\documentclass{article}

\usepackage{multirow}
\usepackage{graphicx}
\usepackage{wrapfig}
\usepackage{subcaption}

\usepackage{amsthm}
\usepackage{amsmath}
\usepackage{algorithm}
\usepackage{algpseudocode} 
\usepackage{float}
\usepackage{placeins}

\theoremstyle{definition}
\newtheorem{definition}{Definition}[section]

\usepackage[normalem]{ulem}

\usepackage[dvipsnames]{xcolor}

    \usepackage[final]{neurips_2025}

\usepackage[utf8]{inputenc} %
\usepackage[T1]{fontenc}    %
\usepackage{hyperref}       %
\usepackage{url}            %
\usepackage{booktabs}       %
\usepackage{amsfonts}       %
\usepackage{nicefrac}       %
\usepackage{microtype}      %
\usepackage{xcolor}         %

\title{LoRA vs Full Fine-tuning: An Illusion of Equivalence}

\author{
Reece Shuttleworth \quad Jacob Andreas \quad Antonio Torralba \quad Pratyusha Sharma \\
MIT CSAIL \\
\texttt{\{rshuttle, jda, torralba, pratyusha\}@mit.edu}
}

\begin{document}

\maketitle

\begin{abstract}
Fine-tuning is a crucial paradigm for adapting pre-trained large language models to downstream tasks. Recently, methods like Low-Rank Adaptation (LoRA) have been shown to effectively fine-tune LLMs with an extreme reduction in trainable parameters. But, \emph{are their learned solutions really equivalent?} We study how LoRA and full-finetuning change pre-trained models by analyzing the model's weight matrices through the lens of their spectral properties. We find that LoRA and full fine-tuning yield weight matrices whose singular value decompositions exhibit very different structure: weight matrices trained with LoRA have new, high-ranking singular vectors, which we call \emph{intruder dimensions}, while those trained with full fine-tuning do not.
Further, we extend the finding that LoRA forgets less than full fine-tuning and find its forgetting is vastly localized to the intruder dimension -- by causally intervening on the intruder dimensions by changing their associated singular values post-fine-tuning, we show that they cause forgetting. Moreover, scaling them down significantly improves modeling of the pre-training distribution with a minimal drop in downstream task performance.
Given this, we should expect accumulating intruder dimensions to be harmful and lead to more forgetting. This will be amplified during continual learning because of sequentially fine-tuning, and we show that LoRA models do accumulate intruder dimensions here tend to perform worse in this setting, emphasizing the practicality of our findings.

\end{abstract}

\section{Introduction}
\label{sec:intro}

\begin{wrapfigure}{hR}{0.50\columnwidth}  %
  \vspace{-19pt}                           %
  \centering
  \includegraphics[width=\linewidth]{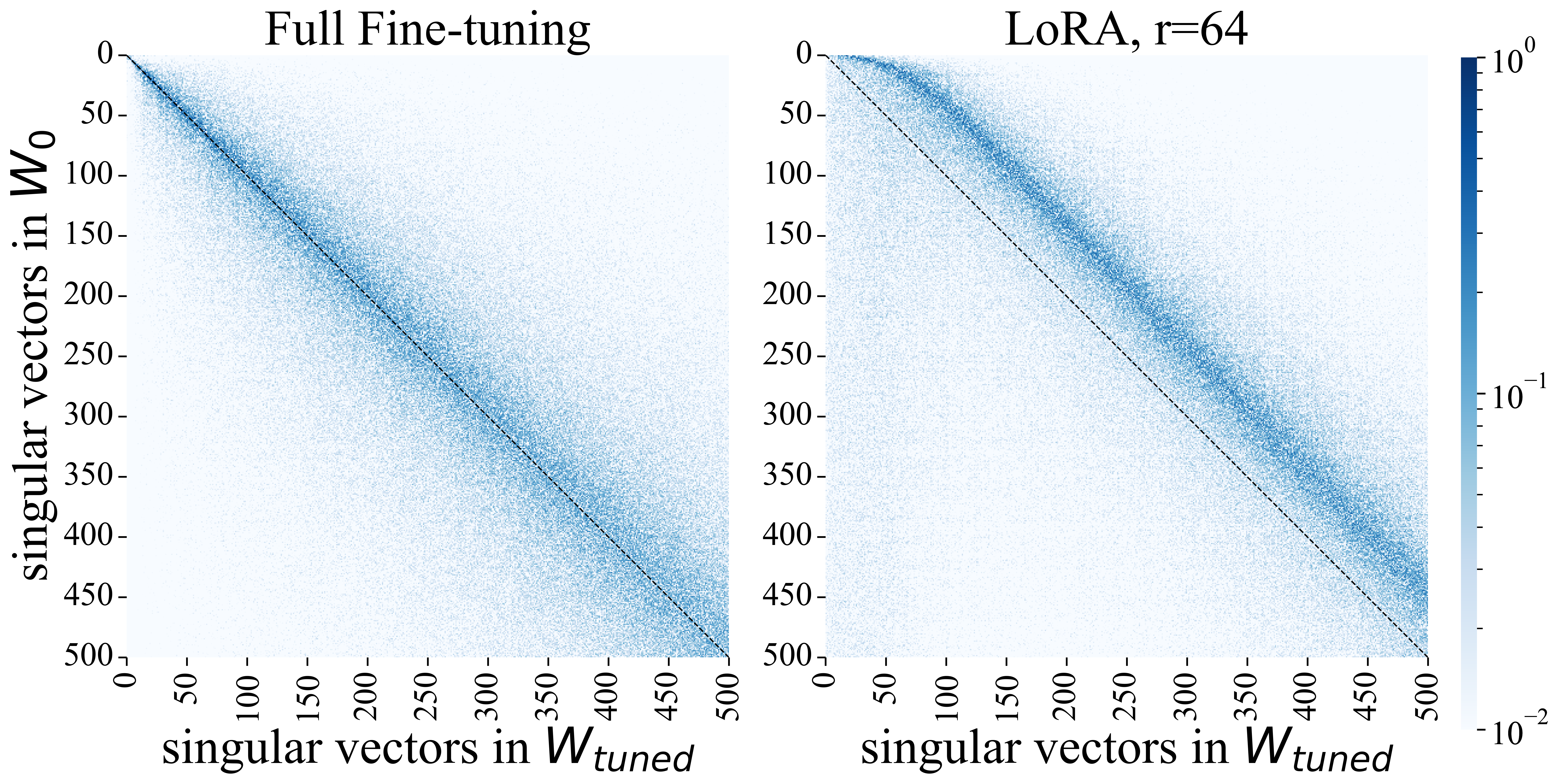}
  \vspace{-6pt}
  \caption{\small \textbf{LoRA and full fine-tuning update the parameter space differently.} Similarity matricies of pre- and post-fine-tuning singular vectors for LLaMA2-7B that characterize the spectral differences introduced during fine-tuning. Full fine-tuning retains most of the pre-training structure, while LoRA has a diagonal shift. Color shows cosine similarity.}
  \vspace{-6pt}
  \label{fig:magicoder-similarity-matrix}
\end{wrapfigure}

Adapting large, pre-trained models to downstream tasks via fine-tuning is a computation- and data-efficient way to create domain-specific models for a variety of tasks. The simplest approach is to fine-tune all parameters of the pre-trained model on downstream task data \citep{devlin2019bertpretrainingdeepbidirectional, instructgpt}. However, as pre-trained models grow larger, full fine-tuning becomes increasingly challenging and expensive. 
Recently, parameter-efficient fine-tuning (PEFT) methods, especially low-rank adaptation (LoRA; \citealp{lora}), have been shown to enable fine-tuning with only a fraction of the trainable parameters. \textbf{While LoRA can match full fine-tuning performance, are the solutions learned by the two methods similar?}

\begin{figure*}[h!]
\centering
\includegraphics[width=1.\textwidth]{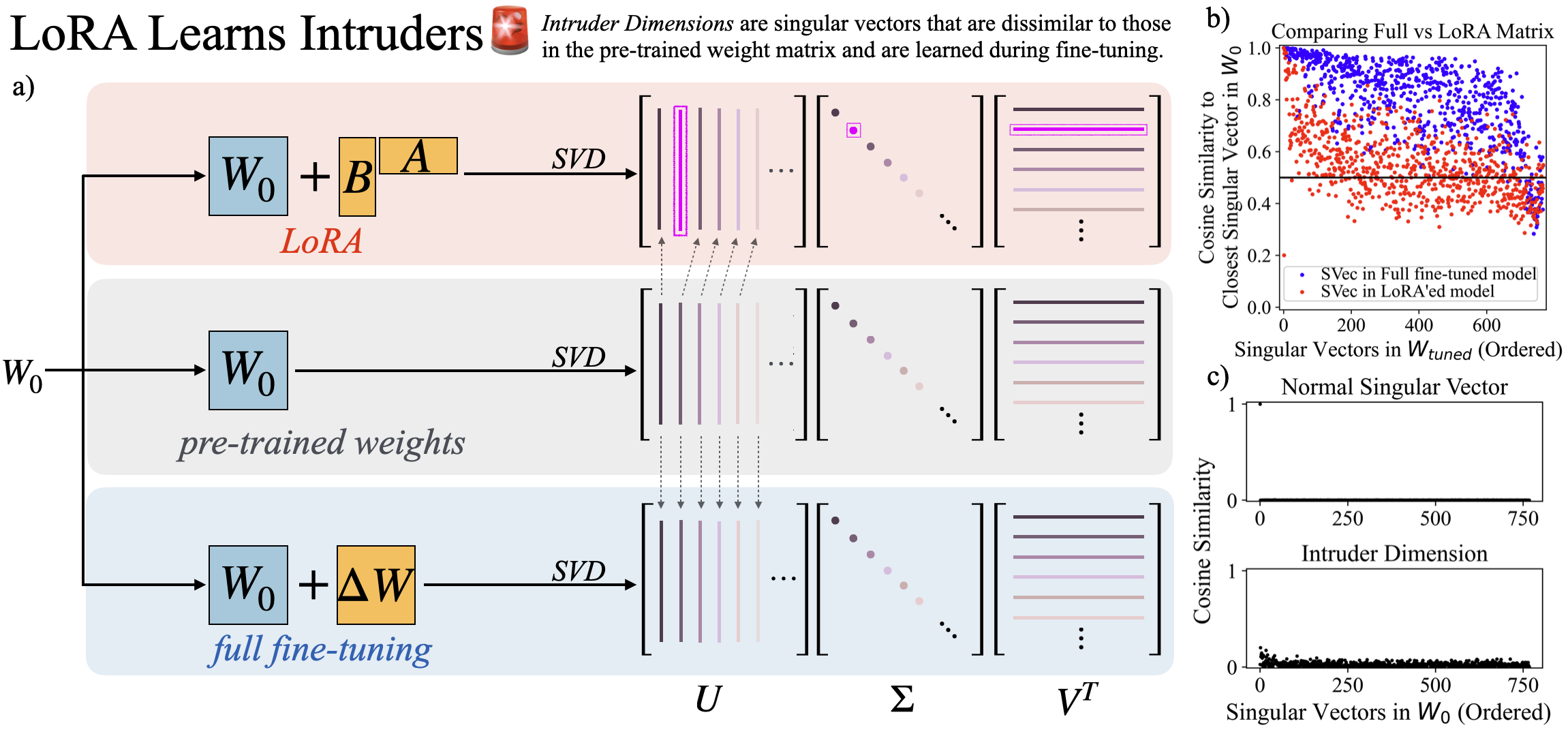}
\vspace{-5pt}
\caption{\small \textbf{Characterizing structural differences between solutions learnt by LoRA \& Full Fine-tuning.} \textbf{a)} We measure the changes to the SVD of the pre-trained weights made during fine-tuning. We observe \textit{intruder dimensions} introduced by LoRA in top ranking singular vectors but not by full fine-tuning. \textbf{b)} Comparing a matrix fine-tuned with full fine-tuning or LoRA. 
\textbf{c)} The intruder dimension shows near-zero absolute cosine similarity with all pre-trained singular vectors, in contrast to other singular vectors of the finetuned matrix.}
\label{ourmethod}
\vspace{-5pt}
\end{figure*}

While full fine-tuning treats every parameter as trainable, LoRA treats the learned update to a weight matrix as the product of two low-rank matrices \citep{lora}. While this parameterization is empirically effective, a principled explanation of the mechanism by which it matches the full fine-tuning performance has remained elusive. One explanation is offered by the \emph{intrinsic dimension hypothesis} \citep{li2018measuringintrinsicdimensionobjective, aghajanyan2020intrinsic}, which posits that the update learned via fine-tuning has a low intrinsic rank, suggesting that LoRA might recover an approximately equivalent solution to full fine-tuning. However, prior work has observed differences in the ability of LoRA and full fine-tuning to independently change the angle and magnitude with which a neuron transforms its input \citep{dora}. Moreover, other work has also observed that LoRA has difficulty matching the performance of full fine-tuning on difficult tasks like code generation \citep{biderman2024loralearnsless, zhuo2024astraiosparameterefficientinstructiontuning} and long-form text generation \citep{ivison2023camelschangingclimateenhancing}. Therefore, it is unclear if these findings indicate a limit in LoRA's ability to fit to a specific downstream task, or if these methods learn inherently different solutions.

In this paper, we show 
that full fine-tuning and LoRA learn different solutions with characteristic differences in their spectral properties (as shown in Fig.~\ref{fig:magicoder-similarity-matrix} for LLaMA2-7B \citep{llama2}) and that these spectral differences are causally related to different model behaviors. We observe:

1. \textbf{LoRA and full fine-tuning produce structurally different parameter updates, characterized by the existence of \textit{intruder dimensions} in weight matrices tuned by LoRA}. Intruder dimensions are singular vectors with large associated singular values that are very dissimilar to the singular vectors in the pre-trained weight matrix. In contrast, fully fine-tuned models remain spectrally similar to the pre-trained model and do not contain intruder dimensions.

2. \textbf{LoRA forgets less than full fine-tuning...but not always.} We extend the findings of \citet{biderman2024loralearnsless} that LoRA forgets less to the case \emph{even} when there is equal fine-tuning performance between LoRA and full fine-tuning, showing that a difference in fit is not simply the cause of this finding but rather is inherent to these methods. 
However, this is not always the case:
despite nearly identical fine-tuning task accuracies, we show that different selections of LoRA alpha and learning rate lead to starkly different generalization behaviors, even leading to LoRA forgetting more than full fine-tuning. We also find that models with the best generalization for each of these hyperparameter settings have the fewest intruder dimensions.

3. \textbf{Intruder dimensions cause forgetting of the pre-training distribution.} Scaling down the associated singular values of high-ranking intruder dimensions leads to a large drop in loss on the pre-training distribution (forgetting) but only a minimal drop in test performance. The drop in forgetting we observe when scaling down singular vectors is unique to intruder dimensions and indicates that they interfere with the pre-trained language modeling ability of these models.
Given this finding, we should expect accumulating intruder dimensions to be harmful and lead to more forgetting. To amplify this accumulation and examine its effect, we fine-tune in a continual learning setting (sequentially fine-tuning on many tasks) and show that LoRA models do indeed tend to forget more on previously learned tasks in this setting, providing additional support for our findings.

\section{Background \& Related Work}

\textbf{Methods for fine-tuning.} Pre-trained language models offer a foundation for downstream applications, eliminating the need to train from scratch \citep{instructgpt, devlin2019bertpretrainingdeepbidirectional}. Full fine-tuning, in which every parameter of a pre-trained model is updated, is commonly used \citep{devlin2019bertpretrainingdeepbidirectional, liu2019robertarobustlyoptimizedbert}. Low Rank Adaptation (LoRA; \citealp{lora}), which represents the update to the weights as a product of two low-rank matrices, reduces computation and memory requirements relative to full fine-tuning. Past work has shown that LoRA matches full fine-tuning performance for tasks like sequence classification \citep{lora}, instruction tuning \citep{qlora, ghosh2024closerlooklimitationsinstruction}, and chat \citep{qlora}. Other work has shown a gap in performance on harder tasks like code generation \citep{biderman2024loralearnsless, zhuo2024astraiosparameterefficientinstructiontuning}. We focus our investigation on both cases to ensure our findings generalize to all use cases.

\textbf{LoRA, formally.} Given a pre-trained weight matrix $W_0 \in \mathbb{R}^{m \times n}$, full fine-tuning treats the learned matrix update as $\Delta W \in \mathbb{R}^{m \times n}$. Instead, LoRA decomposes $\Delta W$ into a product of two matrices such that $\Delta W = BA$, where $B \in \mathbb{R}^{m \times r}$, $A \in \mathbb{R}^{r \times n}$, and where the rank $r$ is generally $r \ll min(m,n)$. During prediction, $$Y=W_{tuned}X=(W_0+\frac{\alpha}{r}BA)X ~ .$$

$B$ is initialized to zero, and $A$ sampled from an isotropic Gaussian. All parameters in $B$ and $A$ are trained. From this we can see that while full fine-tuning has $mn$ trainable parameters per weight matrix, LoRA only has $mr+rn$. See Appendix \ref{derivation_grads} for derivation of LoRA adapter gradients.

\textbf{LoRA Variants.} Many variations of LoRA exist. Methods improve LoRA's performance or memory-efficiency by initializing with the principal \citep{meng2024pissaprincipalsingularvalues} or minor \citep{wang2024miloraharnessingminorsingular} components of the underlying weight matrix, training with quantization \citep{qlora}, adaptively allocating different ranks \citep{zhang2023adaloraadaptivebudgetallocation}, or sequentially training multiple LoRAs \citep{xia2024chainloraefficientfinetuning}. Other methods propose similar but alternative architectures \citep{dora, vera, nola}. Other work has also proposed low rank manipulations to the activations instead of the weights \citep{reft}. Although the primary focus of our study is on the original LoRA setup \citep{lora}, we also study a few LoRA variants (Appendix \ref{lora-variants-text-appendix}).
While we leave a rigorous analysis of all possible variants to future work, our preliminary experiments show that our findings generalize to several variants. 
Additionally, we also demonstrate the robustness of our findings across a range of LoRA hyperparameter settings (Appendices \ref{learning-rate-text}, \ref{case-study-text-appendix}, \ref{random-seeds-text-appendix}).

\textbf{Analysis of Solutions.} The intrinsic dimension measure \citep{li2018measuringintrinsicdimensionobjective} was used by \citet{aghajanyan2020intrinsic} to argue that the fine-tuning update for a pre-trained LLM has low intrinsic rank, explaining why only a small number of trainable parameters are necessary to reach 90\% of full fine-tuning performance. This finding motivated \citet{lora} to hypothesize that LoRA works because solutions of low intrinsic rank exist. But to our knowledge, no past work has compared the rank (or other properties of weight matrices) between LoRA and full-fine tuning on tasks where they are matched in performance.
While \citet{dora} showed that LoRA has difficulty changing directional and magnitude components of a neuron independently, it is unclear if this difference is due to an inability of LoRA to fit as well as full fine-tuning to the adaptation task.

\textbf{Relation to \citet{biderman2024loralearnsless}.} Recent work comparing LoRA to full fine-tuning has found that LoRA forgets less when fine-tuned on math and code \citep{biderman2024loralearnsless} and more closely resembles the pre-trained model \citep{ghosh2024closerlooklimitationsinstruction}. We extend the findings of \citet{biderman2024loralearnsless} to the case when there is equal fine-tuning performance between LoRA and full fine-tuning, showing that a difference in fit to the fine-tuning task is not simply the cause of this finding but rather is inherent to these methods.

\textbf{Singular Value Decomposition.} The SVD decomposes a matrix $M \in \mathbb{R}^{m \times n}$ such that $M=U\Sigma V^T$, where $U \in \mathbb{R}^{m \times m}$ and $V \in \mathbb{R}^{n \times n}$ have orthonormal columns representing the singular vectors of $M$ and $\Sigma \in \mathbb{R}^{m \times n}$ is a diagonal matrix containing the singular values of $M$. $U$ and $V^T$ represent rotations that matrix $M$ performs, while $\Sigma$ represents scaling along those axes. Singular vectors, ordered by singular values, reveal a matrix's most important axes of transformation.
 
\section{Structural Differences}
\label{section:model-diffs}

\begin{wrapfigure}{hR}{0.5\columnwidth}  %

\vspace{-7pt}

\centering
\includegraphics[width=1.0\linewidth]{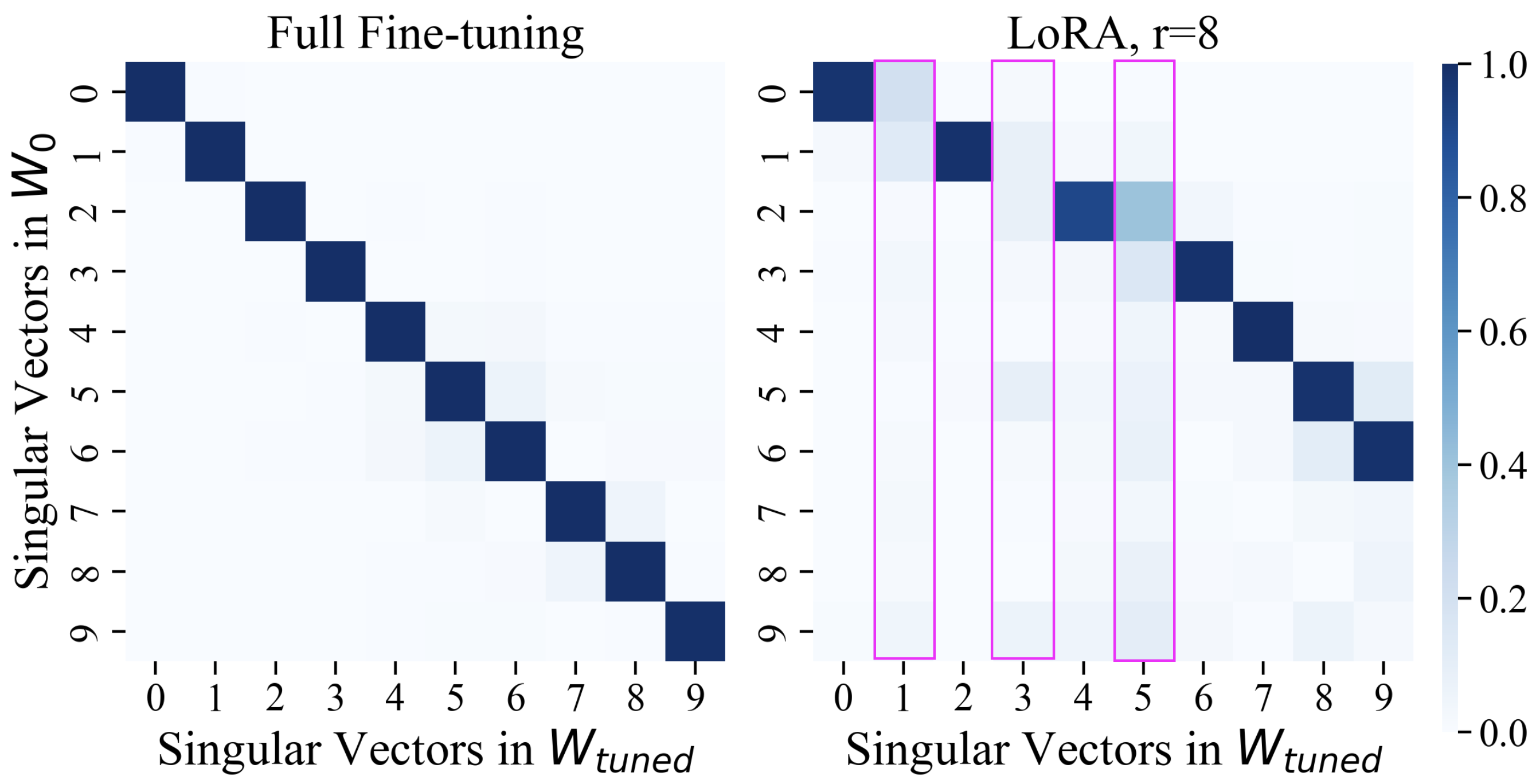}
\vspace{-5pt}
\caption{\small
\textbf{LoRA and full fine-tuning learn distinct structural solutions.} LoRA introduces \emph{intruder dimensions} (represented by outlined columns).
}
\label{similarity_matrix}
\vspace{-10pt}
\end{wrapfigure}

Inspired by \citet{laser}’s finding that the singular value decomposition (SVD, \citealp{svd}) can be used to selectively prune singular vectors to improve model performance, this paper adopts the SVD of neural network parameters as a lens for understanding the changes that fine-tuning makes to pre-trained weights. Understanding how these dimensions change can give us insight into how a particular fine-tuning method changes the pre-trained model. 
In particular, we study how well singular vectors in weight matrices fine-tuned with LoRA or full fine-tuning map to singular vectors in the pre-trained weights (using cosine similarity).

Visually, we observe in Fig.~\ref{ourmethod}(b) that LoRA and full fine-tuning's singular vectors have very different similarities to the pre-trained singular vectors: singular vectors of models fine-tuned with LoRA appear to have, on average, much lower cosine similarity to pre-trained singular vectors in comparison to full fine-tuning. Interestingly, in LoRA fine-tuned models, we also observe the presence of high ranking singular vectors with very low cosine similarity to any pre-trained singular vector.\footnote{Recall that in high dimensions, a vector can have low cosine similarity to a set of orthogonal vectors that span a space; see Appendix \ref{section:orthogonal-cosine-appendix} for discussion.} In Fig.~\ref{ourmethod}(c), we show the difference between these vectors with low cosine similarity to the pre-trained singular vectors and normal singular vectors from the fine-tuned weights. This “new” dimension can be seen in Fig.~\ref{ourmethod}(b) as the lone red dot in the bottom left corner. We name these “new” dimensions \textit{intruder dimensions}, which we define formally as follows:

\begin{definition}
A singular vector $y_j$ from the fine-tuned weight matrix $W_{tuned}$ is an \textbf{intruder dimension} if and only if $\max_{i}(cos(y_j, x_i)) < \epsilon$, where $\epsilon$ is a similarity threshold and $x_i$ are the singular vectors of $W_{0}$.
\label{intruder_definition}
\end{definition}

\begin{wrapfigure}[17]{r}{.50\columnwidth}  %
  \vspace{-8pt}                              %
  \begin{minipage}{\linewidth}
    \begin{algorithm}[H]                     %
      \caption{Finding intruder dimensions.}
      \begin{algorithmic}[1]                 %
        \Require Pre‑trained weights $W_0$, fine‑tuned weights $W_{\text{t}}$,  
                cosine similarity threshold $\epsilon$, \# of fine-tuned singular vectors to examine $k$.
        \State $[U_0,\Sigma_0,V_0^{\!\top}] \gets \text{SVD}(W_0)$
        \State $[U_t,\Sigma_t,V_t^{\!\top}] \gets \text{SVD}(W_{\text{tuned}})$
        \State ${{\rm n\_intruders}} \gets 0$
        \State $n \leftarrow$ \# of pre-trained singular vectors
        \For{$j\gets1$ \textbf{to} $k$}
          \If{$\displaystyle\forall i\le n:\,
               \cos\!\bigl(U_0[i],\,U_t[j]\bigr) < \epsilon$}
             \State ${{\rm n\_intruders}}\gets {{\rm n\_intruders}}+1$
          \EndIf
        \EndFor
        \State \Return ${{\rm n\_intruders}}$
      \end{algorithmic}
      \label{intruder-alg}
    \end{algorithm}
  \end{minipage}
  \vspace{-3pt}
\end{wrapfigure}

Examples of intruder dimensions may be seen in Fig.~\ref{similarity_matrix}. Here, we plot the similarity matrix between the top 10 singular vectors (ranked by singular value) in the pre-trained and fine-tuned matrices. While full fine-tuning appears to have a clear one-to-one mapping, LoRA appears to have its mapping shifted by “blank” columns (outlined in magenta): these are intruder dimensions, with low cosine similarity to every pre-trained singular vector. A zoomed out version of this plot can be seen in Fig.~\ref{fig:magicoder-similarity-matrix}, in which we see an off diagonal shift due to intruder dimensions.

It is important to note that in the case of full fine-tuning, the singular vectors that map to a pre-trained singular vector with high cosine similarity also have similar singular values. From these initial measurements, it appears that LoRA and full fine-tuning have structural differences in the changes they make to the pre-trained weights: while full fine-tuning appears to make small changes to the existing singular vectors and singular values, LoRA introduces new singular vectors that have a large contribution to the norm of the updated parameter matrix.

\textbf{Our Models.} We study LLaMA2-7B \citep{llama2} and RoBERTa-base \citep{liu2019robertarobustlyoptimizedbert}. RoBERTa-base is a pre-trained encoder-only language model and we fine-tune it on six different sequence classification tasks. See Appendix \ref{section:roberta-details} for fine-tuning details.
LLaMA2-7B is a pre-trained decoder-only language model, and we study it when fine-tuned on either code or math. These checkpoints are provided by \citet{biderman2024loralearnsless}. We also study LLaMA-7B \citep{touvron2023llamaopenefficientfoundation} models fine-tuned on instruction following. See Appendix \ref{huggingface-details} for more details about these models. Importantly, these LLaMA models span math, code, and chat, which are considerably harder than sequence classification tasks. This ensures wide coverage of LoRA use cases.

\begin{figure}[t]
    \centering
    \begin{subfigure}[b]{0.255\linewidth}
        \centering
        \includegraphics[width=\linewidth]{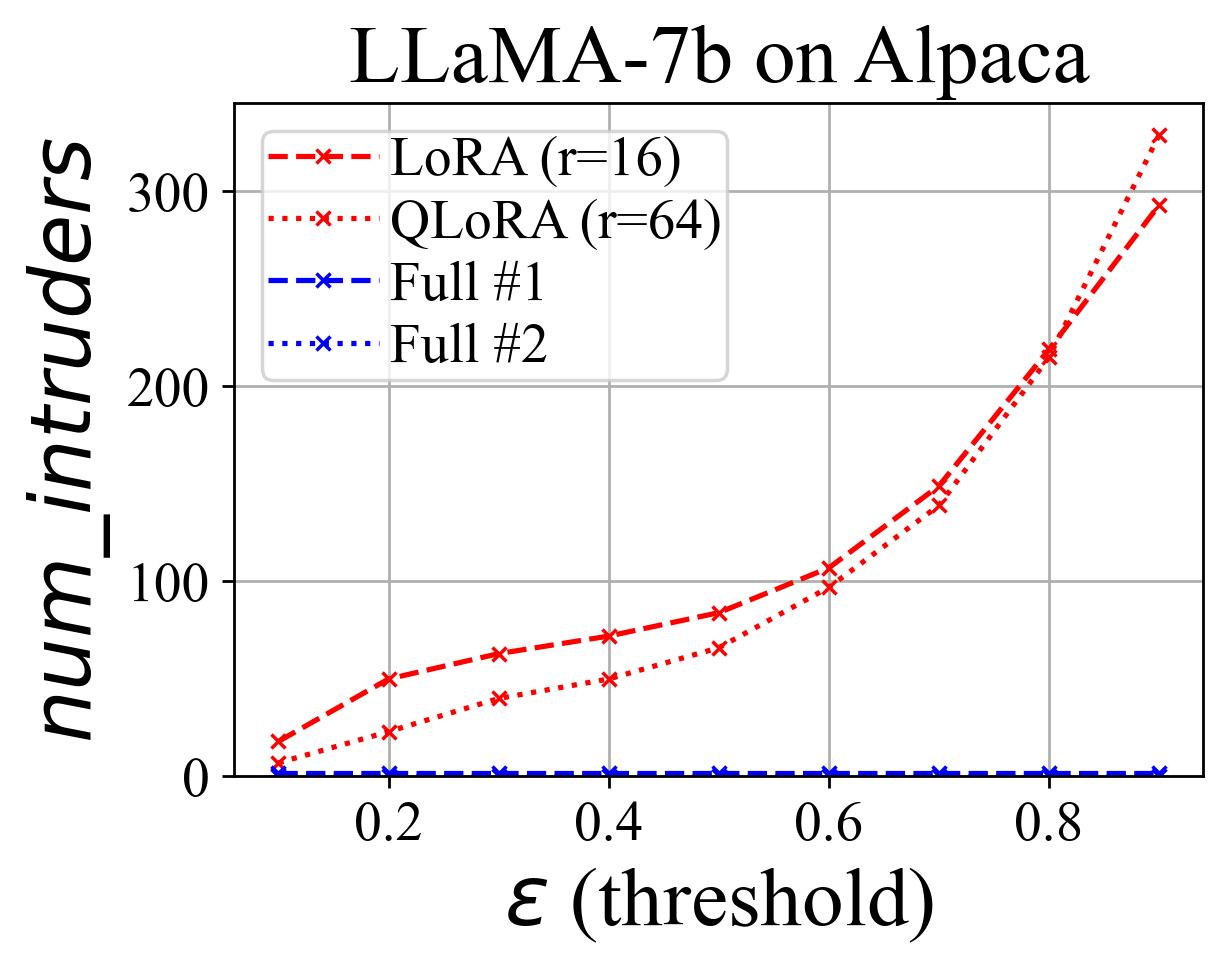}
        \caption{\small \centering LLaMA-7B fine-tuned on Alpaca.}
        \label{llama-epsilon}
    \end{subfigure}
    \begin{subfigure}[b]{0.295\linewidth}
        \centering
        \includegraphics[width=\linewidth]{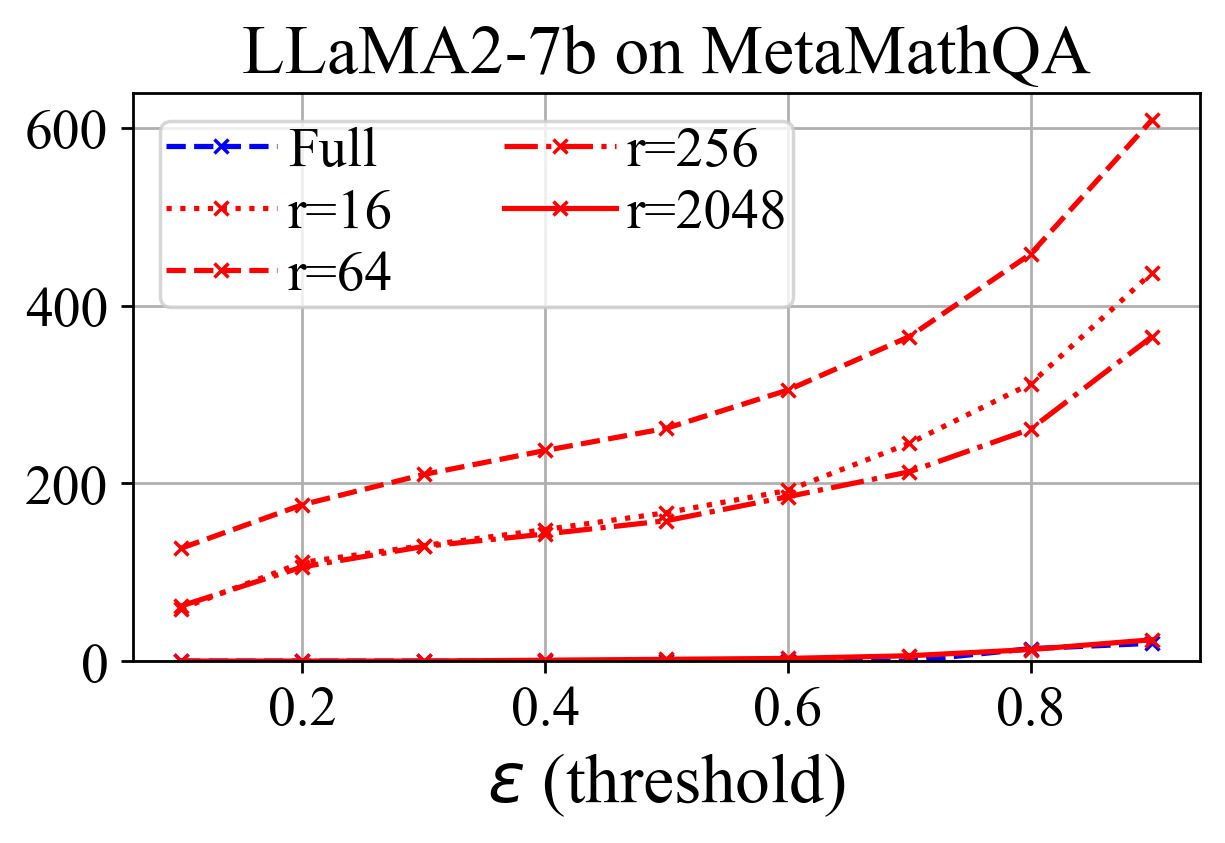}
        \caption{\small \centering LLaMA2-7B fine-tuned on MetaMathQA.}
        \label{llama2-metamath-epsilon}
    \end{subfigure}
    \begin{subfigure}[b]{0.29\linewidth}
        \centering
        \includegraphics[width=\linewidth]{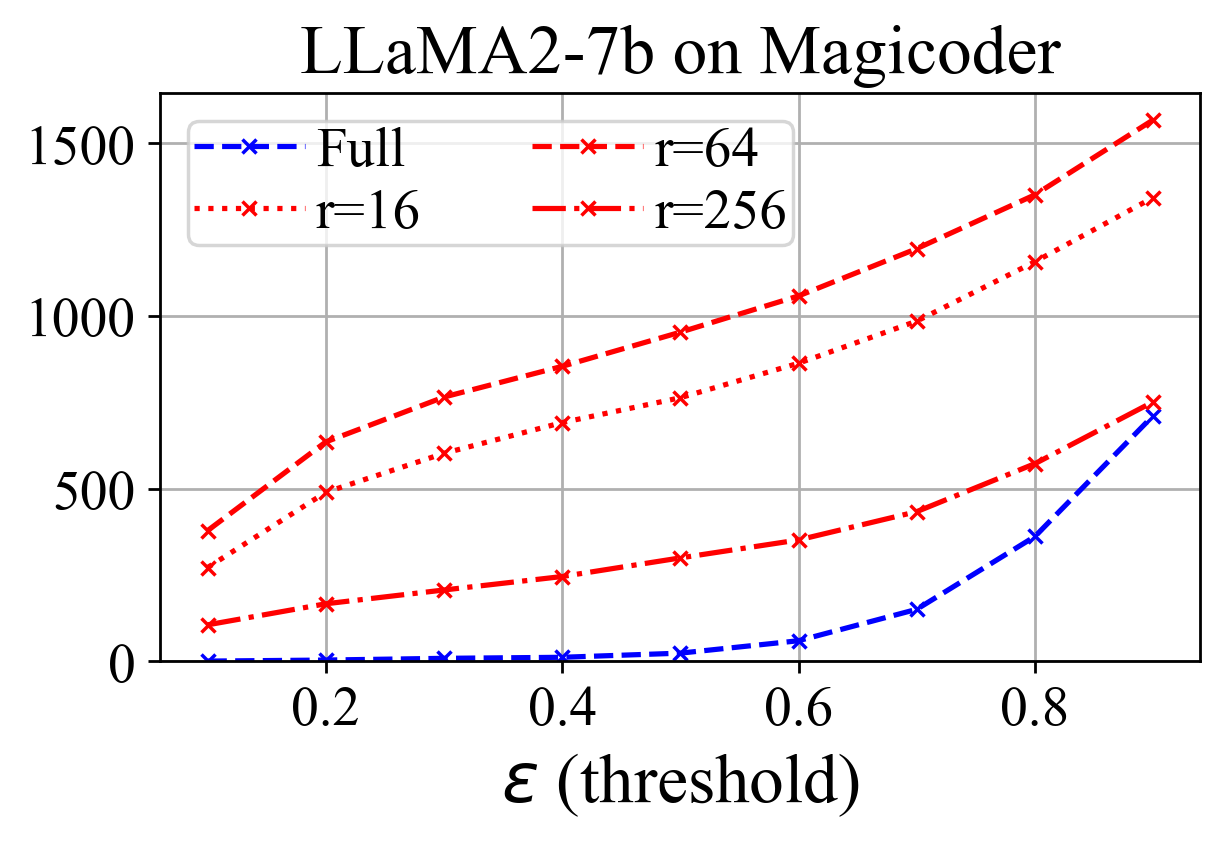}
        \caption{\small \centering LLaMA2-7B fine-tuned on Magicoder-Evol-Instruct.}
        \label{llama2-magicoder-epsilon}
    \end{subfigure}

    \begin{subfigure}[b]{1.0\linewidth}
        \centering
        \includegraphics[width=\linewidth]{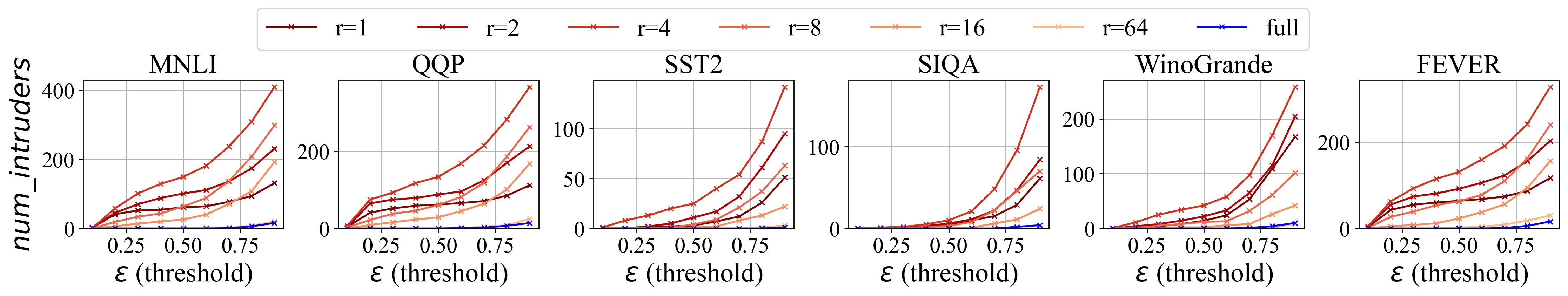}
        \caption{\small Number of intruder dimensions in RoBERTa models fine-tuned on 6 different tasks.}
        \label{roberta-epsilon}
    \end{subfigure}
    \vspace{-5pt}
    \caption{\small \textbf{LoRA has intruder dimensions, whereas full fine-tuning does not.} Here, we set $k=10$ and measure the impact of $\epsilon$ on the number of intruder dimensions measured. LoRA introduces many intruder dimensions in the top 10 ranked singular vectors, while full fine-tuning does not. 
    Numbers are reported are the sums across the entire model.  
    }
    \label{epsilon-sweep}
    \vspace{-5pt}
\end{figure}

\textbf{Our Method.}
To calculate the number of intruder dimensions in a specific weight matrix, we use Algorithm.~\ref{intruder-alg}. In it, we first compute the SVD of both the pre-trained and resulting LoRA and full fine-tuned weights. Then, for each of the top $k$ highest-ranking singular vectors, we measure its maximum cosine similarity with all of the pre-trained singular vectors. If this maximum cosine similarity is less than some threshold $\epsilon$, we classify this singular vector as an intruder dimension. Note that both $k$, the number of fine-tuned singular vectors to examine, and $\epsilon$, the cosine similarity threshold, are hyperparameters; we verify the robustness of our findings for a wide range of $\epsilon$ and $k$ values in Fig.~\ref{epsilon-sweep} and Fig.~\ref{matrix-sweep} respectively. To determine the number of intruder dimensions in a specific model, we run this algorithm
for each weight matrix in the model and sum the total.

\begin{figure*}[t]
        \centering
    \includegraphics[width=1\linewidth]{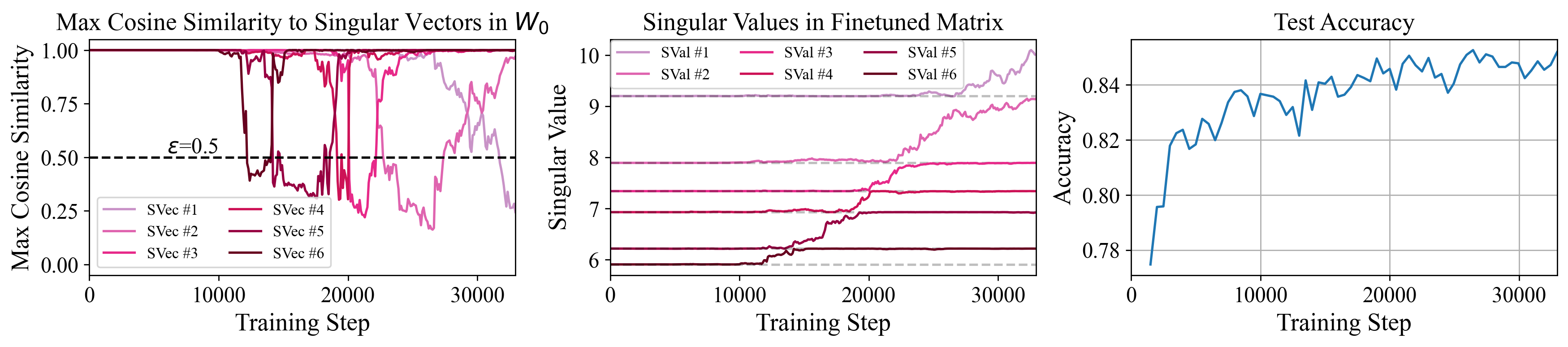}
    \vspace{-5pt}
    \caption{\small \textbf{Evolution of an intruder dimension across training steps.} \emph{(Left)} Intruder dimensions, and their rank, in a LoRA fine-tuned weight matrix during fine-tuning. \emph{(Middle)} Their associated singular values, which shows that the singular value associated with the intruder dimension increases. \emph{(Right)} Test accuracy across training steps.}
    \vspace{-15pt}

    \label{fig:intruders-during-training}
    \end{figure*}

\textbf{LoRA fine-tuned models contain high-ranking intruder dimensions while fully fine-tuned models do not.} To characterize the differences in fine-tuning methods, we first evaluate the differences in the total number of intruder dimensions in the top 10 highest-ranking singular vectors ($k=10$). We repeat this procedure for a range of $\epsilon$ values, our cosine similarity threshold. The results are presented in Fig.~\ref{epsilon-sweep}. For LLaMA2-7B, we find that models trained with LoRA contain intruder dimensions for ranks at least as high as $r \leq 256$. For RoBERTa, we consistently observe intruder dimensions for rank $r \leq 16$, even for low values of $\epsilon$. Interestingly, we observe that fully fine-tuned models, for all model sizes, almost \textit{never} contain intruder dimensions in their top 10 singular vectors, even for epsilon values of about 0.6 to 0.9. This means that full fine-tuning makes smaller changes to the same set of high contribution pre-trained singular vectors, rather than introducing new singular vectors like LoRA.
Importantly, the number of intruder dimensions appears to drop as rank increases past a certain threshold, suggesting that the low-rank nature, as well as the update rule of LoRA, induces them to occur. This is underscored by the $r=2048$ case of LLaMA2-7B fine-tuned on math (Fig~\ref{llama2-metamath-epsilon}), which does not have intruder dimensions and instead has a very similar curve to full fine-tuning. As rank increases past a threshold and LoRA begins to resemble a high rank update, intruder dimensions begin to disappear.

\textbf{LoRA variants have intruder dimensions.} We examine 4 other LoRA variants to ensure that our findings do not only apply to vanilla LoRA. We examine AdaLoRA \citep{zhang2023adaloraadaptivebudgetallocation}, LoRA+ \citep{lora+}, PiSSA \citep{meng2024pissaprincipalsingularvalues}, and VeRA \citep{vera}. In all of these cases, we find intruder dimensions with similar characteristics to vanilla LoRA (see Fig.~\ref{fig:lora-variants}). This shows our findings hold to other variants. For more discussion about these methods, see Appendix \ref{lora-variants-text-appendix}.

\textbf{Intruder dimensions are distributed across both high and low singular values.}
We examine the extent to which intruder dimensions exist throughout the entire weight matrix and how they are distributed. To do this, we hold $\epsilon$ fixed and measure the number of intruder dimensions while varying the proportion of the fine-tuned singular vectors that we examine (Appendix \ref{matrix-sweep-text-appendix}, Fig.~\ref{matrix-sweep}). Here, we can see that LoRA consistently has more intruder dimensions than full fine-tuning, regardless of what fraction of the singular values we examine. 
See Appendix \ref{matrix-sweep-text-appendix} for more discussion.

\textbf{Intruder dimensions increase in magnitude and change in direction as fine-tuning progresses.} To further understand how a particular intruder dimension is introduced during fine-tuning with LoRA, we measure the maximum cosine similarity between the top individual fine-tuned singular vectors and all the pre-trained singular vectors across many intermediate steps in the fine-tuning process, as seen in Fig.~\ref{fig:intruders-during-training} (\emph{left}). In parallel, we track changes in their associated singular values as seen in Fig.~\ref{fig:intruders-during-training} (\emph{middle}). As is evident from the graphs, intruder dimensions appear to gradually increase their ``rank" (\emph{left}) as their singular value is increased (\emph{middle}) while simultaneously changing in direction too as training progresses.

\textbf{Additional empirical observations.} \textbf{1.} We find that the random seed used by LoRA to initialize its adapters plays no role in the resulting structure (see Appendix \ref{random-seeds-text-appendix}). \textbf{2.} We observe that the total number of intruder dimensions increases linearly with respect to the size of the fine-tuning dataset up to a certain point before saturating (Appendix \ref{dataset-size-text-appendix}).
\textbf{3.} We study the effective rank of these fine-tuning updates (Appendix \ref{effective-rank-appendix-text}). However, we find that this measure does not suffice to explain the behavioral differences we observe in LoRA and full fine-tuning. Also, it is important to note that even if it had, its global nature would prevent the precise examinations we conduct in future sections, like in Fig.~\ref{fig:intruder_scaling_onerow}.

\textbf{Experimental and theoretical justification for why intruder dimensions occur.} It is important to note that intruder dimensions are an empirical observation of LoRA. We find that a variety of factors play a role in the introduction of intruder dimensions. In the next section, we present results that suggest that learning rate and LoRA's $\alpha$ contribute to intruder dimensions. In the appendix, we present findings that suggest that tuning the B matrix only leads to fewer intruder dimensions (\ref{tuning_b_text}), and demonstrate how the addition of orthogonal vectors to the pre-trained weight matrix models the introduction of intruder dimensions well (\ref{adding-random-vector-text}).

\section{Model Differences: Forgetting and Out-of-Distribution Generalization}
\label{behavior-section}

\begin{figure}  %
\centering
    \begin{subfigure}[b]{0.45\linewidth}
    \centering
    \includegraphics[width=\linewidth]{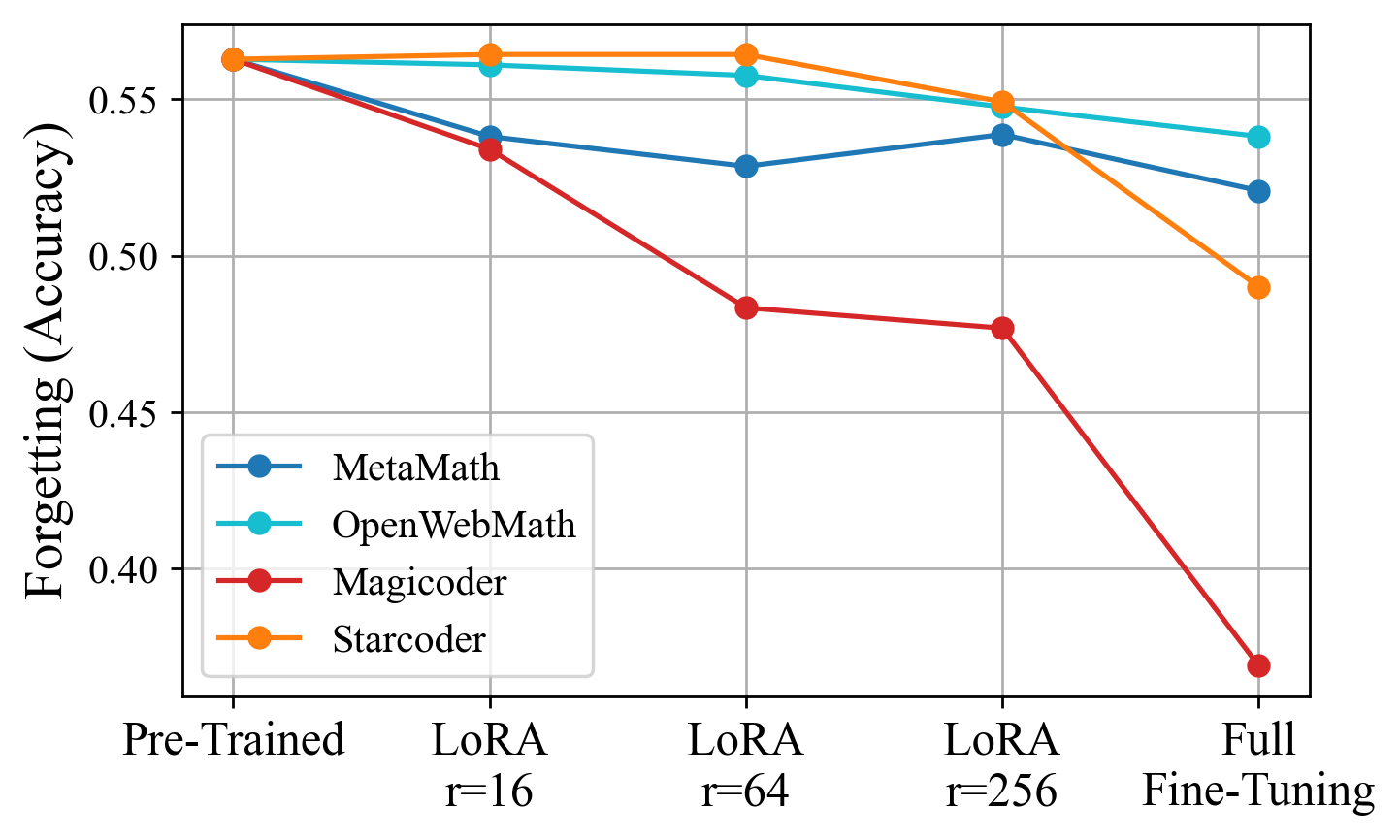}
    \caption{LLaMA2-7B.}
    \end{subfigure}
    \begin{subfigure}[b]{0.45\linewidth}
    \centering
    \includegraphics[width=\linewidth]{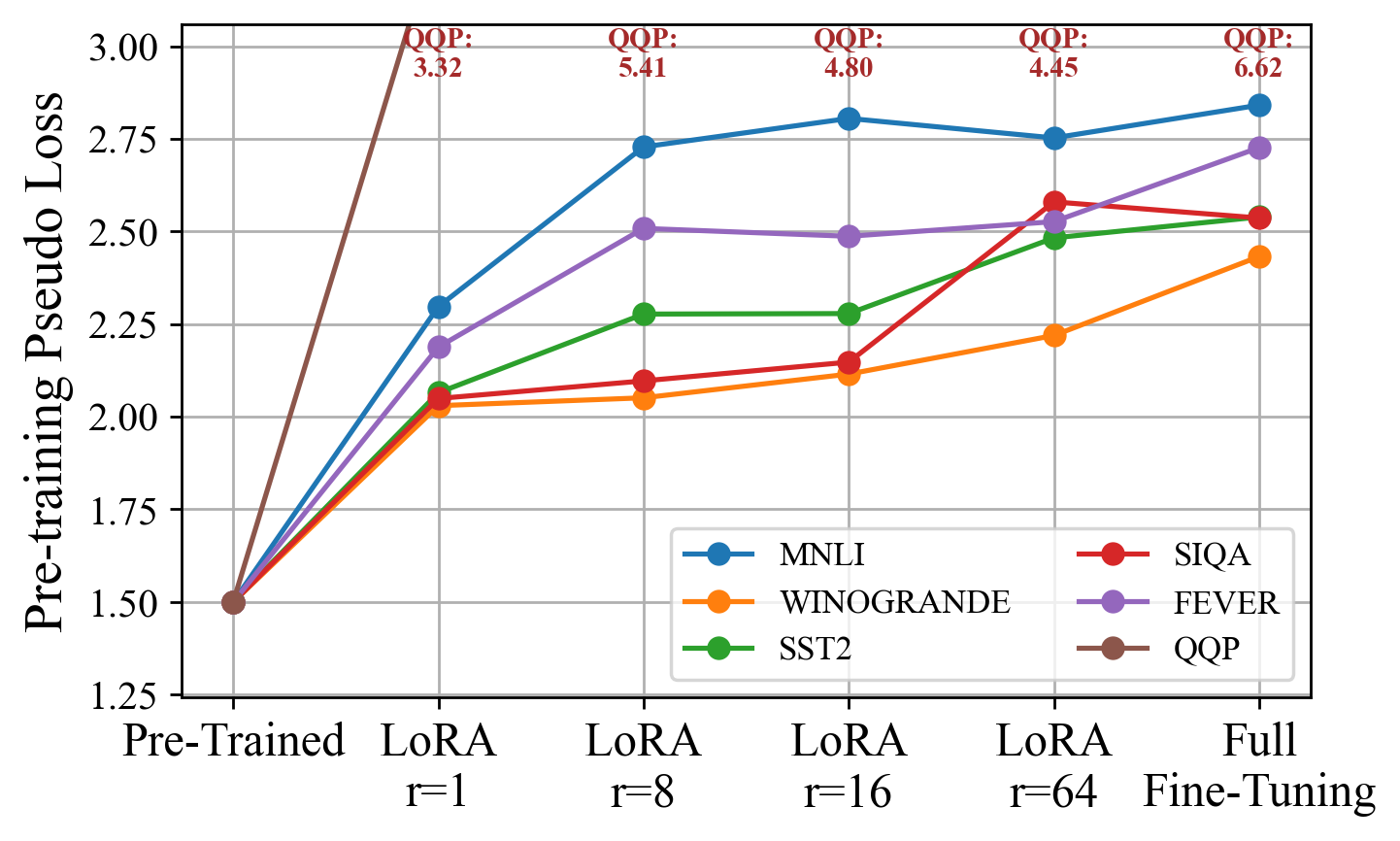}
    \caption{RoBERTa-base.}
    \end{subfigure}
    \caption{\small \textbf{LoRA forgets less, even with same fit to fine-tuning task.} For LLaMA2-7B, forgetting is measured on unrelated tasks, as described in \citet{biderman2024loralearnsless}. For RoBERTa, Pseudo loss on a sample of its pre-training distribution measured as described by \citet{mlm_scoring}. In both, LoRA forgets less than full fine-tuning.}
     \label{pretraining_drift}

\end{figure}

\textbf{LoRA forgets less.} We measure the change in out of distribution performance (forgetting) induced by fine-tuning.
For LLaMA2-7B, we follow \citet{biderman2024loralearnsless} and measure forgetting as the average score on Hellaswag \citep{hellaswag}, WinoGrande \citep{winogrande}, Arc-Challenge \citep{arc-challenge}.
For RoBERTa-base, we measure its ``pseudo-loss'', which is analogous to language modelling loss for encoder-only models, as described by \citet{mlm_scoring} on a sample of its pre-training dataset (as described by \citet{liu2019robertarobustlyoptimizedbert}).
Going forward, we refer to these values as ``forgetting'' and report them in Fig.~\ref{pretraining_drift}. We observe that across all tasks, full fine-tuning forgets more of its pre-training language modeling ability in comparison to LoRA. Importantly, all our RoBERTa-base models fine-tune to equivalent accuracy on the downstream task (Table \ref{a2r_model_accuracies}). This extends the finding that LoRA forgets less \citep{biderman2024loralearnsless} to the case where LoRA and full fine-tuning have equal fit, showing that LoRA forgetting less is not simply a function of it underfitting the fine-tuning task in comparison to full fine-tuning (like in \citet{biderman2024loralearnsless}), but rather a characteristic of LoRA itself.

\textbf{LoRA $\alpha$ impacts generalization and intruder dimensions.} For our experiments, we use the commonly used $\alpha = 2r$ \citep{biderman2024loralearnsless} as well as $\alpha = 8$ \citep{lora}. For both settings of $\alpha$, models obtain equivalent performance on the target task (Tables \ref{model_accuracies} \& \ref{a2r_model_accuracies}). However, when $\alpha = 8$, all ranks of LoRA---even very large ones---exhibit intruder dimensions (Fig.~\ref{roberta-epsilon-a8}), have a much smaller effective rank than when $\alpha = 2r$ (Appendix \ref{effective-rank-appendix-text}), and have much worse generalization (more forgetting, Fig.~\ref{pretraining_drift_a8}). 
Models trained with $\alpha = 2r$ have fewer intruder dimensions and generalize better. This provides additional evidence highlighting the importance of using $\alpha = 2r$ \citep{kalajdzievski2023rankstabilizationscalingfactor, biderman2024loralearnsless}, particularly for higher ranks of LoRA.

\begin{figure}
    \centering
    \includegraphics[width=1.0\linewidth]{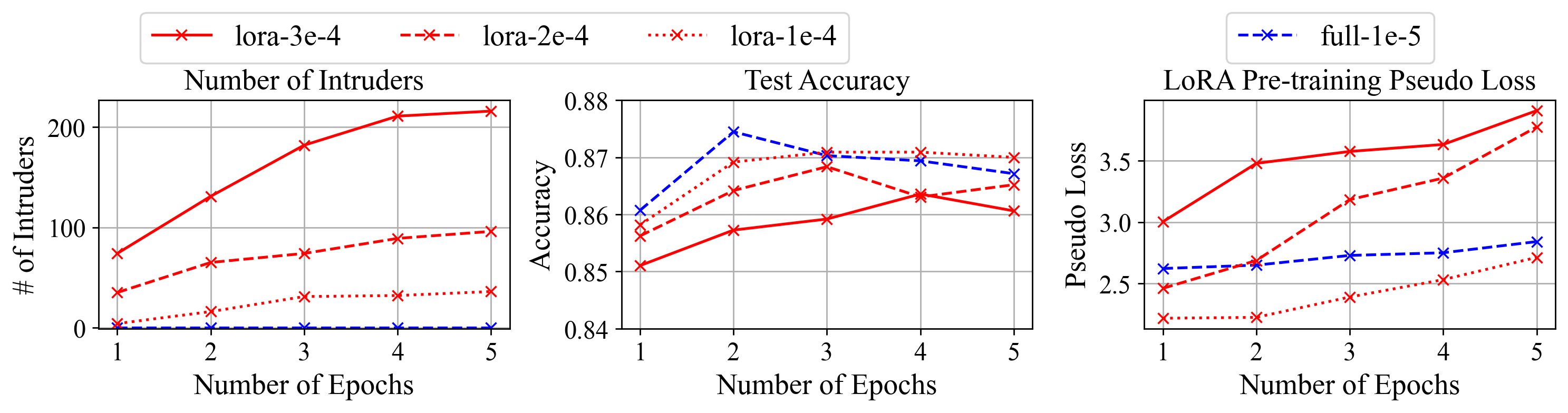}
    \caption{\small
    \textbf{As training progresses, models with growing amount of intruder dimensions continue to forget more, despite non-increasing test performance.} We also measure a strong correlation($\rho$ = 0.971, p-value $\ll 0.001$) between number of intruder dimensions and pre-training pseudo loss. Bigger learning rates lead to more intruder dimensions and forgetting.
    }
    \label{fig:hyperparam_sweep}
\end{figure}

\textbf{An increase in intruder dimensions leads to an increase in forgetting.}
We do a learning rate sweep for RoBERTa-base with LoRA $r=8$ on MNLI to observe its impact. Across epochs, we measure the number of intruder dimensions, test accuracy, and forgetting (pre-training loss). We report the results of these models and our baseline full fine-tuning model in Fig.~\ref{fig:hyperparam_sweep}. We observe that \emph{across} and \emph{within} training runs, as the number of intruder dimensions increase, forgetting (meaning worse generalization) also increases. Test accuracy has no such relation.
Separately, we see that for large learning rates with many intruder dimensions, LoRA models \emph{forget more} than full fine-tuning. This shows that while LoRA in general does forget less than full fine-tuning, it is not a guarantee.

\textbf{Intruder dimensions strongly correlate with forgetting.} When we measure the Spearman correlation between the number of intruder dimensions with forgetting in Fig.~\ref{fig:hyperparam_sweep}, we find an extremely strong fit($\rho=0.971$, p-value $\ll 0.001$). When measuring the same for our LLaMA2-7B models across training epochs (Fig.~\ref{biderman-intruder-correlation}), we still find a strong and still statistically significant relationship ($\rho=0.59$, p-value $= 0.0006$). In contrast, when measuring the correlation between intruder dimensions and test accuracy, we find no statistically significant relationship: for RoBERTa, we measure $\rho = -0.3381$ \& p-value $= 0.218$. For LLaMA2-7B, we measure $\rho= -0.3178$ \& p-value $= 0.0869$. See Appendix \ref{intruder-correlation-text-appendix} for more information.
These results suggest that intruder dimensions are clearly linked with forgetting but \emph{are not necessary for performance.} We examine this claim and whether this relationship is causal in the next section.

\section{Intruder Dimensions Cause Forgetting}

\label{forgetting-text-main}

Previously, we observed that the number of intruder dimensions correlates strongly with forgetting. \textit{Do intruder dimensions cause this forgetting?}

\begin{figure}

    \begin{subfigure}[b]{0.24\linewidth}
        \centering
        \includegraphics[width=1.0\linewidth]{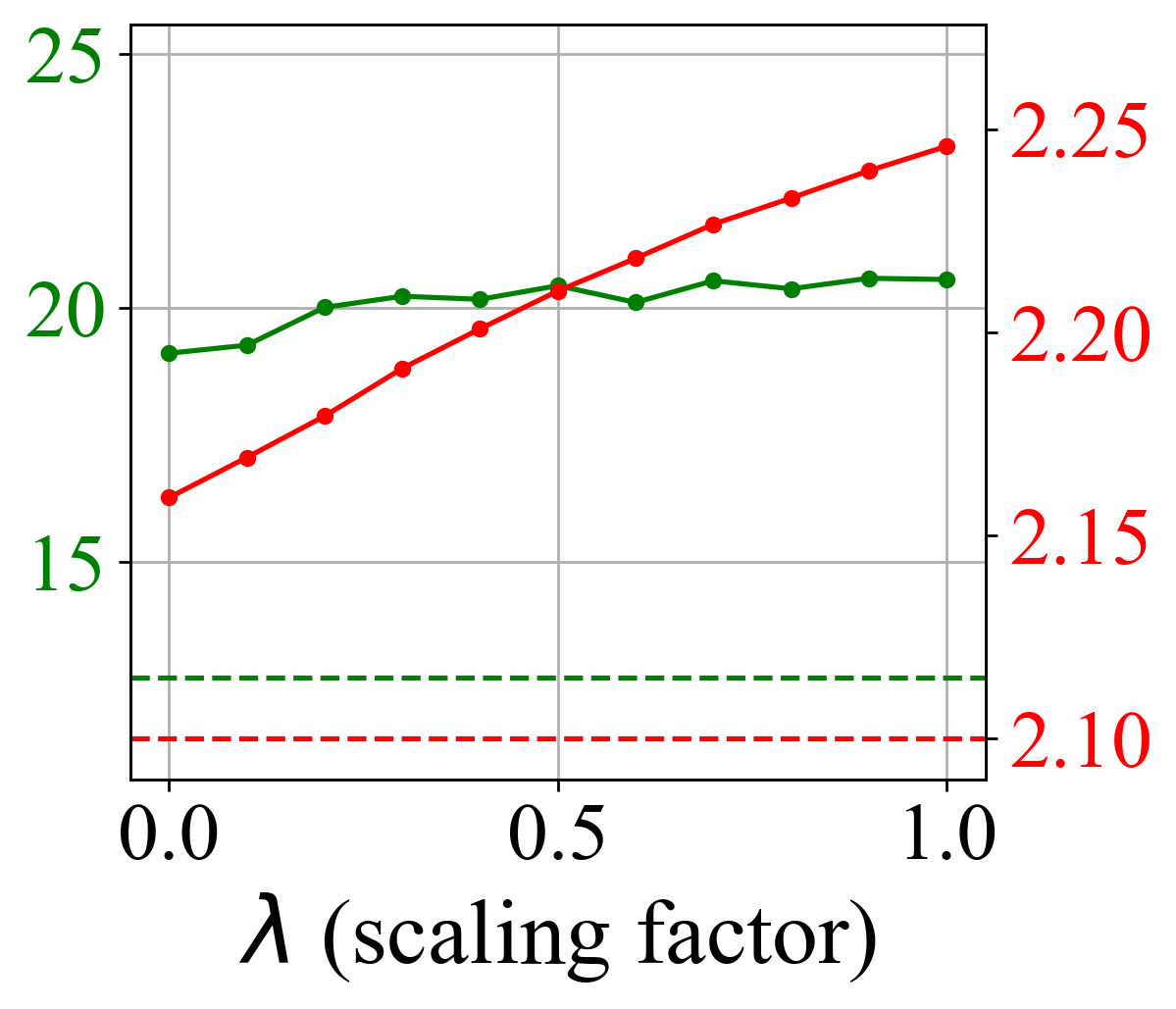}
        \caption{\small \centering LLaMA2-7B\\ on code with r=16.}
        \end{subfigure}
    \begin{subfigure}[b]{0.24\linewidth}
        \centering
        \includegraphics[width=1.0\linewidth]{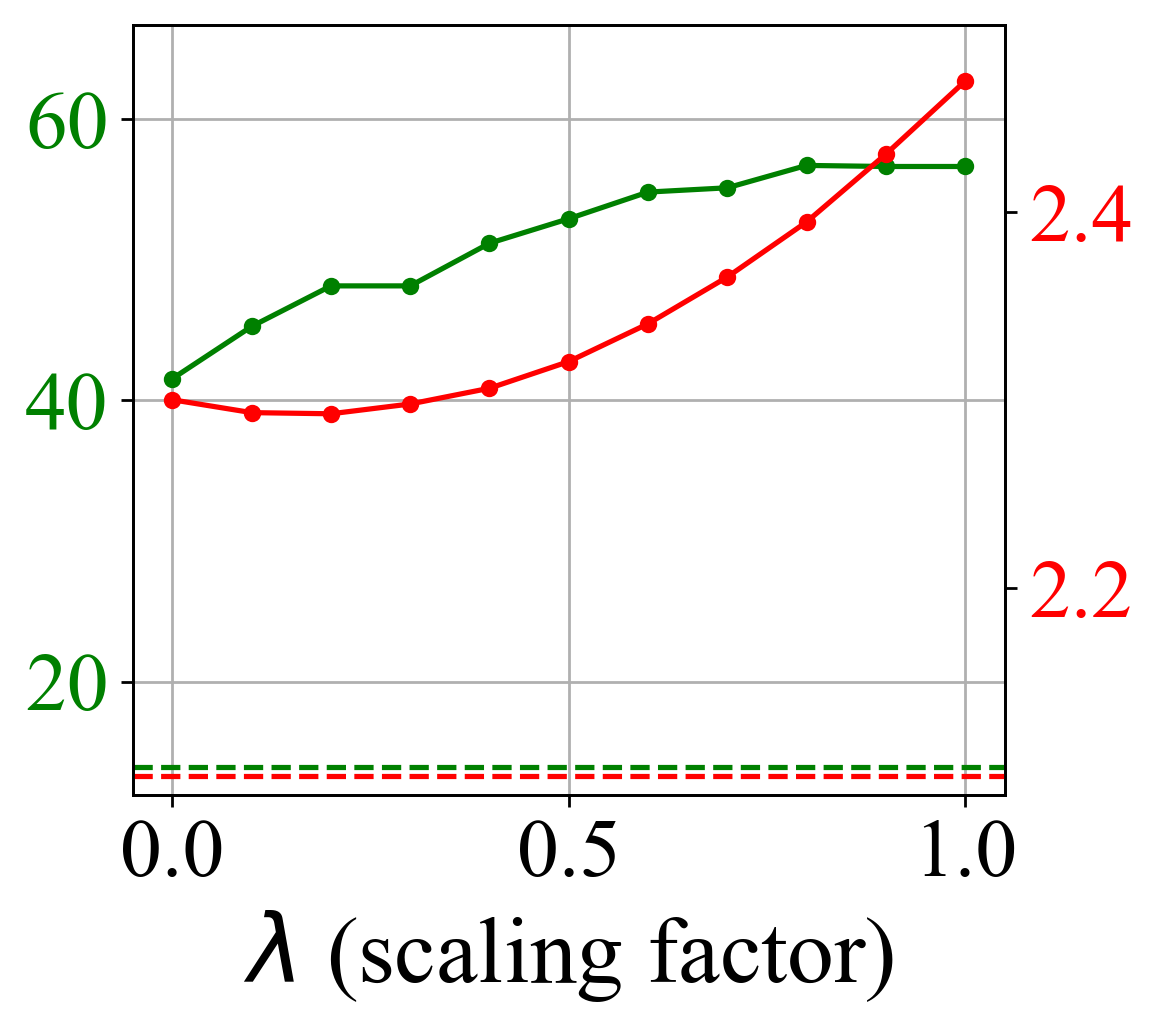}
        \caption{\small \centering LLaMA2-7B\\ on math with r=16.}
    \end{subfigure}
    \begin{subfigure}[b]{0.24\linewidth}
        \centering
        \includegraphics[width=1.0\linewidth]{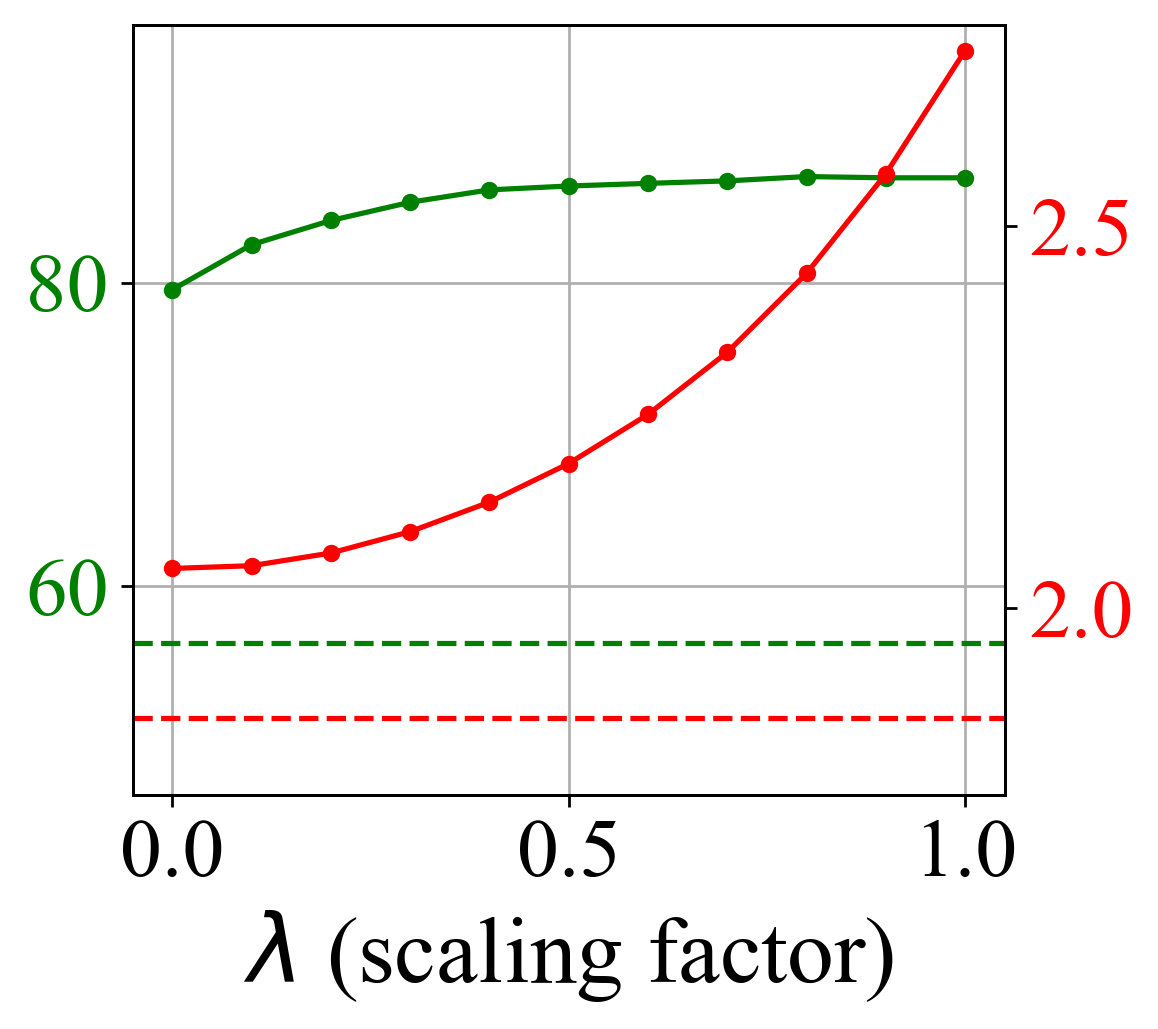}
        \caption{\small \centering RoBERTa-base\\ on MNLI with r=8.}
    \end{subfigure}
    \begin{subfigure}[b]{0.24\linewidth}
        \centering
        \includegraphics[width=1.0\linewidth]{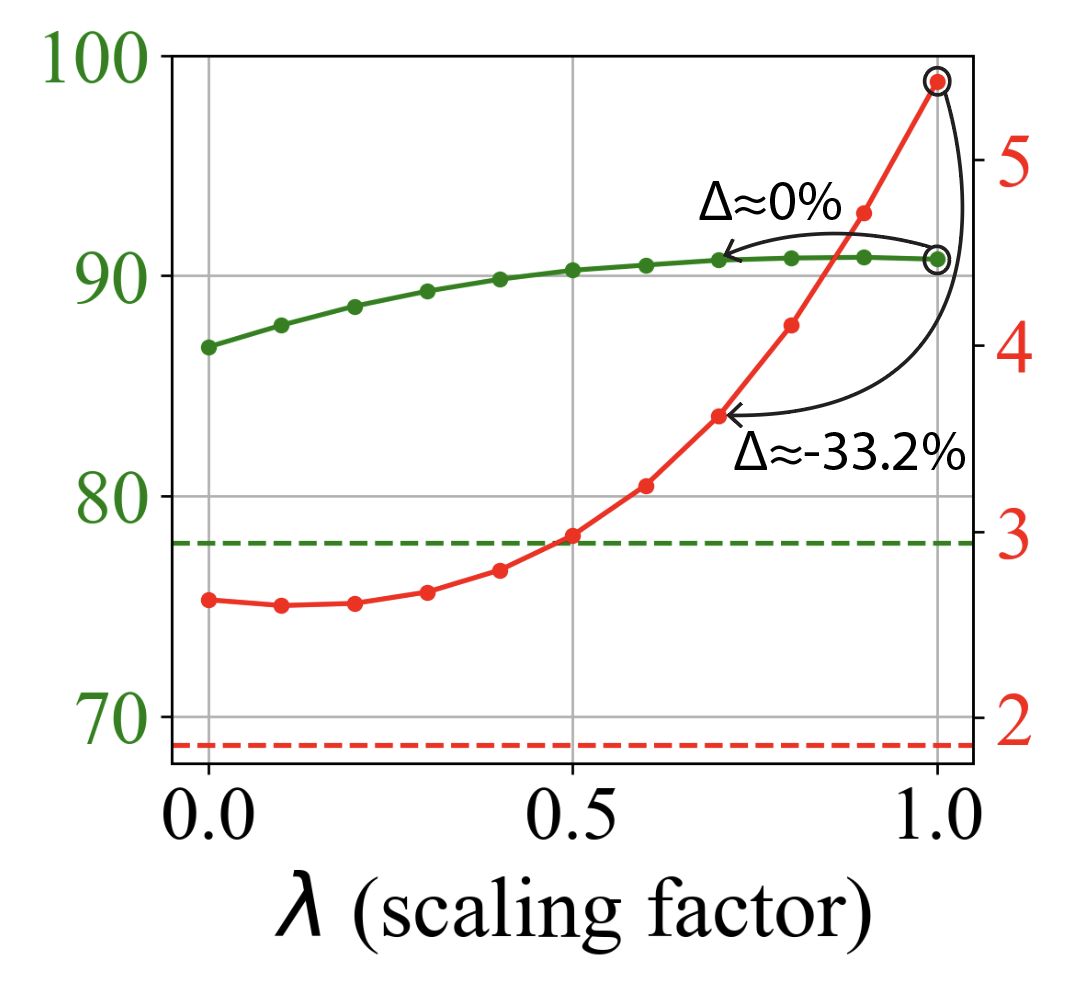}
        \caption{\small \centering RoBERTa-base\\ on QQP with r=8.}
        \label{intruder_scaling_qqp_annotated}
    \end{subfigure}
    \caption{\small  \textbf{Scaling down intruder dimensions in fine-tuned models reduces forgetting but not performance.}  We scale the top intruder dimension in each matrix such that $W = W_0 + \Delta W + (\lambda-1)u_i\sigma_iv^T_i$. Lines represent forgetting (red) and learning (green). Dotted lines represent pre-trained baselines. \textbf{Axis Labels:} \textcolor{ForestGreen}{Green: Test Accuracy (\%).} \textcolor{Red}{Red: Pre-training Loss (Forgetting).}}
    \label{fig:intruder_scaling_onerow}
    \vspace{-10pt}
\end{figure}

\textbf{Scaling the magnitude of intruder dimensions.} To test if intruder dimensions \emph{cause} increased forgetting, we must intervene on intruder dimensions and see the impact. We do this by finding the highest ranked (by singular value) intruder dimension in each weight matrix and scale its contribution such that the new weight matrix is $W = W_0 + \Delta W + (\lambda-1)u_i\sigma_iv^T_i$, where $i$ is the index of the top intruder dimension ($\lambda=0$ is removal and $\lambda=1$ is no change). We sweep $\lambda's$ between 0 and 1, scale the intruder dimensions, and measure test accuracy and pre-training loss. For a comparison baseline, we select the neighbor of the intruder dimension to separately scale. See Fig.~\ref{fig:intruder_scaling_onerow} for results (and Figs.~\ref{roberta_scaling}\&~\ref{llama-intruder-scaling} for full results.)

\textbf{Scaling down intruder dimensions reduces forgetting}. In Fig.~\ref{fig:intruder_scaling_onerow}, we show that when we scale down the top intruder dimension of each weight matrix, we measure a significant reduction in forgetting (pre-training loss) while incurring a minimal drop in test accuracy. 
For all examples in Fig.~\ref{fig:intruder_scaling_onerow}, we observe that when using $\lambda=0.7$ or $0.9$, there is almost no impact on fine-tuning performance, while there is a large percentage drop in forgetting (See Tables \ref{table:scaling_intruder_roberta}\&\ref{table:scaling_intruder_llama}).
In one example for LLaMA2-7B fine-tuned on MetaMath with LoRA $r=256$, we observe that scaling the top intruder dimension in each matrix with $\lambda=0.3$ leads to a 0.1\% drop in test accuracy and a 33.3\% drop in the forgetting induced by fine-tuning. In another for RoBERTa-base fine-tuned on QQP, using $\lambda=0.7$ leads to equivalent in test accuracy and a 33.2\% reduction in the forgetting induced by fine-tuning.
In certain scenarios, we even see test accuracy \emph{improve} along with a drop in forgetting.
If we instead increase their contribution ($\lambda > 1$), we observe more forgetting.
Across the board, scaling down intruder dimensions seems to have little impact on test accuracy but a major impact on forgetting. 
This pattern is exclusive to intruder dimensions (Fig.~\ref{roberta_scaling}): if we instead intervene on pre-trained singular vectors that have a similar singular value to the intruder dimension (ensuring similar contributions to matrix) and scale their magnitude down, we see that forgetting goes up. 
These results indicate that the forgetting observed in LoRA is caused by intruder dimensions interfering with the pre-trained language modeling capabilities. Moreover, the scale (singular value) of these intruder dimensions is not essential for the fine-tuning task performance. See Appendix \ref{scaling-intuders-appendix-text} for further discussion.

\begin{figure}[t]
  \centering
  \begin{subfigure}[b]{0.38\textwidth}
      \includegraphics[width=\linewidth]{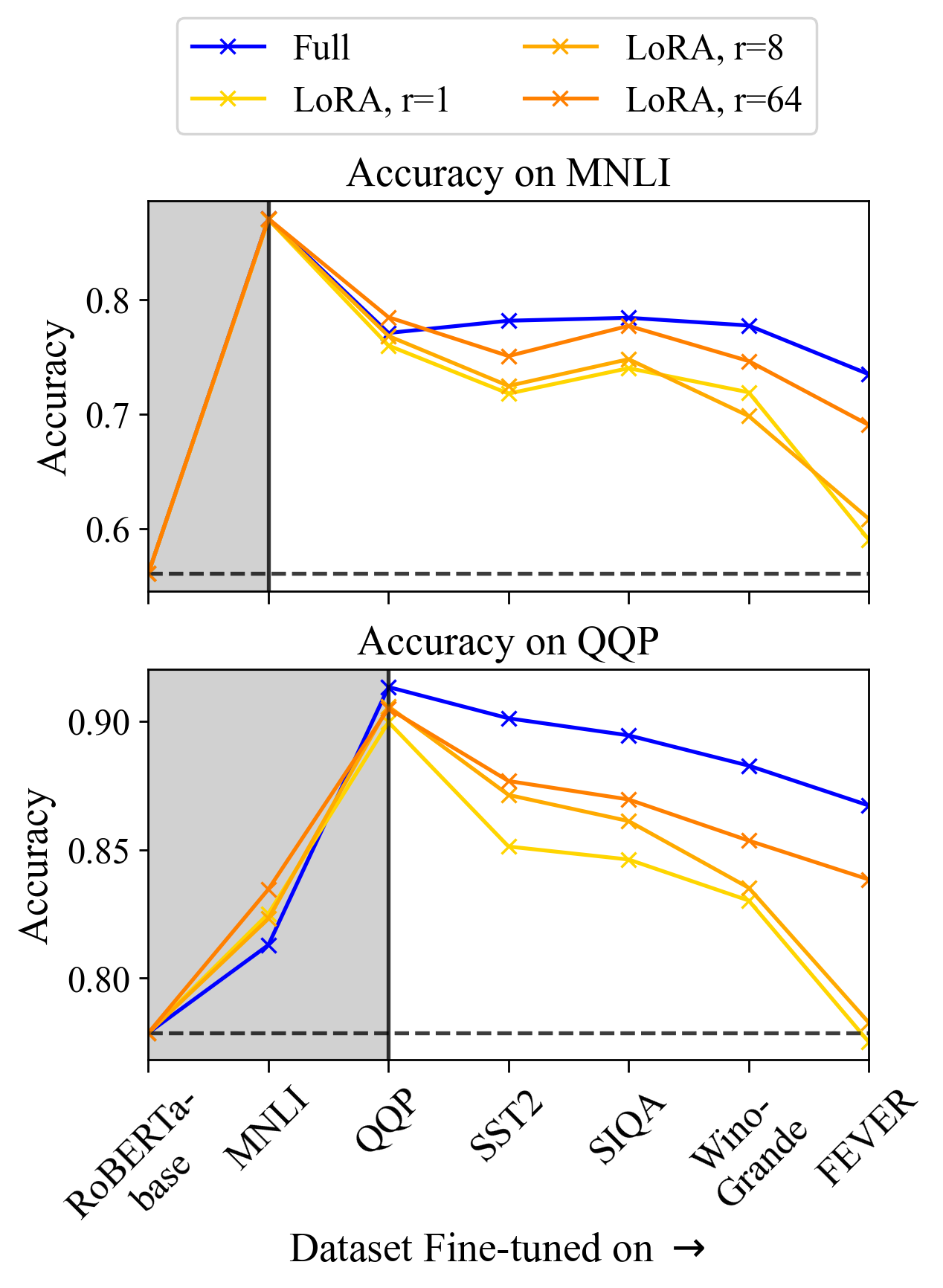}
      \caption{\small Continual learning results.}
      \label{continual-learning-vertical}
  \end{subfigure}
  \hfill
  \begin{minipage}[b]{0.61\textwidth}
      \begin{subfigure}[b]{\linewidth}
          \includegraphics[width=\linewidth]{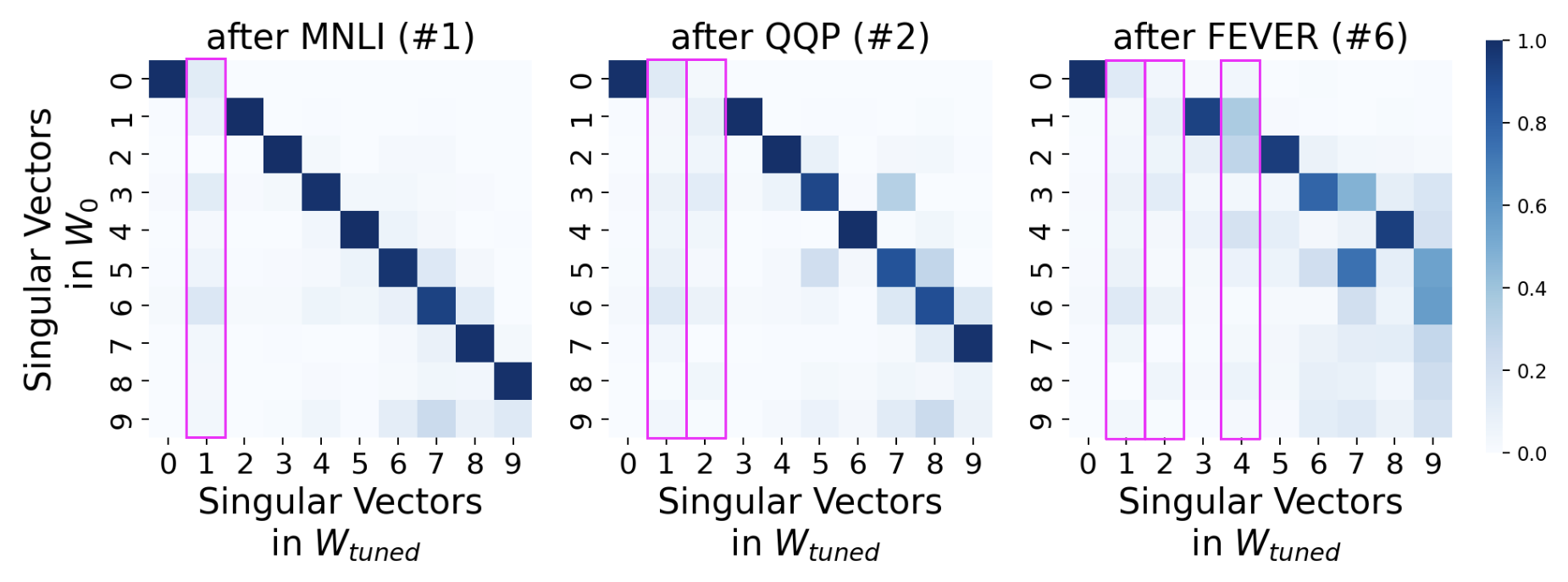}
          \caption{\small Similarity matrices for LoRA r=8 during continual learning.}
          \label{continual-learning-heat_lora_3}
      \end{subfigure}\par\medskip
      \begin{subfigure}[b]{\linewidth}
          \includegraphics[width=\linewidth]{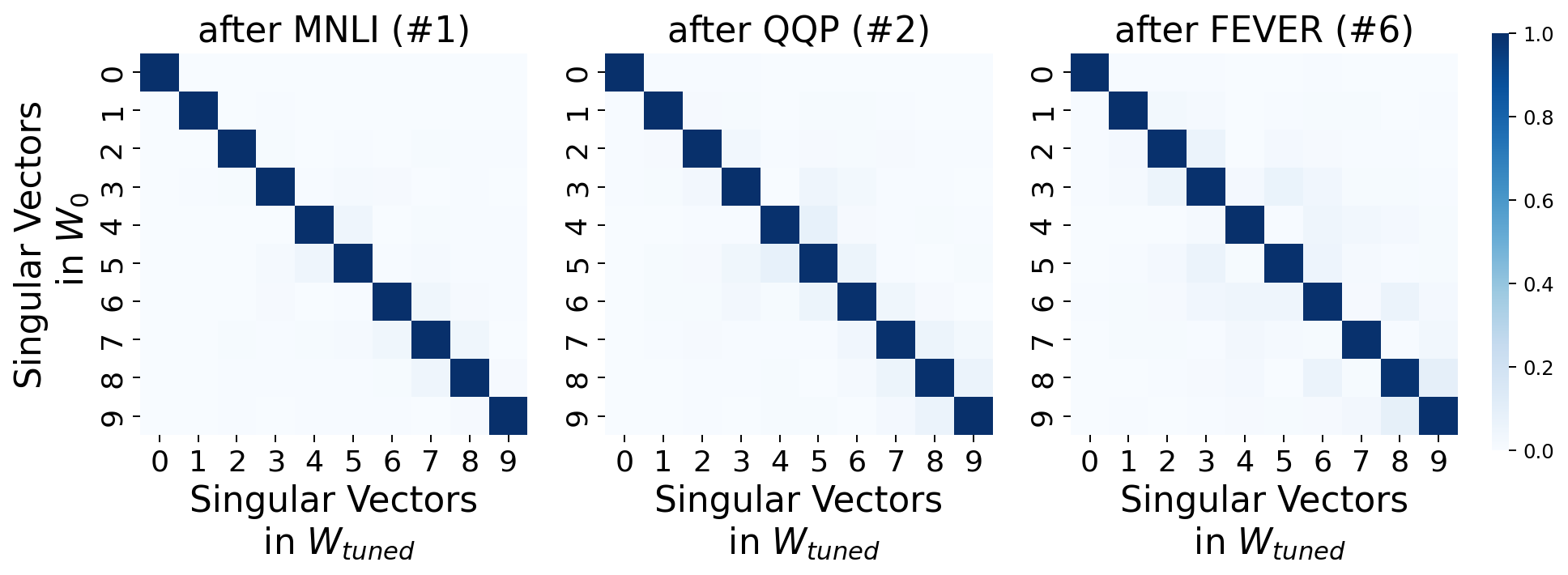}
          \caption{\small Similarity matrices for full fine-tuning during continual learning.}
          \label{continual-learning-heat_full_3}
      \end{subfigure}
  \end{minipage}

  \caption{\small
  \textbf{Full fine-tuning is better than LoRA at continual learning because of accumulating intruder dimensions.}
  When sequentially training on six tasks, full fine-tuning retains performance better than LoRA. in Fig.~\ref{continual-learning-vertical}, horizontal dotted line indicates baseline pre-trained performance. Vertical solid line indicates when a specific dataset is fine-tuned on. Gray region represents performance before the model has been trained on that task. See Appendix \ref{continual-learning-full-text-appendix}, Fig.~\ref{continual_learning_full} for more. In Figs.~\ref{continual-learning-heat_lora_3}\&\ref{continual-learning-heat_full_3}, we see that LoRA accumulates intruder dimensions across tasks and contributes to its degrading performance, whereas full fine-tuning does not.
  }
  \label{continual-learning-plots}
\end{figure}

 \textbf{Continual Learning Setup.} To examine a practical example of how accumulating intruder dimensions, which cause forgetting, may impact performance, we study continual learning, since it requires learning and remembering across a range of tasks. To do this, we train RoBERTa sequentially on multiple tasks and measure performance as new tasks are learned. We use the same training recipe and datasets as before but now sequentially in the following dataset order: MNLI, QQP, SST-2, SIQA, WinoGrande, FEVER. After training on a certain dataset in the sequence, we merge the LoRA weights into the model and reinitialize the LoRA adapter before training on the next task. After training on a specific task, we test on all tasks by, for each task, separately retraining its classification head before testing on its test set. Results are shown in Fig.~\ref{continual-learning-vertical}. 
 
\textbf{Accumulating intruder dimensions hurts LoRA models during continual learning.}
In Fig.~\ref{continual-learning-vertical}, initially both LoRA and full fine-tuning train to equal performance (MNLI), which is consistent with our previous observations. However, we observe that all ranks of LoRA degrade much more rapidly than full fine-tuning.
Low ranks of LoRA, which have the most intruder dimensions, degrade the most.
We attribute this divergence from earlier results—where LoRA appeared to forget less—to the accumulation of intruder dimensions during continual learning, which drive forgetting. 
To show this, we visualize how intruders are added across tasks in Fig.~\ref{continual-learning-heat_lora_3}. Here, we see that each task adds its own intruder dimensions leading to a large amount of intruder dimensions upon the completion of the six task continual learning experiment. In contrast, in Fig.~\ref{continual-learning-heat_full_3}, we see that full fine-tuning retains the pre-trained structure well justifying why full fine-tuning forgets less during continual learning.

\textbf{Implications and prescriptions in fine-tuning.}
These findings suggest several implications and prescriptions during LoRA fine-tuning. We have shown that intruder dimensions drive forgetting, and therefore should be avoided when possible. Interestingly, this presents a data free model evaluation method to examine which model is most overfit to the fine-tuning task (forgotten the most): given two equally performing models on downstream test sets, you should select the one with fewer intruder dimensions. Intruder dimensions appear to be a necessary part of fine-tuning, but they can be mitigated.
Further, these results show the danger of using LoRA during continual learning and justifies using many different adapters without combining them, like advocated in \citet{sheng2024sloraservingthousandsconcurrent}.

\section{Conclusion}

This paper describes the finding that LoRA and full fine-tuning update different parts of the parameter space resulting in distinct spectral properties: LoRA often introduces intruder dimensions—high-ranking singular vectors dissimilar to those in pre-trained weights. These structural differences persist across a series of ablations. Next, we 
find that models with fewer intruder dimensions exhibit better out-of-distribution generalization and forget less of the pretraining distribution. Last, we show that intruder dimensions \emph{cause} increased forgetting: We show that reducing the magnitude of high ranking intruder dimensions leads to minimal changes in test performance but a large drop in pretraining loss. We show that this is particularly relevant during continual training: even though LoRA forgets less than full fine-tuning after training on one task, sequentially training leads to an accumulation of intruder dimensions that causes more forgetting than full-finetuning.

\section*{Acknowledgements}

We would like to thank Jacob Portes and Dan Biderman for corresponding with us and releasing their LLaMA-2 7B checkpoints for us to use. This enabled us to study a more comprehensive range of models. We would also like to thank Leshem Chosen, Lucas Hennigen, Han Guo, Vighnesh Subramaniam, Valerio Pepe, and the entire Language \& Intelligence lab for their helpful feedback on this work. This research was supported in part by the National Science Foundation under grant IIS-2238240.

\bibliographystyle{plainnat}
\bibliography{refs}

\newpage
\appendix
\onecolumn

\section{Why do intruder dimensions exist \& can we alleviate them?}

Here, we discuss possible causes of intruder dimensions. 

\subsection{What do intruder dimensions do?}

\textbf{Conjecture: Intruder dimensions, as high-ranking singular vectors, contribute significantly to the norm and stability of the parameter matrix.} In contrast to pre-trained singular vectors that are learned from large pre-training corpora, LoRA introduces intruder dimensions learned solely from the smaller dataset of the fine-tuning task, which overpower the pre-trained vectors, as seen in the experiments so far. This suggests that these intruder dimensions are very task specific. On the other hand, full fine-tuning, while adapting just as effectively to the fine-tuning task, retains the spectral properties of the pre-trained model effectively. Our experiments that scale down intruder dimensions provide evidence for the claim that intruder dimensions are specialized to the fine-tuning task, since we observe that increasing the norm of an intruder dimension (using $\lambda > 1$) leads to little change in adaptation task performance but leads to a significant increase in forgetting (pre-training loss).

\subsection{Adding a random vector}
\label{adding-random-vector-text}
\textbf{Adding an random vector to a pre-trained matrix introduces an intruder dimension:} To help provide intuition about how new singular vectors in the SVD can be added by LoRA, we examine mathematical conditions that lead to their creation. Certainly, when comparing SVD$(W+\lambda vv^T)$ and SVD$(W)$, where $W$ are the pre-trained weights in $\mathbb{R}^{n \times n}$, $v$ is a randomly sampled vector in $\mathbb{R}^n$, and $\lambda$ is a scalar value greater than the largest singular value of $W$, we expect this update to create an intruder dimension (as $v$ is nearly orthogonal to the existing singular vectors w.h.p.).

\subsection{Differences in update rule}

As described in Appendix \ref{derivation_grads}, LoRA and full fine-tuning have characteristically different update rules, even for the same training examples. We highlight that LoRA has gradients projected into a low-rank space \citep{hao2024floralowrankadapterssecretly}, leading to conditions similar to the toy example in section \ref{adding-random-vector-text} above.

\subsection{Impact of learning rate on intruder dimensions.}
\label{learning-rate-text}
\label{learning_rate_sweep}

 We study the impact learning rate has on intruder dimensions by sweeping a  range of learning rates while keeping all other hyperparameters fixed for LoRA $r=8$ and fine-tune on MNLI. Across training epochs, we report the number of intruder dimensions, test accuracy, and pre-training loss (Fig.~\ref{fig:hyperparam_sweep}). Across models, we see they have similar test accuracies after 5 epochs but very different numbers of intruder dimensions. We note that small learning rates do not converge as fast as the ones we tested and have difficulty reaching the maximum performance that larger learning rates are able to reach. We see that as we increase learning rate, the number of intruder dimensions also increases (left, Fig.~\ref{fig:hyperparam_sweep}). This illustrates a tradeoff in the selection of learning rate of LoRA: picking a larger learning rate may lead to faster convergence and potentially better test accuracy with more intruder dimensions and drift in overall language modeling performance, while smaller learning rates may lead to less drift but potentially lower test accuracy. 

 Because of this experiment, one may dispute our findings with the claim that they are due to our specific selection of hyperparameters. Therefore, we find it important to note that we adopt hyperparameters from prior literature \citep{lora}, default settings in common machine learning libraries\footnote{PEFT, the most popular LoRA library, use learning rates $\geq \mathrm{1e\text{-}3}$ in their tutorials and states ``With LoRA-like methods, you can afford to use a higher batch size and learning rate." \citep{peft}.}, and also study externally trained open-sourced models. This means that our findings are a reflection of common practices and not due to a selection bias by us. Learning rates used by prior work were likely determined based on a variety of factors like speed of convergence and best resulting test accuracy and therefore selected large learning rates that still converge well but coincidentally result in intruder dimensions.

\subsection{Matrix product parameterization of LoRA}

\begin{figure}
    \centering
    \includegraphics[width=1.0\linewidth]{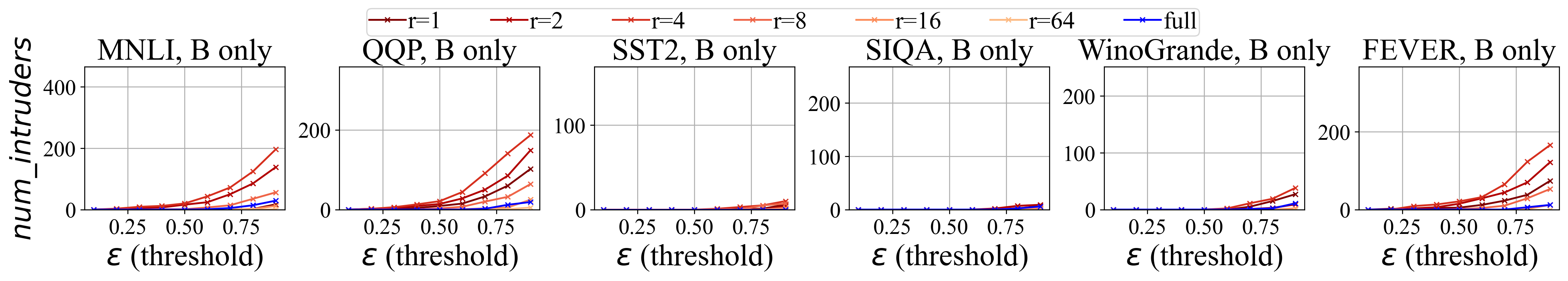}
    \caption{\textbf{Impact of only tuning B on the number of intruder dimensions.} We randomly initialize A such that it has singular values of 1, freeze it, and only train B. When we do this, we see a sharp reduction in high ranking intruder dimensions in comparison to those in normal LoRA (reported in Fig.~\ref{roberta-epsilon}). Graphs for a specific dataset have the same range as Fig.~\ref{roberta-epsilon} for easy comparison.}
    \label{fig:freeze_A_a2r}
\end{figure}

\label{tuning_b_text}
Multiplying matrices together amplifies their spectral differences (their singular values) and in most cases leads to a lower effective rank. To test the impact of the product $BA$ on the introduction of intruder dimensions, we randomly initialize $A$ such that all its singular values are 1 and freeze it. We only tune $B$ and keep the rest of our fine-tuning recipe the same. Comparing this with vanilla LoRA is fair because \citet{zhu2024asymmetrylowrankadaptersfoundation} found that tuning $B$ is more impactful and important for generalization in comparison to $A$ and \citet{hao2024floralowrankadapterssecretly} showed that only tuning $B$ effectively approximates LoRA.
As we can see in Fig.~\ref{fig:freeze_A_a2r}, we see a sharp drop in the number of high ranking intruder dimensions when only tuning $B$ in comparison to the vanilla LoRA case where we train $A$ and $B$ separately, as reported in Fig.~\ref{epsilon-sweep}. This suggests that the matrix product of LoRA is an important component in the introduction of intruder dimensions because of how it amplifies the spectral differences of $B$ and $A$.

\section{Implementation Details}
\label{implementation-details}
\subsection{Evaluation details}

We follow the precedence of \citet{biderman2024loralearnsless} when evaluating LLaMA2-7B: 
When fine-tuned on code, we evaluate on HumanEval \citep{for_humaneval} using the bigcode-eval-harness \citep{bigcode-evaluation-harness}. 
When fine-tuned on math, we evaluate on GSM8K \citep{gsm8k} using the lm-eval-harness \citep{eval-harness}.
On both, we evaluate task forgetting by evaluating on Hellaswag, WinoGrande, and Arc-Challenge using the lm-eval-harness \citep{eval-harness}.

We measure language modeling loss for all our LLaMA2-7B models on a random sample of its pre-training data distribution, according to \citet{llama2}.
We measure ``pseudo-loss'' for all our fine-tuned RoBERTa models on a random sample of the four datasets that RoBERTa used for pre-training(OpenWebText \citep{OpenWebText}, CCNews \citep{CC_News}, Stories \citep{Stories_dataset}, and bookcorpus \citep{bookcorpus}) and weigh them proportionally to their contribution as described by \citet{liu2019robertarobustlyoptimizedbert}.

\subsection{Compute Resources}
All experiments were run on an internal, shared 8xA100-SXM4-80GB machine. All RoBERTa-base fine-tuning runs required a single A100 GPU. All evaluations and analyses also required a single A100 GPU. Many experiments were run sequentially due to need to share these computing resources.
Due to these constraints, instead of fine-tuning our own LLaMA2-7B models, we use publicly released fine-tuned models. For more information on these models, see Section \ref{huggingface-details}.
Each RoBERTa-base fine-tune run takes at most 6 hours on a single GPU. Evaluating an arbitrary LLaMA2-7B model for both test accuracy and forgetting takes about 45 minutes on a single GPU.

\subsection{RoBERTa fine-tuning details}
\label{section:roberta-details}

We generally follow the procedure used by \citet{lora}. For all models, we use a linear learning rate schedule with 0.06 linear warmup ratio and train for a maximum of 5 epochs with batch size 16. We use the Adam optimizer \citep{kingma2017adammethodstochasticoptimization} with no weight decay and a maximum sequence length of 512. We fine-tune all linear layers besides the embedding matrix. For full fine-tuning, we use a learning rate of 1e-5. For LoRA, we set $\alpha=2r$, and train for all ranks in \{1, 2, 4, 8, 16, 64\}. We hold the ``total learning rate of LoRA", which is $\alpha*\eta$, fixed as we sweep rank such that this product always equals 2.4e-3. We fine-tune these models to equivalent accuracy on their downstream task. We fine-tune on six sequence classification tasks: sentiment analysis \citep{sst2}, entailment \citep{mnli}, duplicate identification \citep{wang2019gluemultitaskbenchmarkanalysis}, fact verification \citep{fever}, and common sense reasoning \citep{siqa, winogrande}.

\section{Model Accuracies}

\label{model-acc-text-appendix}

We report the accuracies that our RoBERTa models achieve in Table \ref{model_accuracies} and Table \ref{a2r_model_accuracies}. Our main results are based on the models in Table \ref{a2r_model_accuracies}.

\begin{table}[h!]
\centering
\begin{tabular}{cc|cccccc} \toprule
 Model & Type & MNLI & SST-2 & QQP & WinoGrande & SIQA & FEVER \\
\hline
\multirow{7}{*}{RoBERTa-base} & Full & 0.8745 & 0.9438 & 0.9152 & 0.6582 & 0.6499 & 0.6892\\  
 & r=1 & 0.8647 & 0.9358 & 0.9045  & 0.6251 & 0.672 & 0.6712 \\ 
 & r=2 & 0.8604 & 0.9415 & 0.9058  & 0.6172 & 0.6581 & 0.6673 \\
 & r=4 & 0.8607 & 0.9369 & 0.9079   & 0.6472 & 0.6505 & 0.6694 \\
 & r=8 & 0.8648 & 0.9438 & 0.9108  & 0.6417 & 0.6586 & 0.6582 \\ 
 & r=16 & 0.8604 & 0.9427 & 0.9095  & 0.6235 & 0.6853 & 0.663 \\
 & r=64 & 0.8671 & 0.9484 & 0.9117  & 0.6614 & 0.6638 & 0.6601 \\ 
 \hline 
\end{tabular}
\caption{Model accuracies on their given downstream task after fine-tuning for $\alpha = 8$.}
\label{model_accuracies}
\end{table}

\begin{table}[h!]
\centering
\begin{tabular}{cc|cccccc} \toprule
 Model & Type & MNLI & SST-2 & QQP & WinoGrande & SIQA & FEVER \\
\hline
\multirow{7}{*}{RoBERTa-base} & Full & 0.8745 & 0.9438 & 0.9152 & 0.6582 & 0.6499 & 0.6892\\
 & r=1 & 0.8677 & 0.9415 & 0.9042 & 0.6275 & 0.6418 & 0.687\\
 & r=2 & 0.869 & 0.945 & 0.9054 & 0.6551 & 0.6438 & 0.6822\\
 & r=4 & 0.8698 & 0.9472 & 0.9089 & 0.6361 & 0.6602 & 0.6827\\
 & r=8 & 0.8704 & 0.9472 & 0.9093 & 0.6346 & 0.6607 & 0.6928\\
 & r=16 & 0.8739 & 0.9461 & 0.9093 & 0.6417 & 0.6571 & 0.6924\\
 & r=64 & 0.8719 & 0.9472 & 0.9061 & 0.6212 & 0.6167 & 0.6864\\
 \hline 
\end{tabular}
\caption{Model accuracies on their given downstream task after fine-tuning for $\alpha = 2r$. Our main results are based on these models.}
\label{a2r_model_accuracies}
\end{table}

\section{Cosine Similarity with Orthogonal Vectors that Span a Space}
\label{section:orthogonal-cosine-appendix}

Here we demonstrate why it is possible for a vector to have low cosine similarity with every orthogonal vector that collectively span a space if the dimensionality of the vectors is high.

 \paragraph{Minimizing the Maximum Cosine Similarity.} Lets take $Z = \min\limits_{v \in \mathbb{R}^n}\max\limits_i cos(v, x_i)$, where $v$ is an arbitrary vector and each vector $x_i$, which we collectively call $X$, make up an orthonormal basis that span the space. $Z$ can be small in a high dimensional space.

 \paragraph{2-D case.} Assume $X=I$ without loss of generality. It is trivial to see that $Z=\frac{1}{\sqrt{2}}$, and is when $v=\begin{bmatrix}\frac{1}{\sqrt{2}}&  \frac{1}{\sqrt{2}}\end{bmatrix}$.

 \paragraph{3-D case.} Assume $X=I$ without loss of generality. $Z=\frac{1}{\sqrt{3}}$ when $v=\begin{bmatrix}\frac{1}{\sqrt{3}}& \frac{1}{\sqrt{3}} & \frac{1}{\sqrt{3}}\end{bmatrix}$.

 \paragraph{N-D case.} In the N-D case, we can see, via induction, that $Z=\frac{1}{\sqrt{n}}$. 
 
As we can see here, if $n$ is large, the value of $Z$ will be low, even though we are doing the cosine similarity of a vector with respect to a set of orthonormal vectors that span a space.

\section{Derivation of LoRA Adapter's Gradients}

Our calculations were derived independently but follow a similar line to that of \citet{hao2024floralowrankadapterssecretly}.

\label{derivation_grads}

\paragraph{Derivation for Full Fine-tuning.}
Full fine-tuning is structured such that $$Y=W_{tuned}X=(W_0+\Delta W)X,$$ where $X \in \mathbb{R}^{n \times b}$ are the inputs, $Y \in \mathbb{R}^{m \times b}$ are the outputs, $W_0 \in \mathbb{R}^{m \times n}$ are the pre-trained weights, and $\Delta W \in \mathbb{R}^{m \times n}$ is the fine-tuning update. Accordingly, $\frac{\partial L}{\partial \Delta W} = \frac{\partial L}{\partial Y}X^T$, and the update is $$\Delta W_{n}=\Delta W_{n-1}-\eta\frac{\partial L}{\partial Y}_nX^T_n,$$ where $\eta$ is the learning rate.

\paragraph{Derivation for LoRA.}
LoRA is structured such that $$Y=W_{tuned}X=(W_0+\frac{\alpha}{r}BA)X,$$ where $X \in \mathbb{R}^{n \times b}$ are the inputs, $Y \in \mathbb{R}^{m \times b}$ are the outputs, $W_0 \in \mathbb{R}^{m \times n}$ are the pre-trained weights, $B \in \mathbb{R}^{m \times r}$ is initialized to zero, $A \in \mathbb{R}^{r \times n}$ is randomly initialized, and $\alpha$ is a hyperparameter. Accordingly, $\frac{\partial L}{\partial B} = \frac{\alpha}{r}\frac{\partial L}{\partial Y}X^TA^T$ and $\frac{\partial L}{\partial A} = \frac{\alpha}{r}B^T\frac{\partial L}{\partial Y}X^T$. Therefore, their respective updates are $$B_n=B_{n-1}-\eta \frac{\alpha}{r}\frac{\partial L}{\partial Y}X^TA^T$$ and $$A_n=A_{n-1}- \eta \frac{\alpha}{r}B^T\frac{\partial L}{\partial Y}X^T,$$ where $\eta$ is the learning rate.

\paragraph{Differences in First Step.}

During the very first step of training, given identical examples both full fine-tuning and LoRA have the same $X$ and $Y$ for each layer since $B$ is initialized to zero. After the first step, full fine-tuning has a update matrix equal to $$\Delta W_{full}=-\eta \frac{\partial L}{\partial Y}X^T.$$ In contrast, LoRA has an update matrix equal to $$\Delta W_{lora}= (\frac{\alpha}{r})(B_{0}-\eta \frac{\alpha}{r}\frac{\partial L}{\partial Y}X^TA_0^T)(A_{0}- \eta \frac{\alpha}{r}B_0^T\frac{\partial L}{\partial Y}X^T).$$ Since $B_0=0$,

$$\Delta W_{lora}= (\frac{\alpha}{r})(-\eta \frac{\alpha}{r}\frac{\partial L}{\partial Y}X^TA_0^T)(A_{0}).$$

From this, we can see that the gradient steps are clearly different, even with the same training examples.

\section{Intruder Dimensions Correlate with Forgetting}
\label{intruder-correlation-text-appendix}

\subsection{For RoBERTa}

As mentioned in the main text, when measuring the Spearman correlation between the number of intruder dimensions and forgetting in Fig.~\ref{fig:hyperparam_sweep} we find an extremely strong fit, with $\rho=0.971$ and p-value $\ll 0.001$. This shows us that intruder dimensions strongly correlate with forgetting. In contrast, when we correlate intruder dimensions with performance, we find no such correlation: for RoBERTa, we measure $\rho = -0.3381$ with p-value $= 0.218$.

\subsection{LLaMA2-7B}

\begin{figure}
        \centering
        \includegraphics[width=0.9\linewidth]{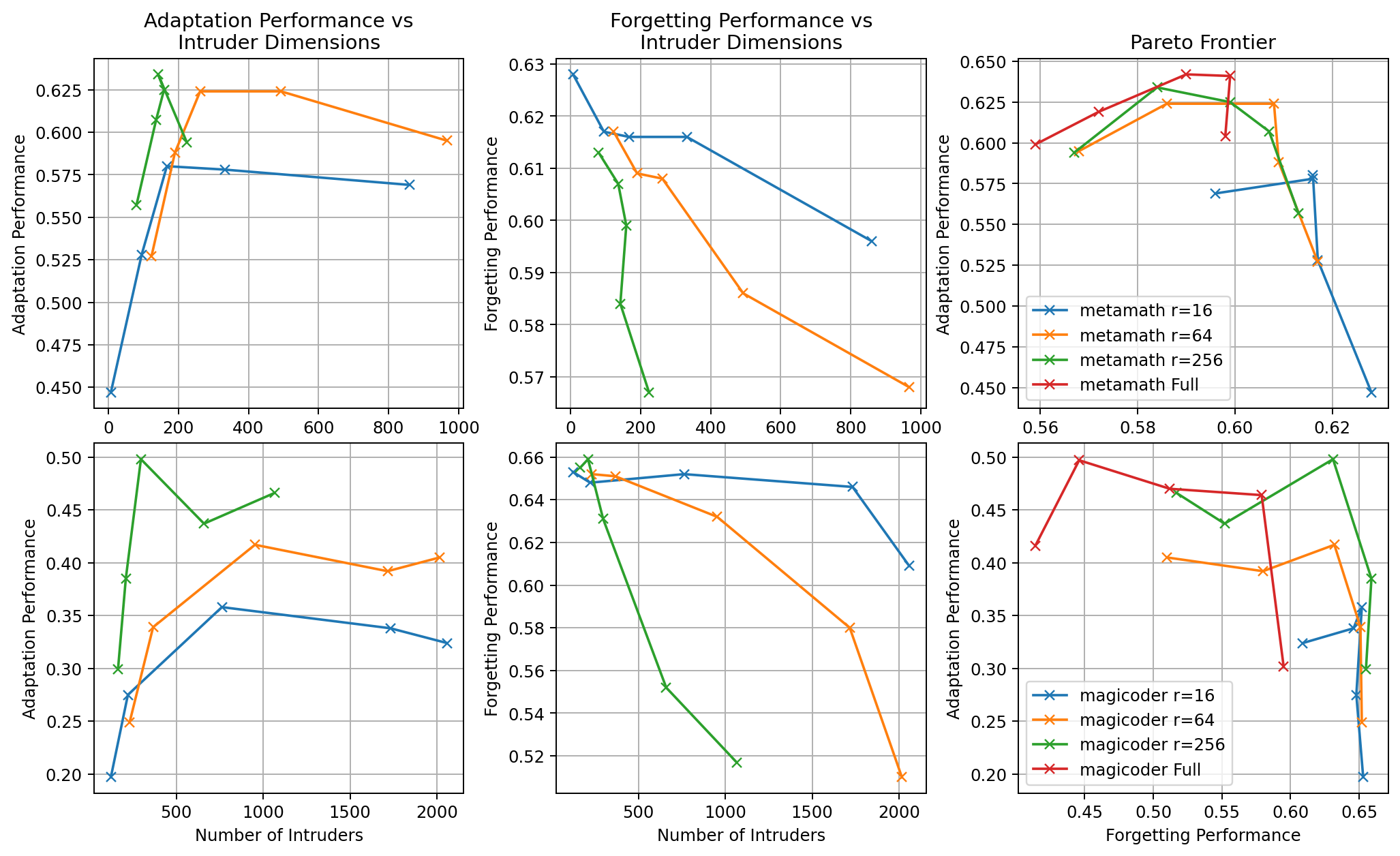}
        \caption{
        \textbf{For LLaMA2-7B, intruder dimensions correlate with forgetting.} Top row: MetaMath. Bottom row: Magicoder. We display intruder dimensions vs test accuracy and intruder dimensions vs forgetting.
        }
        \label{biderman-intruder-correlation}
    \end{figure}

When measuring the Spearman correlation between number of intruder dimensions and forgetting for our LLaMA2-7B models across training epochs (Fig.~\ref{biderman-intruder-correlation} (\emph{middle})), we find a statistically significant relationship with $\rho=0.59$ and p-value $= 0.0006$. When correlating intruder dimensions and test accuracy (Fig.~\ref{biderman-intruder-correlation} (\emph{left})), we instead measure $\rho= -0.3178$ with p-value $= 0.0869$. Again, we see that intruder dimensions correlates with forgetting.

\section{Intruder Dimensions Cause Forgetting (Scaling Experiments)}
\label{scaling-intuders-appendix-text}
\subsection{Performance Differences When Scaling Down Intruder Dimensions}

\subsubsection{RoBERTa}
We report our findings for RoBERTa models fine-tuned on MNLI, QQP, and FEVER in Table \ref{table:scaling_intruder_roberta}. Remember that we scale down using the equation $W = W_0 + \Delta W + (\lambda-1)u_i\sigma_iv^T_i$. Here, we see that scaling down intruder dimensions leads to a sharp drop in forgetting (pre-training loss) and a much smaller drop in test accuracy. Scaling down an intruder dimension by two ($\lambda=0.5$) results always leads to less than a two percent drop in test accuracy but double digit percentage drops in forgetting. One particularly compelling example, as shown in Fig.~\ref{intruder_scaling_qqp_annotated}, shows how scaling down the top intruder dimensions with $\lambda=0.7$ when fine-tuning on QQP and using LoRA r=8 result in essentially no (0.0\%) drop in adaptation performance but a large (-33.2\%) drop in forgetting. Our findings of the impact of scaling down intruder dimensions hold across three datasets and both LoRA r=1 and r=8, which were the two ranks that we found to have many intruder dimensions. Note that this experiment is meaningless if a model has no intruder dimensions, since no singular vectors will be removed.

\begin{table}[h!]
    \centering
    \begin{tabular}{|cc|cc|cc|cc|cc|cc|}
    \hline
\multirow{2}{*}{Task} & LoRA & \multicolumn{2}{|c|}{$\lambda=0.1$} & \multicolumn{2}{|c|}{$\lambda=0.3$} & \multicolumn{2}{|c|}{$\lambda=0.5$} & \multicolumn{2}{|c|}{$\lambda=0.7$} & \multicolumn{2}{|c|}{$\lambda=0.9$}\\ 
& Rank & TA & PTL & TA & PTL & TA & PTL & TA & PTL & TA & PTL \\ \hline
\multirow{2}{*}{MNLI} & r=1 & -18.7 & -13.3 & -8.3 & -14.5 & -2.7 & -13.7 & -0.6 & -11.0 & 0.0 & -5.1 \\ & r=8 & -5.1 & -24.7 & -1.9 & -23.1 & -0.6 & -19.8 & -0.2 & -14.5 & 0.0 & -5.9 \\ \hline 
\multirow{2}{*}{QQP} & r=1 & -8.6 & -35.5 & -4.4 & -35.8 & -1.6 & -34.1 & -0.4 & -28.9 & 0.1 & -14.6 \\ & r=8 & -3.3 & -52.0 & -1.6 & -50.6 & -0.6 & -45.0 & -0.0 & -33.2 & 0.1 & -13.0 \\ \hline 
\multirow{2}{*}{FEVER} & r=1 & -11.1 & -10.3 & -4.2 & -11.8 & -0.4 & -11.1 & 0.6 & -8.7 & 0.6 & -3.8 \\ & r=8 & -5.6 & -14.6 & -1.0 & -15.3 & 0.7 & -13.4 & 1.3 & -9.7 & 0.6 & -4.0 \\ \hline 
    \end{tabular}
    \caption{
    \textbf{Impact of scaling RoBERTa-base's intruder dimensions on test accuracy (TA) and pre training loss (PTL).} Numbers reported are the percent change in test accuracy and percent reduction in forgetting induced by fine-tuning.
    \textbf{Scaling down intruder dimensions leads to less forgetting.} On RoBERTa. PTPL is Pre-training loss and TA is test accuracy. Both are reported as percent change with respect to the unchanged fine-tuned model. Scaling down intruder dimensions has large impact on forgetting but little impact on test accuracy. 
    }
    \label{table:scaling_intruder_roberta}
\end{table}

\subsubsection{LLaMA2-7B}

Our finding that scaling down intruder dimensions leads to less forgetting but similar test accuracy holds to LLaMA2-7B. One particularly interesting example is for LLaMA2-7B fine-tuned on MetaMath with $r=256$: when scaling the top intruder dimensions down with $\lambda=0.5$, we see a large drop (-25.2\%) in forgetting and an \emph{increase} (+1.8\%) in test accuracy.

\begin{table}[h!]
    \centering
    \begin{tabular}{|cc|cc|cc|cc|cc|}
    \hline
\multirow{2}{*}{Dataset} & LoRA  & \multicolumn{2}{|c|}{$\lambda=0.3$} & \multicolumn{2}{|c|}{$\lambda=0.5$} & \multicolumn{2}{|c|}{$\lambda=0.7$} & \multicolumn{2}{|c|}{$\lambda=0.9$}\\ 
& Rank & TA & PTL & TA & PTL & TA & PTL & TA & PTL \\ \hline

\multirow{3}{*}{MetaMath} & r=16 & -15.0 & -46.5 & -6.6 & -40.4 & -2.7 & -28.2 & 0.0 & -10.5 \\ & r=64 & -5.8 & -29.1 & -4.4 & -22.3 & 0.3 & -14.2 & 0.3 & -4.9 \\ & r=256 & -0.1 & -33.3 & 1.8 & -25.2 & 0.5 & -15.5 & -0.8 & -5.2 \\ \hline 
\multirow{3}{*}{Magicoder} & r=16 & -1.6 & -37.5 & -0.6 & -24.4 & -0.1 & -13.2 & 0.1 & -4.1 \\ & r=64 & -5.0 & -12.7 & -3.3 & -8.3 & 0.3 & -4.3 & 2.2 & -1.4 \\ & r=256 & -4.3 & -12.8 & -1.3 & -10.2 & -0.9 & -7.0 & -1.6 & -2.8 \\ \hline 

    \end{tabular}
    \caption{
    \textbf{Impact of scaling LLaMA2-7B's intruder dimensions on test accuracy (TA) and pre training loss (PTL).} Numbers reported are the percent change in test accuracy and percent reduction in forgetting induced by fine-tuning.
    }
    \label{table:scaling_intruder_llama}
\end{table}

\subsection{Intruder Dimensions Cause Worse OOD Performance}

As we discussed in section \ref{forgetting-text-main} of the main text, for LoRA models with main intruder dimensions ($r=1$ and $r=8$ in our experiments) we measure the impact of intruder dimensions by identifying and scaling the top intruder dimension in every weight matrix such that $W = W_0 + \Delta W + (\lambda-1)u_i\sigma_iv^T_i$, where $i$ is the index of the top intruder dimension (note that $\lambda=0$ is removal, $\lambda=1$ is no change, and $\lambda = 2$ doubles the intruder dimension). For our RoBERTa models fine-tuned on MNLI, QQP, and FEVER, we use $\lambda \in \{0, 0.25, 0.5, 0.75, 1.0, 1.5, 2.0\}$ and for each measure the test accuracy and pre-training loss. For a comparison baseline, we select the neighbor of the intruder dimension to separately scale. 

\begin{figure}
        \centering
        \begin{subfigure}[b]{0.49\linewidth}
            \centering
            \includegraphics[width=1.0\linewidth]{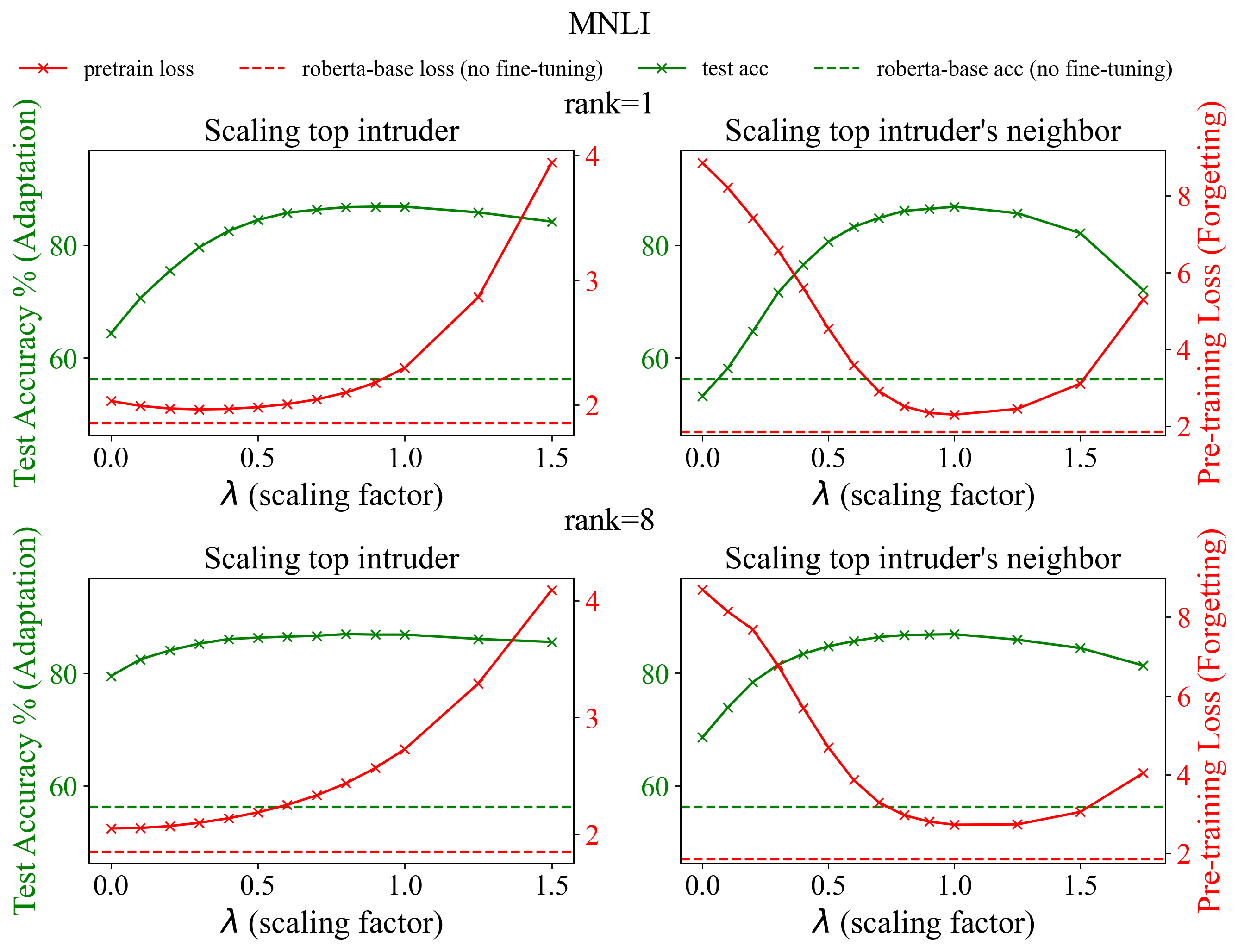}
            \caption{RoBERTa-base fine-tuned on MNLI.}
        \label{fig:intruder_scaling_mnli}
        \end{subfigure}
        \begin{subfigure}[b]{0.49\linewidth}
            \centering
            \includegraphics[width=1.0\linewidth]{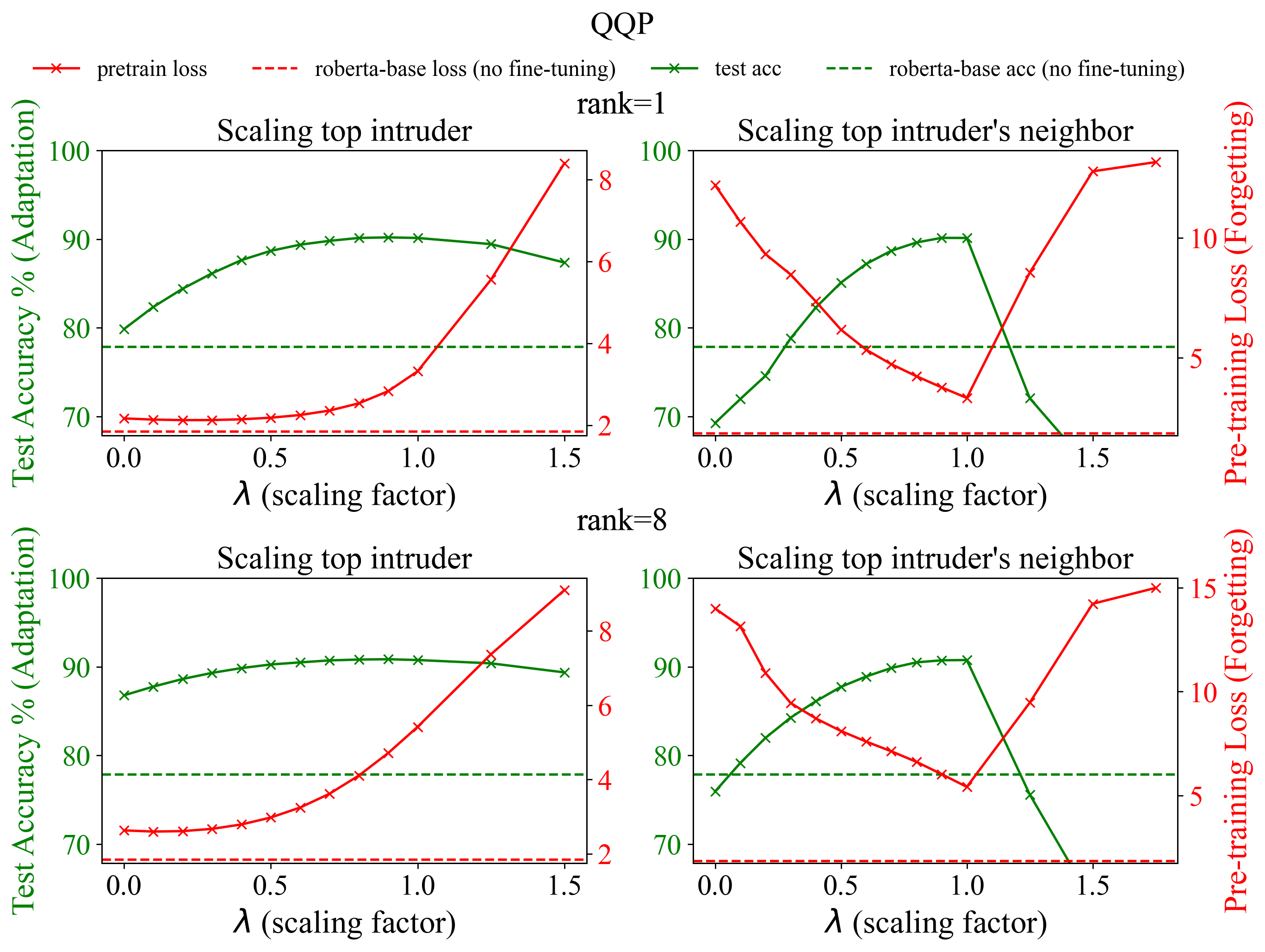}
            \caption{RoBERTa-base fine-tuned on QQP.}
            \label{fig:intruder_scaling_qqp}
        \end{subfigure}

        \begin{subfigure}[b]{0.49\linewidth}
            \centering
            \includegraphics[width=1.0\linewidth]{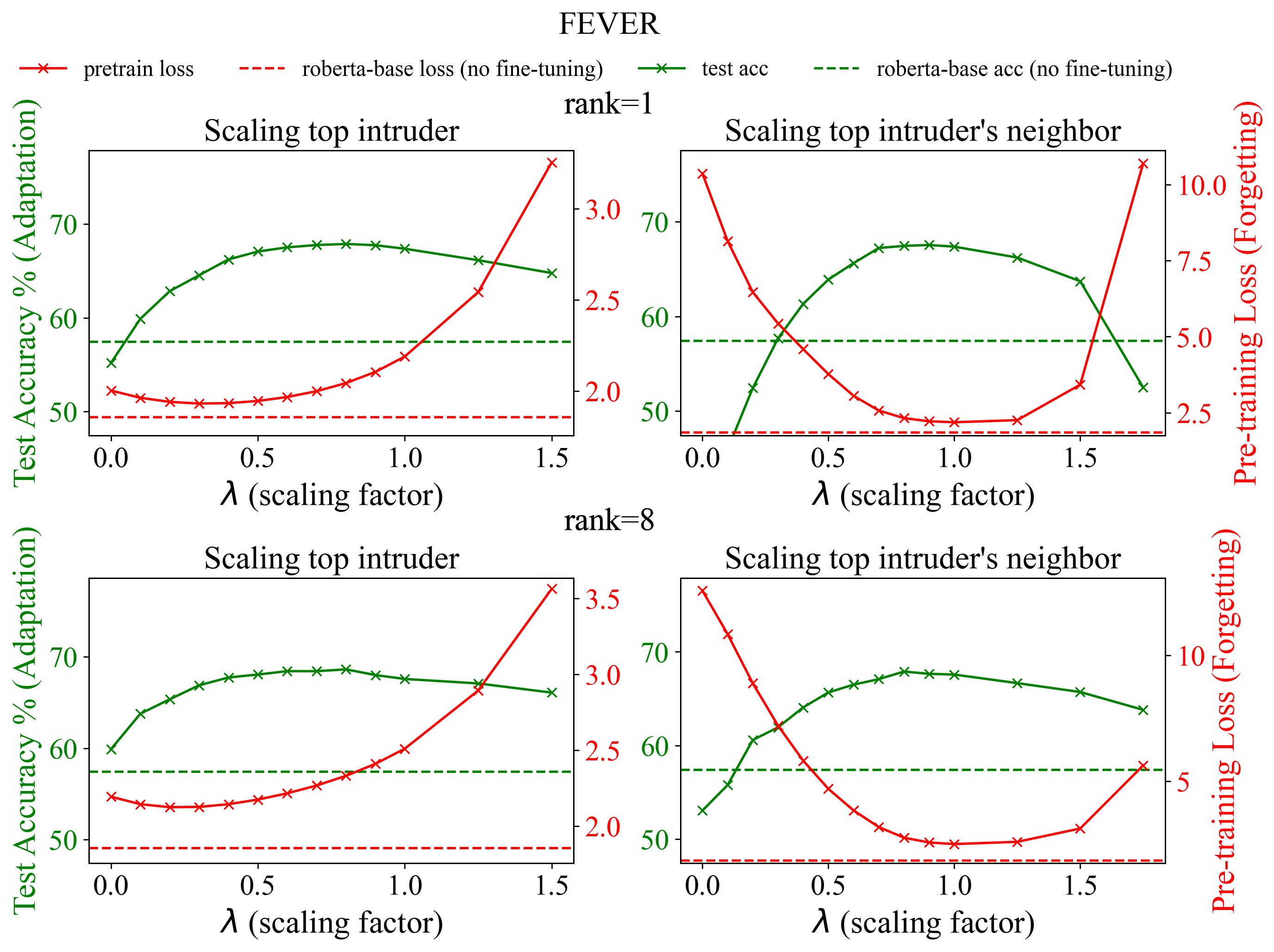}
            \caption{RoBERTa-base fine-tuned on FEVER.}
            \label{fig:intruder_scaling_fever}
        \end{subfigure}
        
        \caption{\textbf{Scaling RoBERTa-base's intruder dimensions. }We scale the top intruder dimension in each matrix by $\lambda$, a multiplicative constant, such that $W = W_0 + \Delta W + (\lambda-1)u_i\sigma_iv^T_i$. Using $\lambda < 1$ leads to a large drop in pre-training loss while only slightly impacting the test accuracy. For Figs.~\ref{fig:intruder_scaling_mnli}, \ref{fig:intruder_scaling_qqp}, and \ref{fig:intruder_scaling_fever}, we also scale the intruder dimension's neighbor, which is a pre-trained singular vector. Changing these vectors negatively impacts both pre-training loss and test accuracy.}
        \label{roberta_scaling}
    \end{figure}

We report these results in Fig.~\ref{fig:intruder_scaling_mnli} (MNLI), Fig.~\ref{fig:intruder_scaling_qqp} (QQP), and Fig.~\ref{fig:intruder_scaling_fever} (FEVER). We find that when scaling down intruder dimensions ($\lambda < 1$), we observe a clear and significant drop in forgetting (pre-training loss) but a negligible drop in test accuracy (adaptation).  When we instead scale up intruder dimensions ($\lambda > 1$), we observe that forgetting increases significantly. We observe that when scaling up top intruder dimensions by 50\% ($\lambda = 1.5$), adaptation performance remains relatively flat with a large increase in forgetting, providing further evidence that intruder dimensions are task specific. These trends hold across all 6 models we study (3 datasets with two different LoRA ranks).

In contrast, when we scale a neighboring (pre-trained) singular vector of intruder dimensions instead, we observe starkly different behaviors. When scaling down ($\lambda < 1$) these pre-trained singular vectors, we observe we see that forgetting sharply increases and adaptation performance drops more sharply than when scaling intruders instead. When scaling up ($\lambda > 1$) these pre-trained singular vectors, we observe similar drops in forgetting and adaptation performance. This is likely because pre-trained singular vectors are well tuned for language modeling, and therefore any change to them will have negative downstream impacts on performance. This shows that are observations are not due to the robustness of the model to scaling down specific singular vectors, but rather the difference in contribution to model performance of intruder dimensions vs. pre-trained singular vectors.

Here, we clarify some possible points of confusion. We only scale the top intruder dimension in each weight matrix, if and only if an intruder exists in that matrix, so even when $\lambda = 0$ we do not recover the pre-trained model because not all weight matrices have intruder dimensions and it is possible for multiple intruder dimensions to exist in a weight matrix if LoRA $r > 1$. Furthermore, intruder dimensions are not perfectly orthogonal to the pre-trained singular vectors. Due to the orthogonality constraint imposed by the SVD, all singular vectors will be changed slightly in the matrix, so that even when we remove the intruder dimension, the resulting matrix will be slightly different. These reasons are why when $\lambda = 0$ our LoRA $r=1$ models do not return to baseline performance.

\begin{figure}
        \centering
        \includegraphics[width=0.8\linewidth]{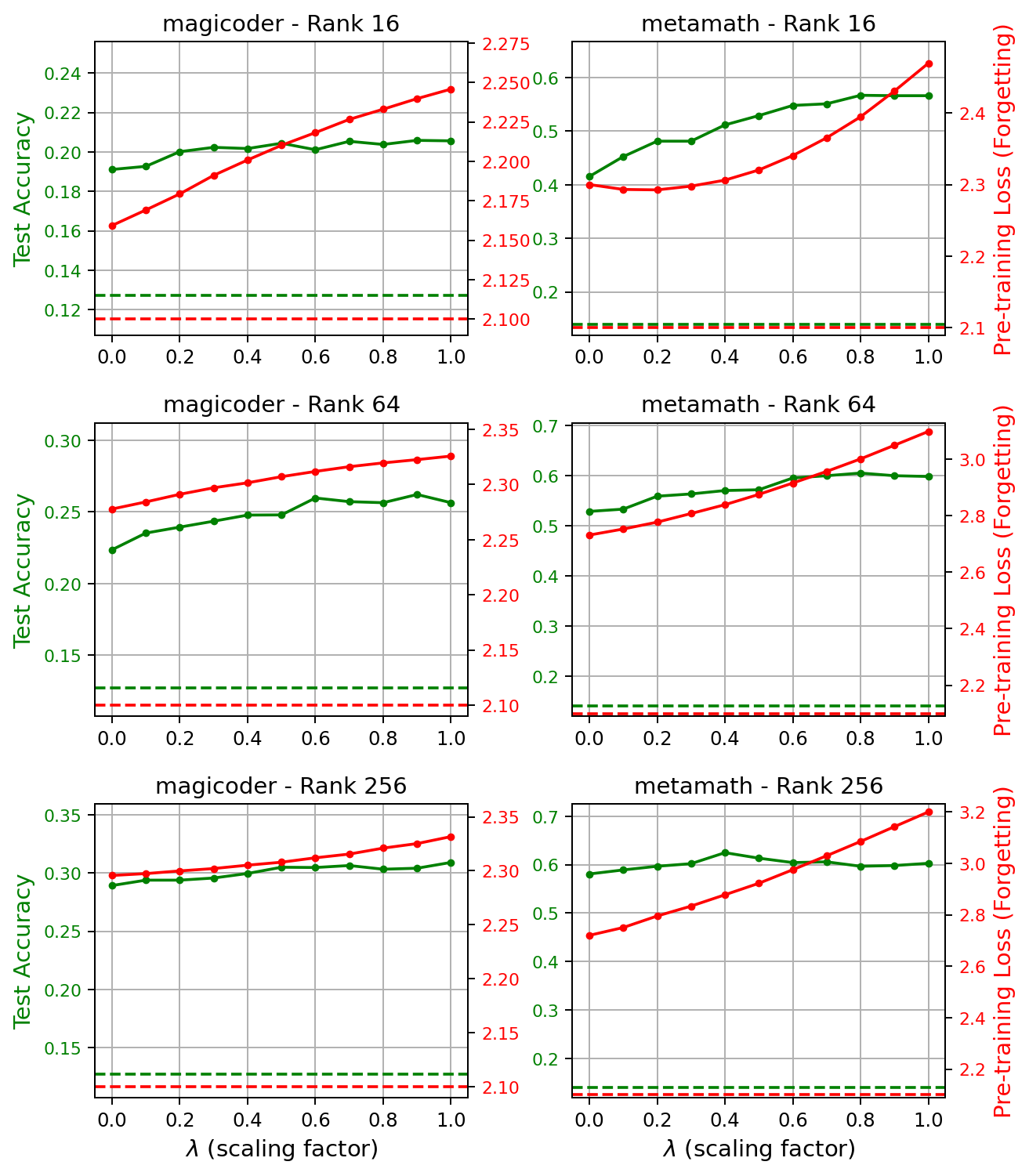}
        \caption{
        \textbf{Scaling down LLaMA2-7B's intruder dimensions leads to less forgetting and nearly equivalent test accuracy.}
        }
        \label{llama-intruder-scaling}
    \end{figure}

These findings hold to LLaMA2-7B: in Fig~\ref{llama-intruder-scaling}, we see that scaling down intruder dimensions leads to much less forgetting but similar test accuracy.

\FloatBarrier
\newpage
\section{The Effective Rank of the Update Matrix Depends on Alpha}
\label{effective-rank-appendix-text}

\begin{figure}[h]
    \centering
    \begin{subfigure}[b]{0.495\linewidth}
        \centering
        \includegraphics[width=0.95\linewidth]{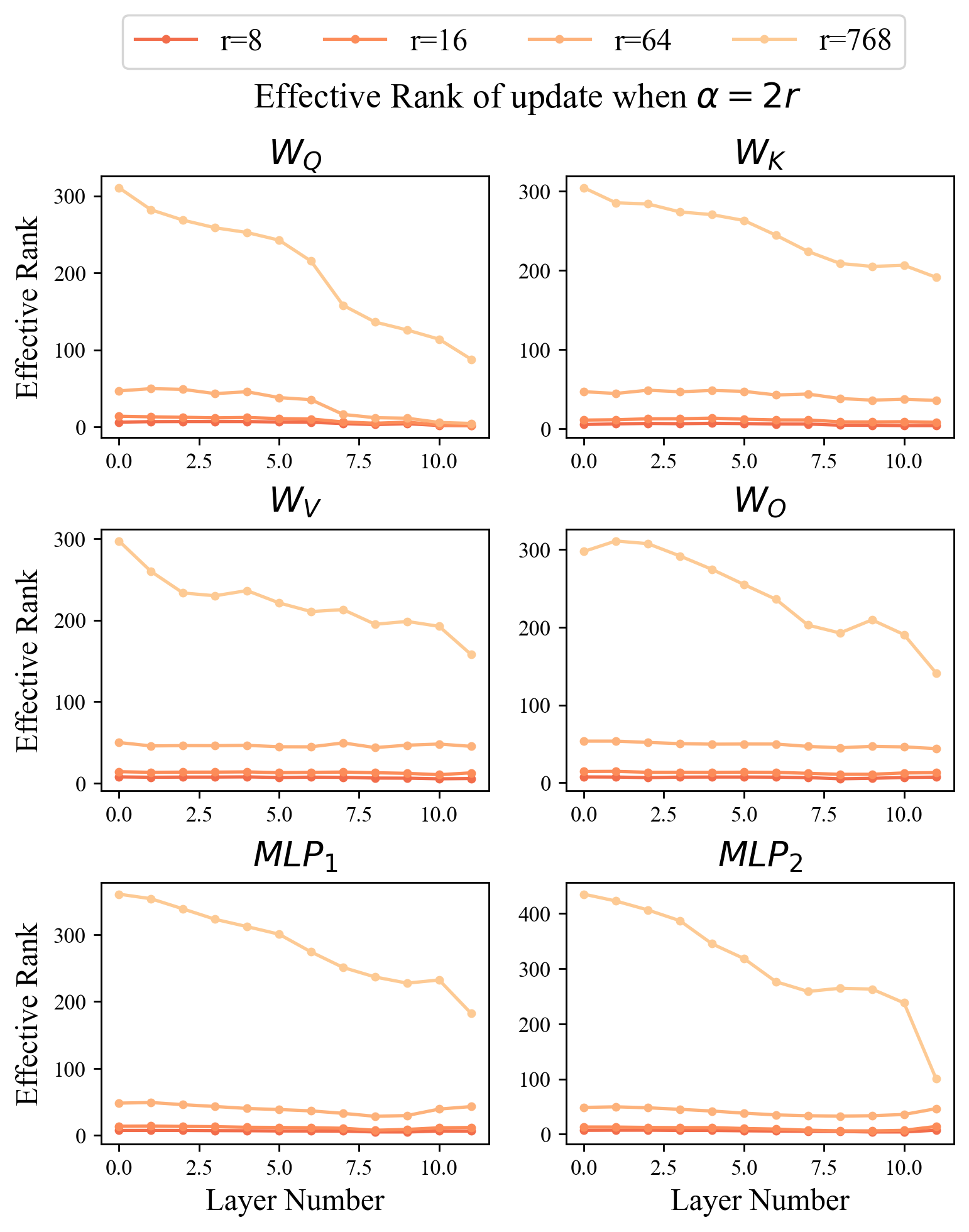}
    \caption{Effective Rank of the LoRA update when $\alpha=2r$.}
    \label{fig:all-effective-ranks-a2r}
    \end{subfigure}
    \begin{subfigure}[b]{0.495\linewidth}
        \centering
        \includegraphics[width=0.95\textwidth]{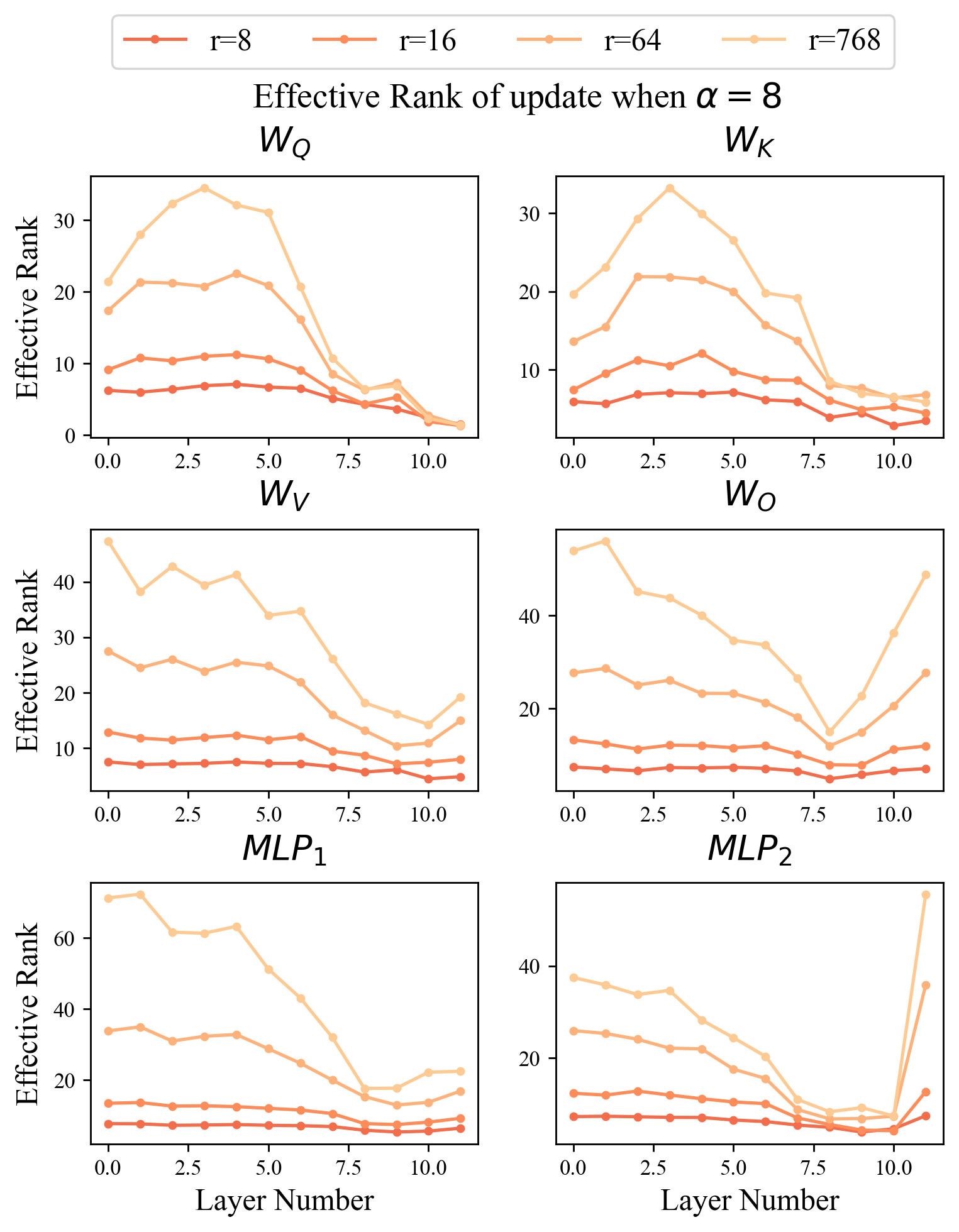}
        \caption{Effective Rank of the LoRA update when $\alpha = 8$.}
        \label{roberta-all-effective-ranks=a8}        
    \end{subfigure}
    \caption{Effective rank of LoRA update matrices ($\Delta W$) for RoBERTa fine-tuned on MNLI. We observe that when $\alpha=2r$, higher ranks of LoRA ($r=64,768$) have much higher effective rank than the same ranks of LoRA but instead with $\alpha = 8$. Building on \citet{kalajdzievski2023rankstabilizationscalingfactor, biderman2024loralearnsless}, this suggests that $\alpha = 2r$ is necessary for high ranks of LoRA to utilize their expressive capacity. Note: full fine-tuning consistently has updates with effective rank above 400.}
    \label{effective-rank-plot}
\end{figure}

\citet{kalajdzievski2023rankstabilizationscalingfactor} found that LoRA can have gradient collapse when rank is high if alpha is not set properly. \citet{biderman2024loralearnsless} found that setting $\alpha = 2r$ is very important for the performance of high rank LoRA. In this section, we provide additional evidence of the importance of setting $\alpha = 2r$ and show that if $\alpha$ is held fixed, high rank LoRA converges to low rank solutions. To do this, for both $\alpha = 2r$ and $\alpha = 8$, we measure the effective rank \citep{effective_rank} of the weight matrix updates for different LoRA ranks. 
Effective rank is a measure of the information density of a matrix and can be thought of as an estimation of the rank needed to capture the information held in the weight matrix. It is computed using the singular values of a matrix and we expect the LoRA rank to be the upper bound on what the effective rank can be: LoRA $r=8$ should have an effective rank update of at most 8.
We present the effective rank measurements in Fig.~\ref{effective-rank-plot}. In this plot, we find that when $\alpha = 2r$ (Fig.~\ref{fig:all-effective-ranks-a2r}), LoRA $r=64$ and $r=768$ have much higher effective rank than $r=8$ and $r16$, with $r=768$ appearing to always have effective rank above 100. In stark contrast, we see that when $\alpha = 8$ (Fig.~\ref{roberta-all-effective-ranks=a8}), high ranks of LoRA have much lower effective ranks, frequently even converging to the effective rank of much lower rank updates (like $r=8$ and $r=16$). For example, LoRA $r=768$ has an effective rank that is consistently below 50 when $\alpha = 8$. We note that full fine-tuning has an effective rank update of above 400 consistently. These plots suggest that when $\alpha$ is kept fixed when scaling LoRA rank, the solutions are uable to take advantage of their higher expressability and instead \textit{converge to low rank solutions}.

\section{Impact of Matrix Percentage on Number of Intruder Dimensions}

\label{matrix-sweep-text-appendix}

\begin{figure}[h!]
    \centering
    \begin{subfigure}[b]{1.0\linewidth}
        \centering
        \includegraphics[width=\linewidth]{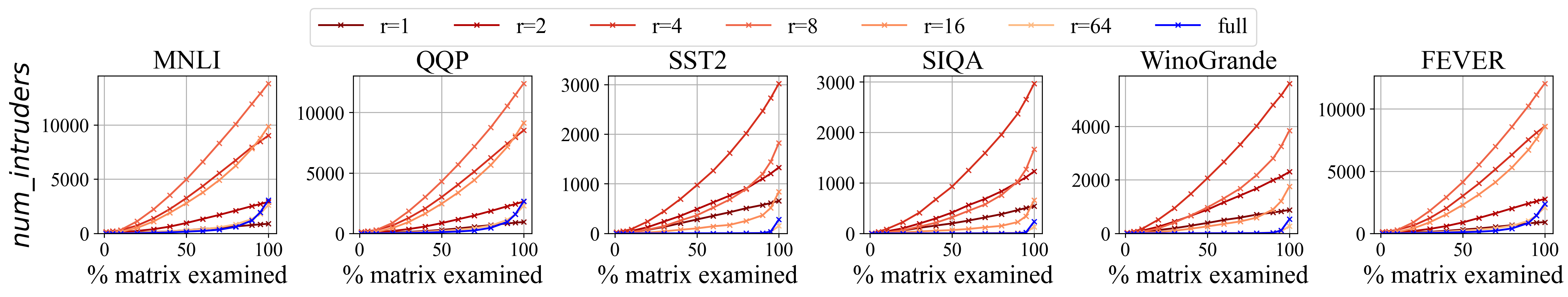}
        \caption{Impact of the number of singular vectors in the fine-tuned matrix we examine, $k$, on the number of intruder dimensions for RoBERTa models fine-tuned on 6 different tasks. Here, we set $\epsilon=0.5$.}
        \label{roberta-matrix}
    \end{subfigure}

    \begin{subfigure}[b]{0.26\linewidth}
        \centering
        \includegraphics[width=\linewidth]{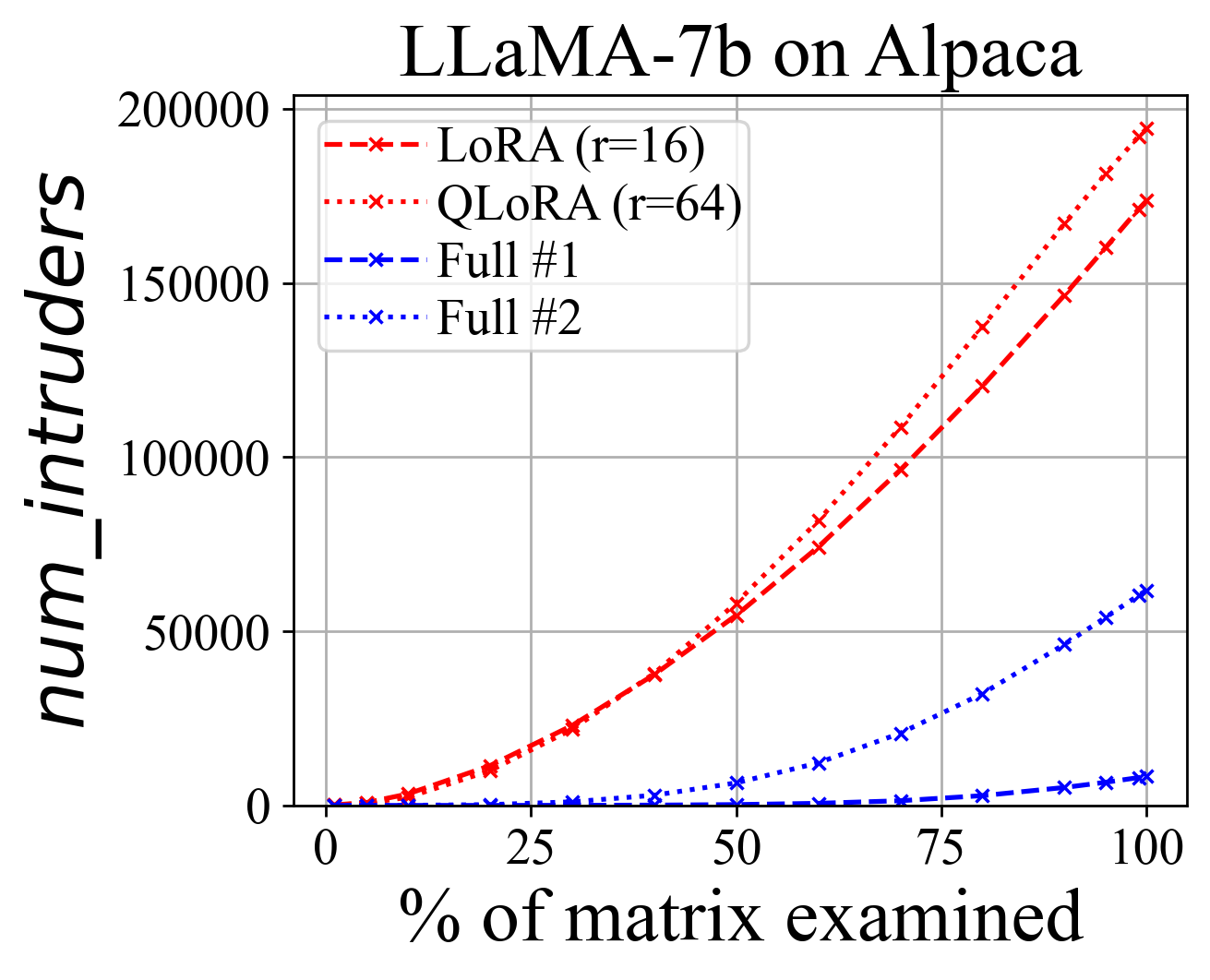}
        \caption{\centering LLaMA-7B fine-tuned on Alpaca.}
        \label{llama-matrix}
    \end{subfigure}
    \begin{subfigure}[b]{0.28\linewidth}
        \centering
        \includegraphics[width=\linewidth]{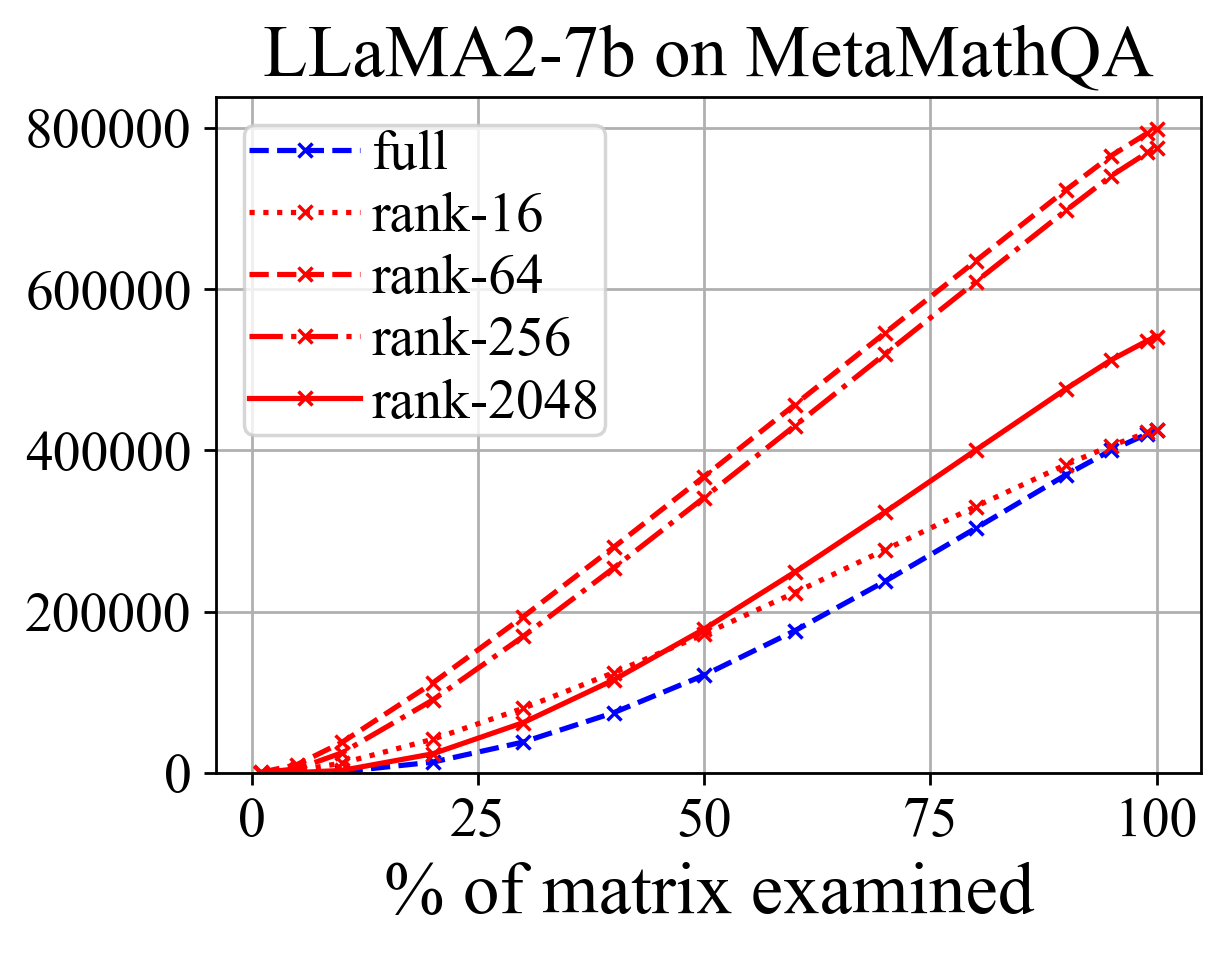}
        \caption{\centering LLaMA2-7B fine-tuned on MetaMathQA.}
        \label{llama2-metamath-matrix}
    \end{subfigure}
    \begin{subfigure}[b]{0.28\linewidth}
        \centering
        \includegraphics[width=\linewidth]{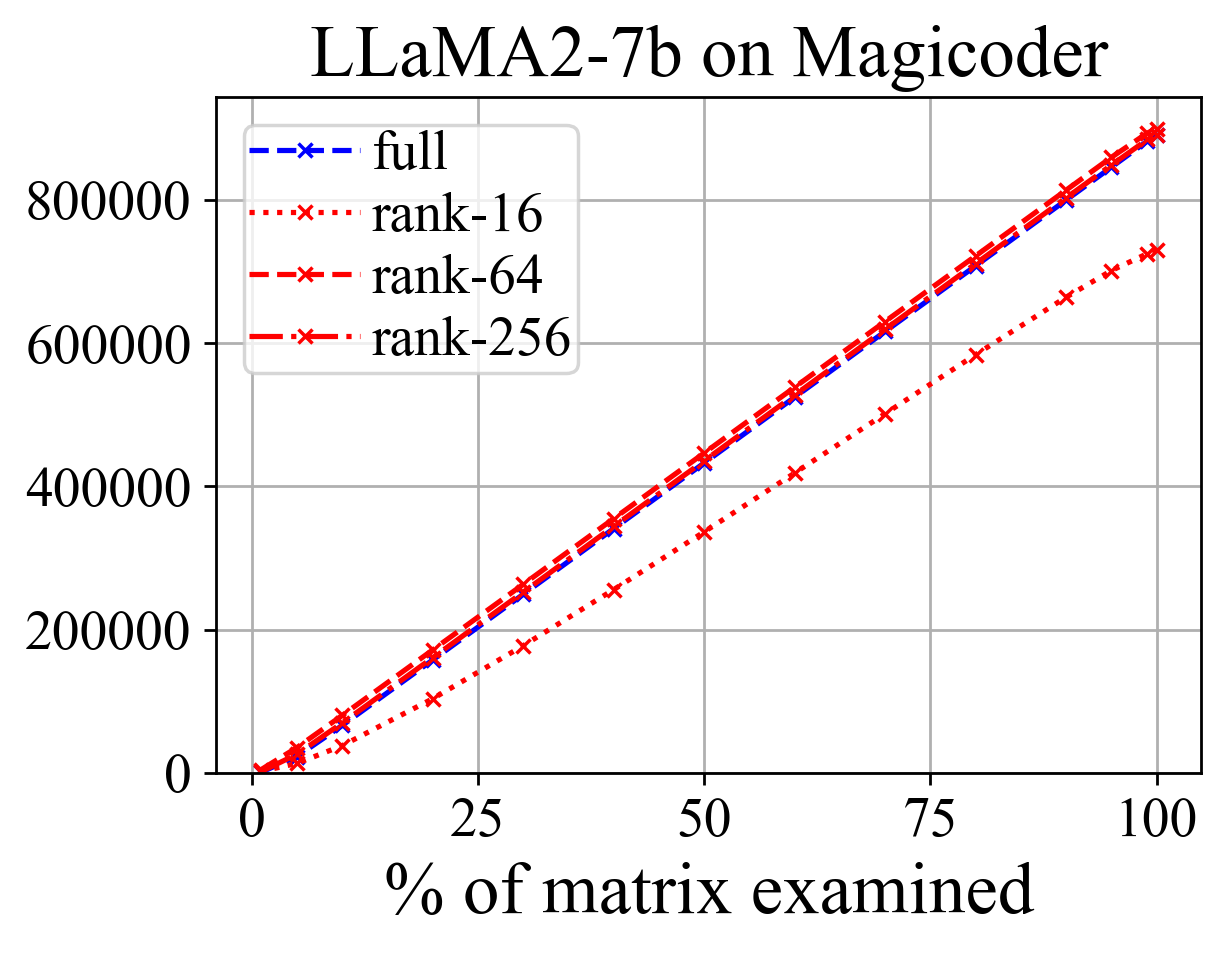}
        \caption{\centering LLaMA2-7B fine-tuned on Magicoder-Evol-Instruct.}
        \label{llama2-magicoder-matrix}
    \end{subfigure}
    \caption{\textbf{Impact of $k$, the number of fine-tuned singular vectors we examine, on the number of intruder dimensions.} We see that models fine-tuned with LoRA tend to have more intruder dimensions than full fine-tuning, regardless of the value of $k$ used.}
    \label{matrix-sweep}
    
\end{figure}

In this section, we examine the extent to which intruder dimensions exist throughout the entire weight matrix and how they are distributed. As described in the main text, we hold $\epsilon$ fixed as $\epsilon = 0.5$ and measure the number of intruder dimensions while varying the proportion of the fine-tuned singular vectors that we examine (this means varying our $k$ parameter in Algorithm \ref{intruder-alg}). Here, we can see that LoRA consistently has more intruder dimensions than full fine-tuning, regardless of what fraction of the singular values we examine. The only caveat to this is that, for some datasets, full fine-tuning passes LoRA $r=1$ when examining the last 20\% of the fine-tuned singular vectors. This is likely due to the limited expressivity of rank 1 updates and is interesting because it suggests that in this case, full fine-tuning may be changing lower-ranking singular vectors more than LoRA. One interesting contradiction to our findings is in Fig.~\ref{llama2-magicoder-matrix}, which shows that full fine-tuning and LoRA appear to have very similar distributions of intruder dimensions within their matrix when fine-tuned on code. This is likely due to the large domain shift from natural language to coding tasks (\citet{biderman2024loralearnsless} also make this observation of a large domain shift required for models fine-tuned on Magicoder \citep{wei2024magicoderempoweringcodegeneration}).

\FloatBarrier
\section{Impact of Dataset Size On Intruder Dimensions}

\label{dataset-size-text-appendix}
\begin{figure}[h]
    \centering
    \begin{subfigure}[b]{0.49\linewidth}
        \centering
        \includegraphics[width=\linewidth]{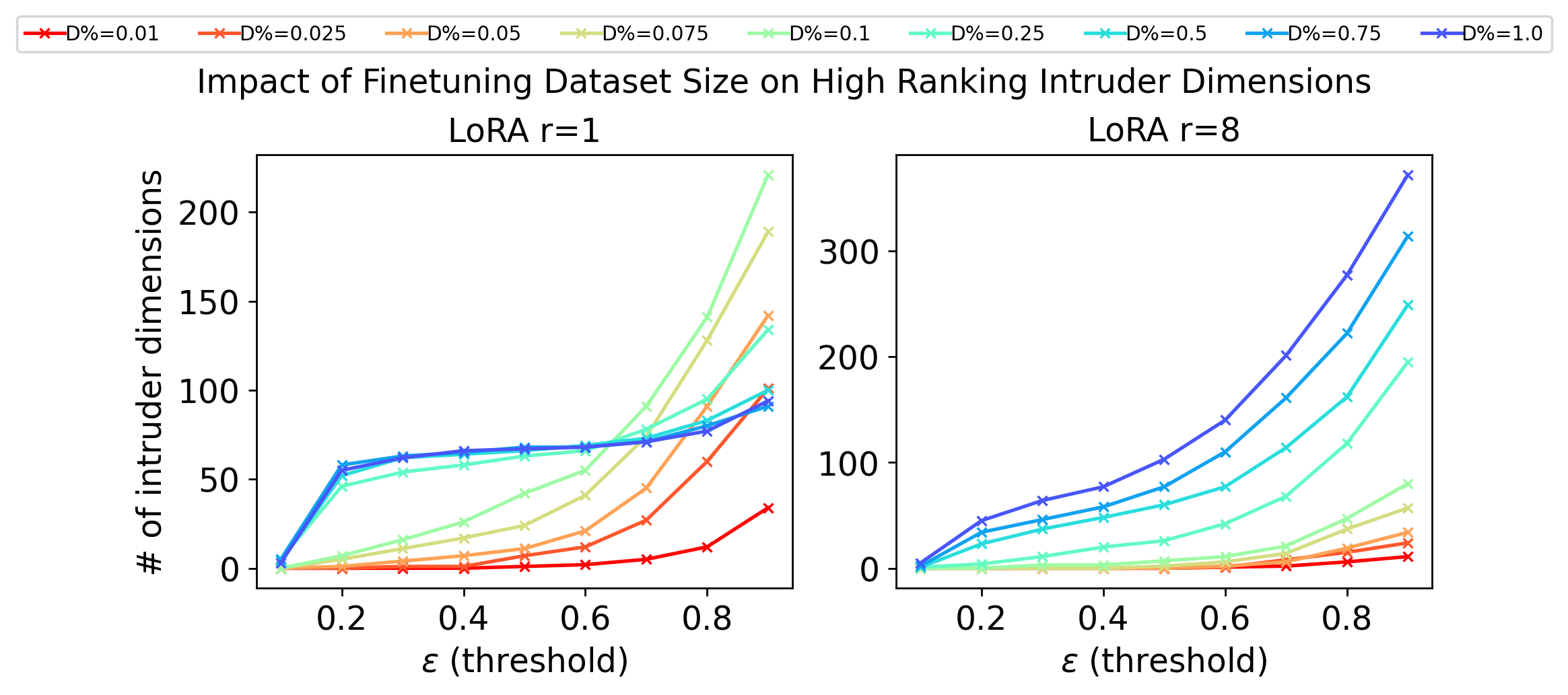}
    \end{subfigure}
    \begin{subfigure}[b]{0.49\linewidth}
        \centering
        \includegraphics[width=\linewidth]{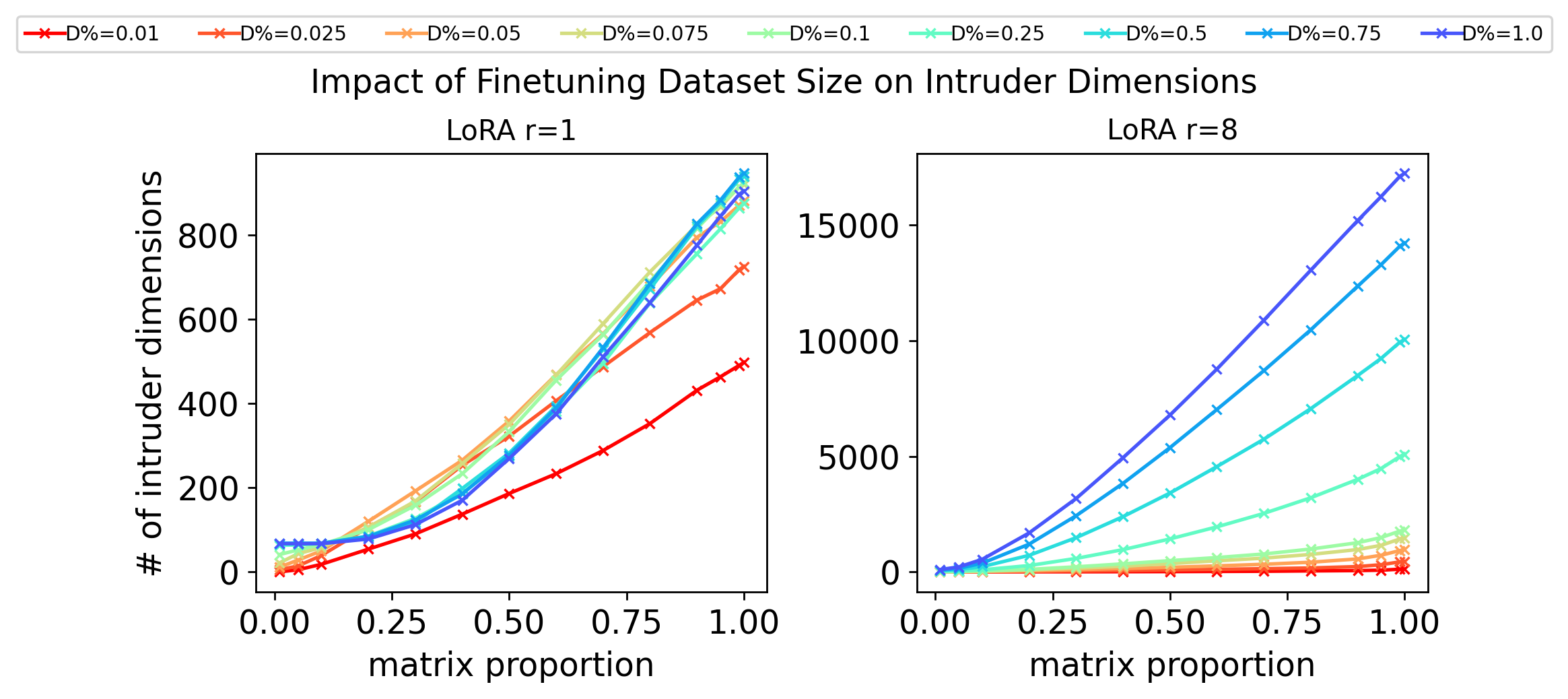}
    \end{subfigure}
    \caption{\textit{(Left)} Impact of cosine similarity threshold, $\epsilon$, on the number of intruder dimensions for LoRA models trained on different proportions of the MNLI dataset. \textit{(Right)} Impact of the number of fine-tuned singular vectors we examine, $k$, on the number of intruder dimensions for LoRA models trained on different proportions of the MNLI dataset. We see that training on a larger proportion of the dataset increases the number of intruder dimensions in the model. }
    \label{dataset-size-exp}
\end{figure}

\textbf{The total number of intruder dimensions increases proportionally to the size of the fine-tuning dataset.} Using our training recipe (Appendix \ref{section:roberta-details}), we fine-tuned models on data subsets of varying sizes. We trained RoBERTa-base on MNLI using LoRA with rank 1 and 8 (cases where we originally saw intruder dimensions) and measure the number of intruder dimensions along with the impact of $\epsilon$ and $k$ (Fig.~\ref{dataset-size-exp}). For $r=8$, as we train on more data, more intruder dimensions are introduced. Interestingly, however, LoRA with rank 1 appears to converge to similar amounts of intruder dimensions, regardless of the dataset size. This may be because of the limited expressivity of models with $r=1$. This experiments suggest that with smaller datasets, fewer intruder dimensions may be introduced by LoRA.

\pagebreak

\section{LLaMA/LLaMA-2 Instruction Tuned Models}

\label{huggingface-details}

 Our LLaMA-7B checkpoints were fine-tuned on the Alpaca \citep{alpaca_dataset} and consist of two fully fine-tuned models, one LoRA model with rank 16, and one QLoRA \citep{qlora} model with rank 64. Our LLaMA2-7B checkpoints were fine-tuned on either code (IFT with Magicoder-Evol-Instruct-110K \citep{wei2024magicoderempoweringcodegeneration} or CPT with StarCoder \citep{li2023starcodersourceyou}) or math (IFT with MetaMathQA \citep{yu2024metamathbootstrapmathematicalquestions} or CPT with OpenWebMath \citep{paster2023openwebmathopendatasethighquality}) and consist of one fully fine-tuned model and 3-4 LoRA'ed models of different ranks for each dataset and generously provided by \citet{biderman2024loralearnsless}. In Fig.~\ref{llama-epsilon}, Full \#1 refers to ``PKU-Alignment/alpaca-7b-reproduced" and Full \#2 refers to ``chavinlo/alpaca-native".

\begin{table}[h]
\centering
\small
\begin{tabular}{|c|c|c|c|c|}
\hline
 \textbf{Hugging Face Path} & Base Model & IT Dataset \\
\hline
timdettmers/qlora-alpaca-7b & LLaMA-7b & Alpaca\\
\hline
tloen/alpaca-lora-7b & LLaMA-7b & Alpaca\\
\hline
PKU-Alignment/alpaca-7b-reproduced & LLaMA-7b & Alpaca \\
\hline
chavinlo/alpaca-native & LLaMA-7b & Alpaca\\
\hline
LoRA-TMLR-2024/magicoder-lora-rank-16-alpha-32 & LLaMA2-7b & Magicoder \\
\hline
LoRA-TMLR-2024/magicoder-lora-rank-64-alpha-128 & LLaMA2-7b & Magicoder \\
\hline
LoRA-TMLR-2024/magicoder-lora-rank-256-alpha-512 & LLaMA2-7b & Magicoder \\
\hline
LoRA-TMLR-2024/magicoder-full-finetuning-lr-5e-05 & LLaMA2-7b & Magicoder \\
\hline
LoRA-TMLR-2024/metamath-lora-rank-16-alpha-32 & LLaMA2-7b & MetaMath \\
\hline
LoRA-TMLR-2024/metamath-lora-rank-64-alpha-128 & LLaMA2-7b & MetaMath \\
\hline
LoRA-TMLR-2024/metamath-lora-rank-256-alpha-512 & LLaMA2-7b & MetaMath \\
\hline
LoRA-TMLR-2024/metamath-full-finetuning-lr-1e-05 & LLaMA2-7b & MetaMath \\
\hline
LoRA-TMLR-2024/starcoder-lora-rank-16-20B-tokens & LLaMA2-7b & StarCoder \\
\hline
LoRA-TMLR-2024/starcoder-lora-rank-64-20B-tokens & LLaMA2-7b & StarCoder \\
\hline
LoRA-TMLR-2024/starcoder-lora-rank-256-20B-tokens & LLaMA2-7b & StarCoder \\
\hline
LoRA-TMLR-2024/starcoder-full-finetuning-lr-1e-05-20B-token & LLaMA2-7b & StarCoder \\
\hline
LoRA-TMLR-2024/openwebmath-lora-rank-16-20B-tokens & LLaMA2-7b & OpenWebMath \\
\hline
LoRA-TMLR-2024/openwebmath-lora-rank-64-20B-tokens & LLaMA2-7b & OpenWebMath \\
\hline
LoRA-TMLR-2024/openwebmath-lora-rank-256-20B-tokens & LLaMA2-7b & OpenWebMath \\
\hline
LoRA-TMLR-2024/openwebmath-full-finetuning-lr-1e-05-20B-tokens & LLaMA2-7b & OpenWebMath \\
\hline

\end{tabular}
\caption{Hugging Face model paths for LLaMA-7b/LLaMA2-7b IT models.}
\end{table}
\label{huggingface-llama-models}

\FloatBarrier

\section{Continual Learning}

\label{continual-learning-full-text-appendix}
\subsection{Performance during continual learning}
\begin{figure}
    \centering
    \includegraphics[width=0.9\linewidth]{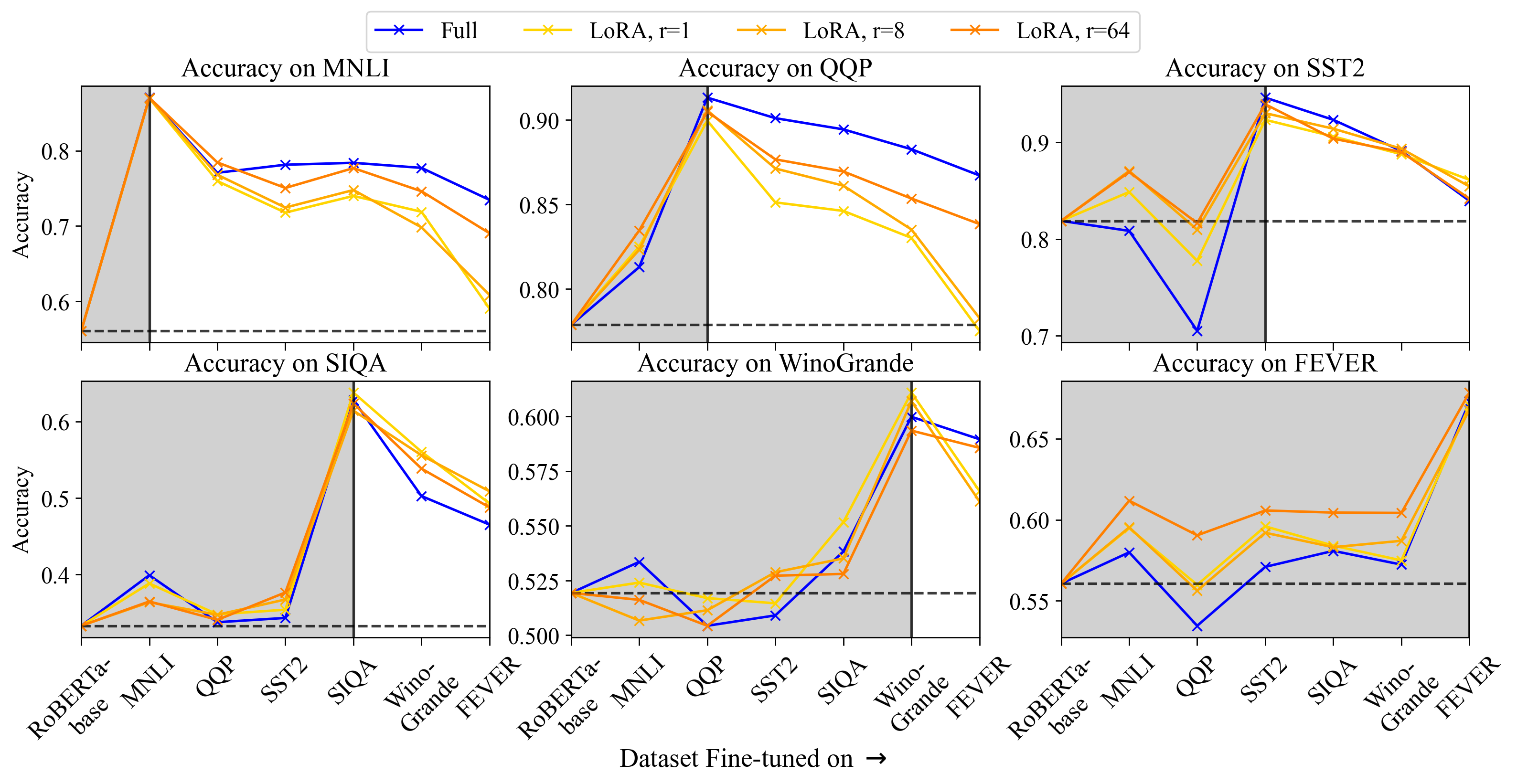}
    \caption{Full plot of Fig.~\ref{continual-learning-vertical}. Continual Learning performance of RoBERTa for full fine-tuning and LoRA. We sequentially train on six tasks, in order from left to right. Horizontal dotted line indicates baseline pre-trained performance. Vertical solid line indicates when a specific dataset is fine-tuned on. Gray region represents performance before the model has been trained on that task. We are interested in the differences in accuracies of these methods both right after training (at the vertical black line) and later (in the white region). We see that low ranks of LoRA forget previously learned tasks more.}
    \label{continual_learning_full}
\end{figure}

As described in Section \ref{behavior-section} in the main text, we train sequentially on 6 tasks and measure task performance on all of these tasks across tasks trained on (continual learning). We report the full graph of our findings in Fig.~\ref{continual_learning_full}. In it, we find that when all our models are trained to similar accuracy, lower ranks of LoRA, which coincide with more intruder dimensions, forget more of their previously learned tasks than higher ranks of LoRA and full fine-tuning. 

\subsection{Similarity matrices during continual learning}

\begin{figure}
    \centering
    \begin{subfigure}[b]{1.0\linewidth}
        \centering
        \includegraphics[width=0.9\linewidth]{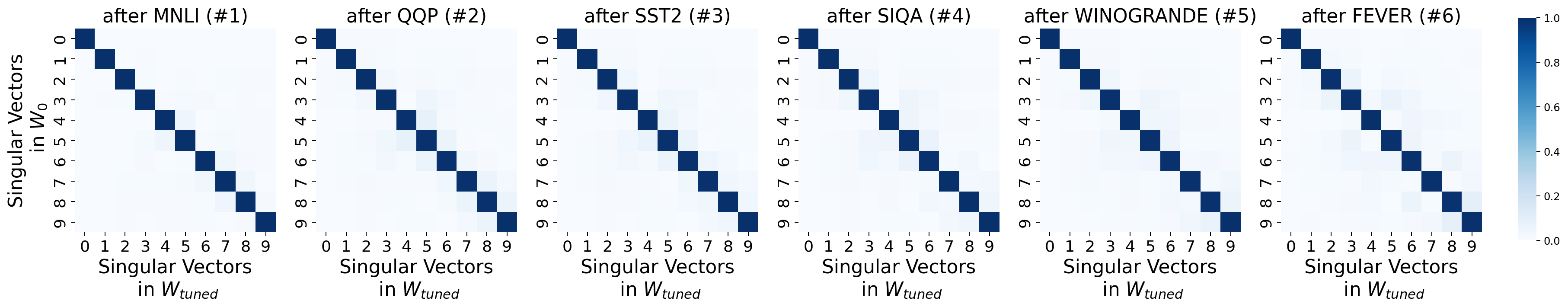}
        \caption{Continual learning similarity matrices for full fine-tuning.}
        \label{continual-learning-heatmap-full}
    \end{subfigure}
    \begin{subfigure}[b]{1.0\linewidth}
        \centering
        \includegraphics[width=0.9\linewidth]{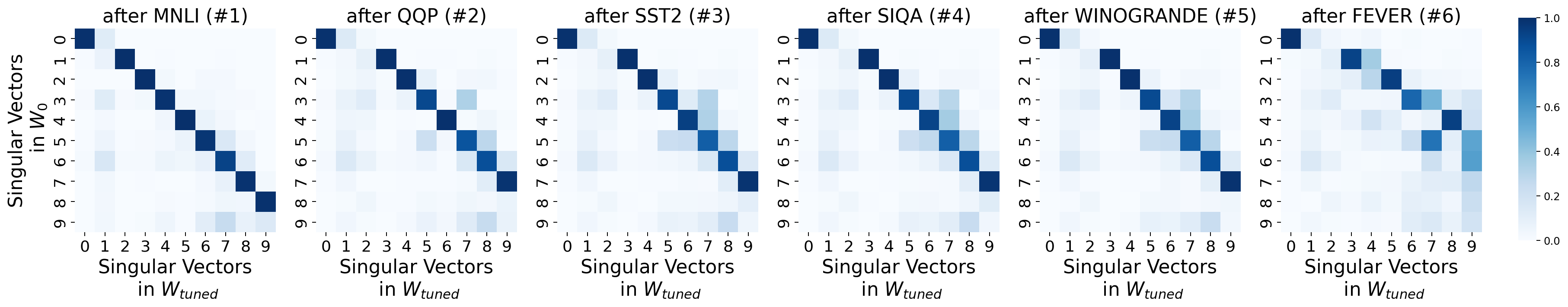}
        \caption{Continual learning similarity matrices for LoRA.}
        \label{continual-learning-heatmap-lora}
    \end{subfigure}

    \caption{\textbf{LoRA accumulates intruder dimensions, while full fine-tuning does not.} The pre-trained structure of the model degrades across tasks trained on.}
    \label{continual-learning-heatmaps}
\end{figure}

After each continual learning dataset we fine-tune on, we measure the similarity matrix between the current model and the pre-trained model. In Fig.~\ref{continual-learning-heatmap-lora}, we observe that LoRA accumulates intruder dimensions across fine-tuning datasets. In contrast, in Fig.~\ref{continual-learning-heatmap-full} we observe that the pre-trained structure of the model is retained well across fine-tuning datasets. These experiments suggest why LoRA appears to degrade faster during continual learning.

\FloatBarrier
\section{Case Study: Setting Alpha=8 instead of Alpha=2r}
\label{case-study-text-appendix}

\begin{figure}[h!]
        \centering
        \begin{subfigure}[b]{1.0\linewidth}
            \centering
            \includegraphics[width=\textwidth]{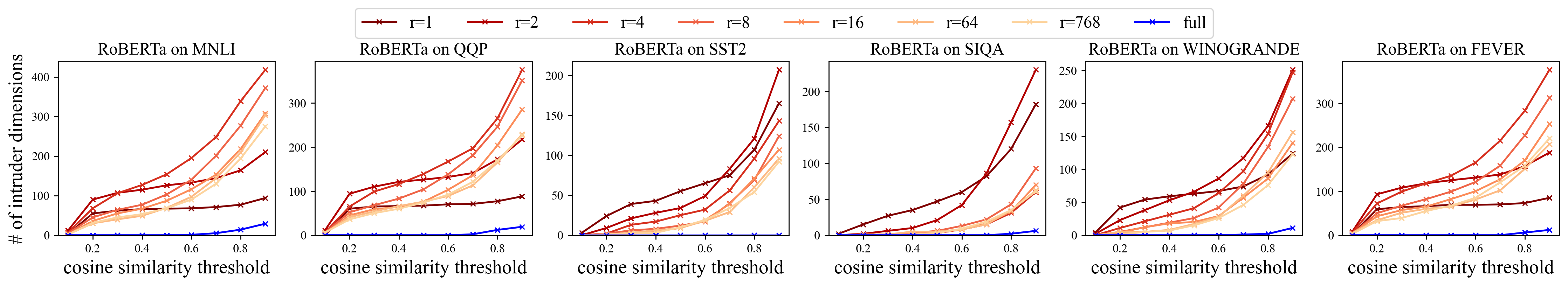}
            \caption{Number of intruder dimensions in RoBERTa models fine-tuned on 6 different tasks. Here, we set $k=10$. We use the same conditions as in Fig.~\ref{roberta-epsilon} but instead now set $\alpha = 8$ instead of $\alpha = 2r$.}
            \label{roberta-epsilon-a8}

        \end{subfigure}

        \begin{subfigure}[b]{1.0\linewidth}
            \centering
            \includegraphics[width=\textwidth]{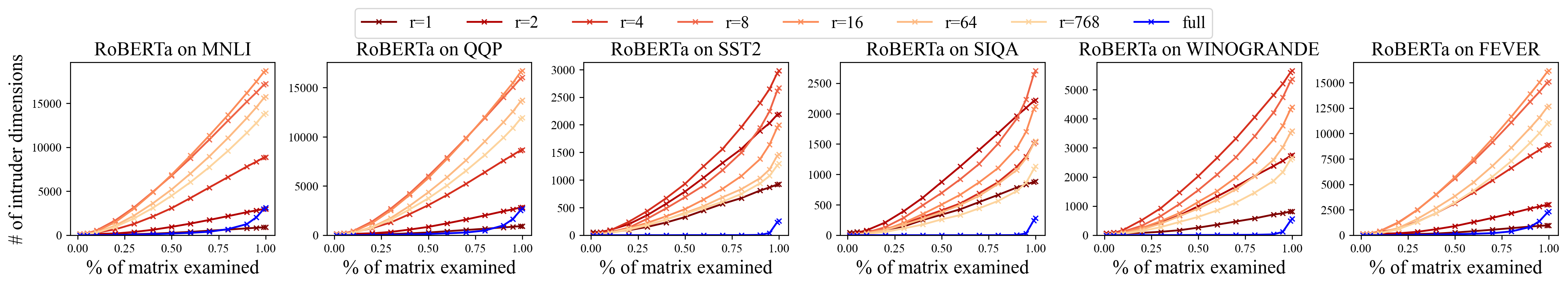}
            \caption{Impact of the number of singular vectors in the fine-tuned matrix we examine, $k$, on the number of intruder dimensions for RoBERTa models fine-tuned on 6 different tasks. Here, we set $\epsilon=0.5$. We use the same conditions as in Fig.~\ref{roberta-matrix} but instead now set $\alpha = 8$ instead of $\alpha = 2r$.}
            \label{roberta-matrix-a8}
        \end{subfigure}

        \caption{We find that when $\alpha=8$ instead of $\alpha = 2r$, our models have more intruder dimensions. \textit{(Top)} Replication of Fig.~\ref{roberta-epsilon} with $\alpha = 8$ instead of $\alpha = 2r$. \textit{(Bottom)} Replication of Fig.~\ref{roberta-matrix} with $\alpha = 8$ instead of $\alpha = 2r$.}

    \end{figure}

\begin{figure}[h!]
        \centering
        \includegraphics[width=0.8\linewidth]{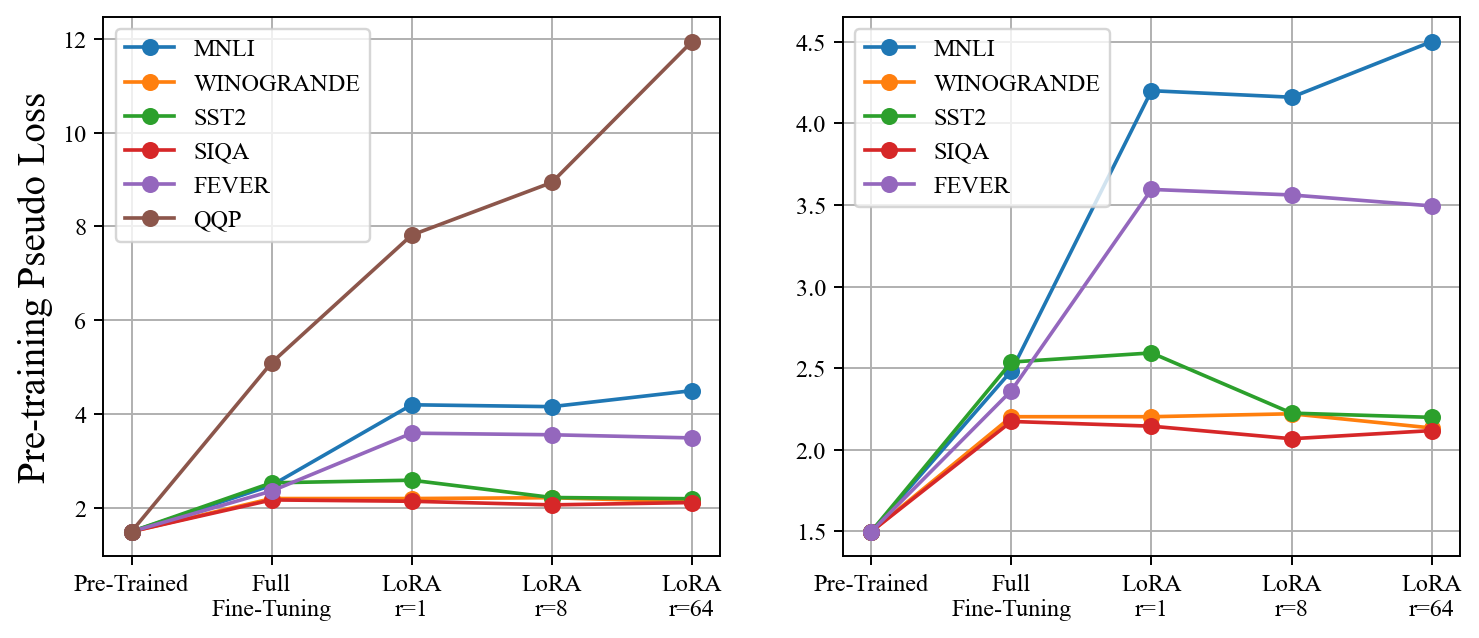}
        \caption{For $\alpha = 8$. RoBERTa's performance on its pre-training data distribution after fine-tuning on a particular task. We measure pseudo loss as described by \citet{mlm_scoring}. We compare these results to when $\alpha=2r$ (Fig.~\ref{pretraining_drift}).}
        \label{pretraining_drift_a8}
    \end{figure}

\begin{figure}[h!]
        \centering
        \includegraphics[width=0.95\textwidth]{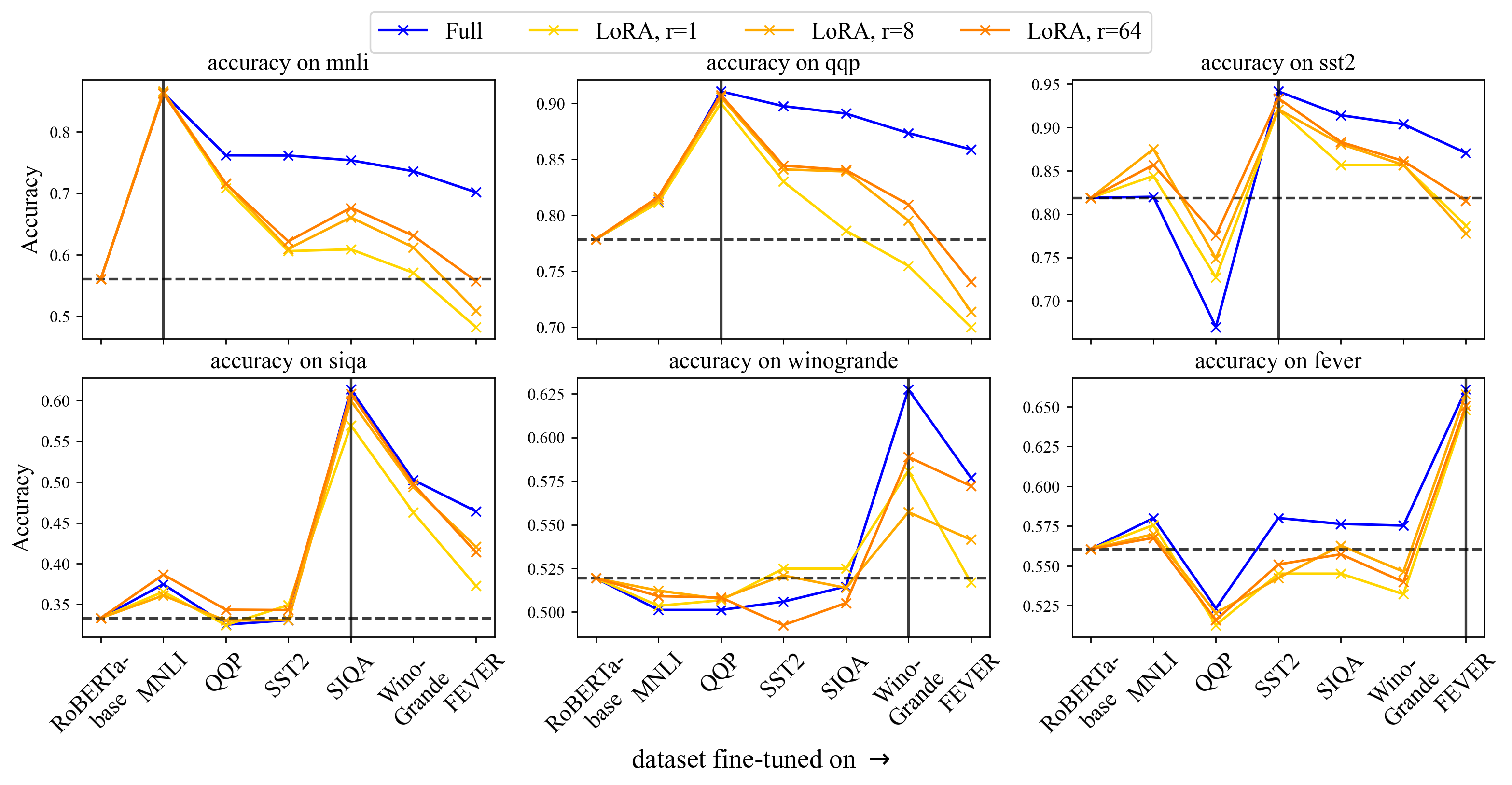}
        \caption{For $\alpha = 8$. RoBERTa's performance on six datasets during continual learning. We sequentially train on six tasks, in order from left to right. Horizontal dotted line indicates baseline pre-trained performance. Vertical solid line indicates when a specific dataset is fine-tuned on. We compare these results to when $\alpha=2r$ (Fig.~\ref{continual_learning_full}).}
        \label{continual_learning_a8}
    \end{figure}

Our main experiments were conducted with $\alpha = 2r$. However, \citet{lora} instead set $\alpha = 8$ for RoBERTa-base. While not the recommended practice now, we explore what impact this selection has on our findings. We report our key plots in Fig.~\ref{roberta-epsilon-a8}, \ref{roberta-matrix-a8}, \ref{pretraining_drift_a8}, \ref{continual_learning_a8}, \& \ref{roberta-all-effective-ranks=a8}. 
In Fig.~\ref{roberta-epsilon-a8} \& \ref{roberta-matrix-a8} we see that LoRA'd models with high rank have significantly more intruder dimensions in comparison to when $\alpha=2r$. Interestingly, whereas when $\alpha=2r$ LoRA models with ranks like 64 had no or very few intruder dimensions (see Fig.~\ref{epsilon-sweep}), they now have numerous intruder dimensions.
These differences are corroborated by Fig.~\ref{roberta-all-effective-ranks=a8}, where we see that the learned solutions of LoRA have significantly lower effective rank in comparison to when $\alpha = 2r$. For example, we see in Fig.~
\ref{roberta-all-effective-ranks=a8} that when LoRA has a rank of 768, the effective rank is never above 100. In contrast, we see in Fig.~
\ref{fig:all-effective-ranks-a2r} that with the same rank of 768, LoRA always has an effective rank above 768. This suggests that when $\alpha=8$, LoRA is converging to lower rank solutions than when $\alpha=2r$. This supports the finding that setting $\alpha=2r$ improves LoRA's performance when a high rank is used \citep{biderman2024loralearnsless, kalajdzievski2023rankstabilizationscalingfactor}.
Behaviorally, we see in Fig.~\ref{continual_learning_a8} that LoRA models with high rank have much more forgetting on previously learned tasks in comparison to full fine-tuning and LoRA when $\alpha = 2r$ is used ($\alpha = 2r$ results are in Fig.~\ref{continual_learning_full}). Likewise, in Fig.~\ref{roberta-all-effective-ranks=a8} we see that when LoRA has high rank, it has much more forgetting on the pre-trained distribution in comparison to LoRA when $\alpha = 2r$.

    \section{Impact of Random Seeds}

    \label{random-seeds-text-appendix}

    To ensure that random seed does not play a role in the number of intruder dimensions we observe in a model, we sample 5 different seeds and fine-tune RoBERTa-base on MNLI using the same methodology as in Fig.~\ref{roberta-epsilon}. We find that the initialization has a negligible role on the number of intruder dimensions. This shows that our findings are not dependent on the random initialization of the LoRA modules.

    \begin{figure}
        \centering
        \includegraphics[width=0.3\linewidth]{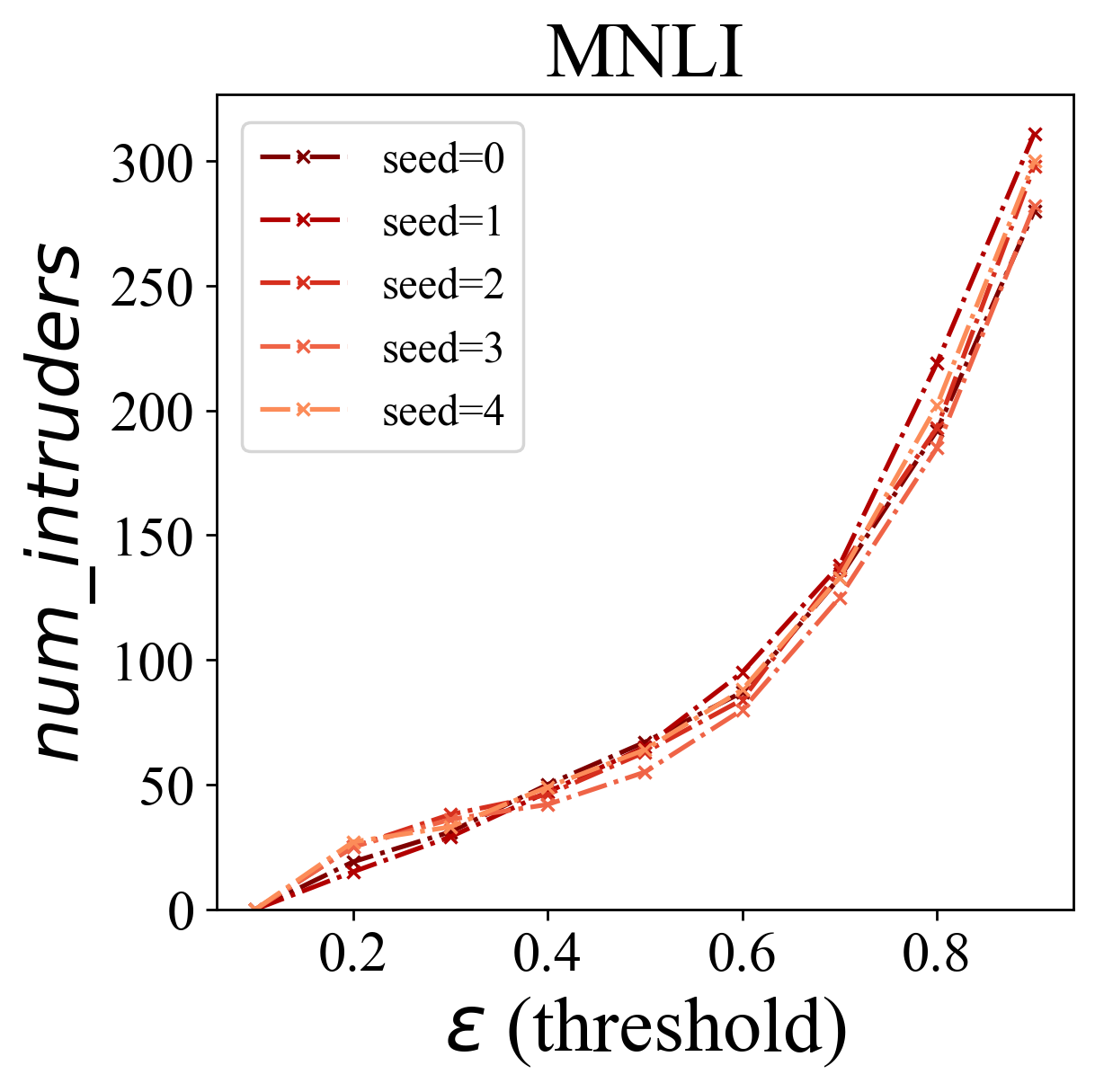}
        \caption{\textbf{Impact of Random Seeds on intruder dimensions.} We fine-tune RoBERTa-base across 5 random seeds and use our same methodology as in Fig.~\ref{roberta-epsilon}. We find that the initialization has a negligible role on the number of intruder dimensions. This shows that our findings are not dependent on the random initialization of the LoRA modules.}
        \label{fig:random_seeds}
    \end{figure}

    \section{LoRA Variants}

    \label{lora-variants-text-appendix}

    We focus significantly on the standard LoRA method proposed by \citet{lora} in order to study it in depth. However, many variants of LoRA have been proposed recently. AdaLoRA \citep{zhang2023adaloraadaptivebudgetallocation} adaptively allocates LoRA rank to different modules in order to ensure optimal allocation of trainable parameters to certain modules. LoRA+ \citep{lora+} sets different learning rates for the A and B modules in LoRA. PiSSA \citep{meng2024pissaprincipalsingularvalues} initializes the A and B modules with the top ranking singular vectors of the pre-trained weights. VeRA \citep{vera} models LoRA as the product of two random matrices with trainable parameters doing elementwise operations on the resulting vectors. These variants may have important impacts on the presence of intruder dimensions. For example, PiSSA initializes with the singular vectors and therefore may have an easier time changing them, possibly leading to more intruder dimensions. In contrast, LoRA+ in effect lowers the learning rate, which we found to be important to introducing intruder dimensions, and may therefore reduce the number of intruder dimensions.

\begin{figure}
        \centering
        \includegraphics[width=0.55\linewidth]{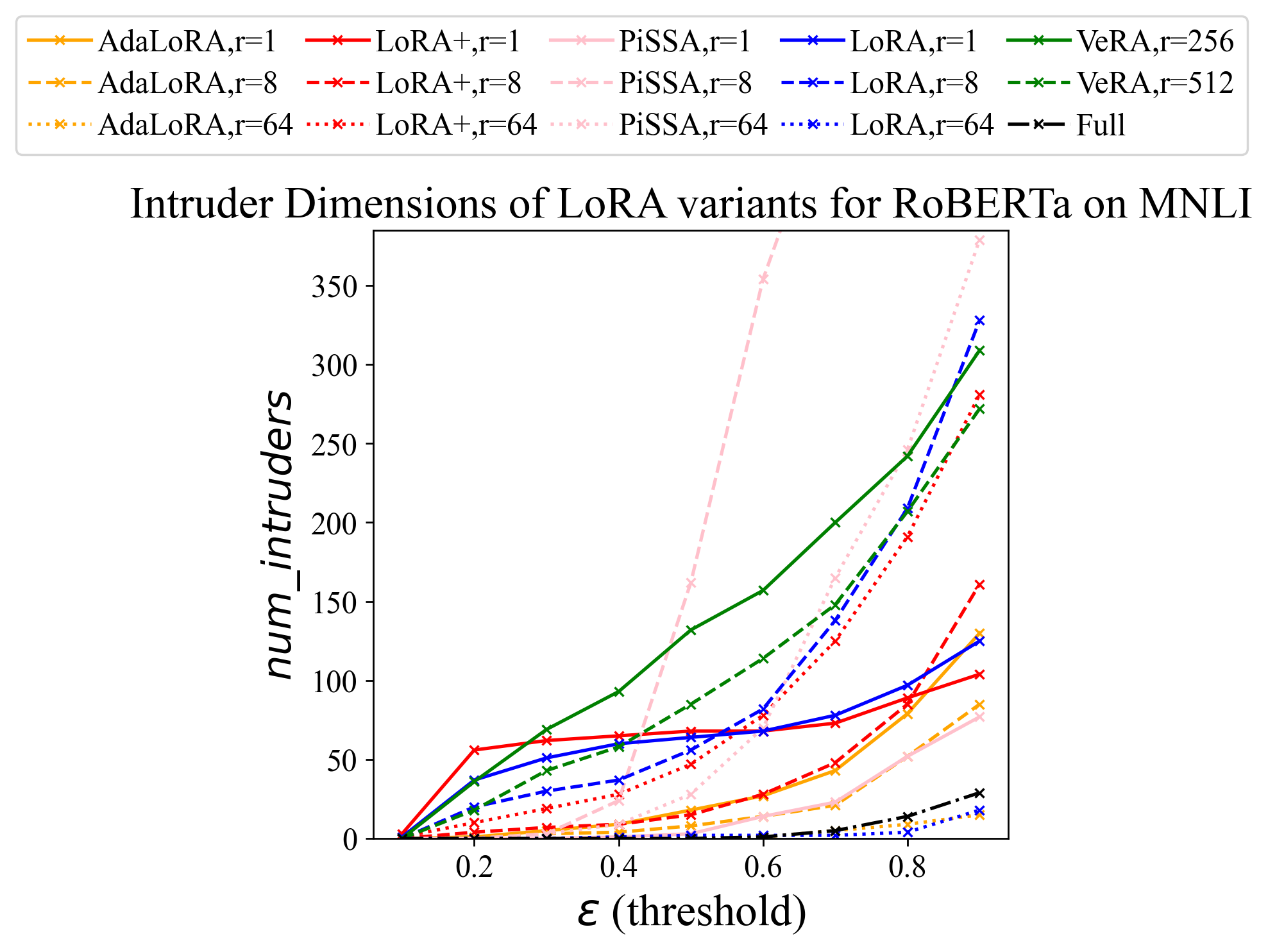}
        \caption{\textbf{Measuring LoRA variants for intruder dimensions.} $k=10$. We compare variants of LoRA to normal LoRA (blue) and full fine-tuning (black). We find that the LoRA variants we examine still have intruder dimensions and shows that our findings are not just exclusive to normal LoRA.}
        \label{fig:lora-variants}
    \end{figure}

     In order to examine if intruder dimensions are still relevant for these methods, we rerun our MNLI fine-tuning experiment with RoBERTa with each of these methods with default hyperparameters that they provide. These results are supplied in Fig.~\ref{fig:lora-variants}. Interestingly, we see that all the variations we examine have intruder dimensions. Some interesting observations include: LoRA+ and LoRA $r=1$ appear to have nearly identical curves, suggesting they have very ismilar intruder dimension characteristics. We again see that with higher ranks ($r=64$) these LoRA variants tend to have very few intruder dimensions. However, it does appear that methods that explicitly modify the singular vectors, like PiSSA, have many intruder dimensions. This makes sense since they are explicitly constructed to modify the singular vectors on the pre-trained model. These findings emphasize that intruder dimensions are not just an observed phenomenom in normal LoRA and suggests to future work the examination of LoRA variants.

\end{document}